\documentclass{article}

\usepackage[final,nonatbib]{neurips_2020}

\usepackage[utf8]{inputenc} 
\usepackage[T1]{fontenc}    
\usepackage{hyperref}       
\usepackage{url}            
\usepackage{booktabs}       
\usepackage{amsfonts}       
\usepackage{nicefrac}       
\usepackage{microtype}      
\usepackage{graphicx}
\usepackage{amsmath}
\usepackage{xcolor}
\usepackage{tabularx}
\usepackage{multirow}
\usepackage[percent]{overpic}
\usepackage{textcomp}

\title{The Utility of Decorrelating Colour Spaces in Vector Quantised Variational Autoencoders}

\author{%
  Arash Akbarinia \And Raquel Gil-Rodr{\'\i}guez \And Alban Flachot \And Matteo Toscani \And \vspace{-0.8cm}  \\
   Department of General Psychology \\
   Justus-Liebig University, Giessen, Germany \\
   \texttt{arash.akbarinia@psychol.uni-giessen.de} \\
}

\begin{document}

\maketitle

\begin{abstract}

    Vector quantised variational autoencoders (VQ-VAE) are characterised by three main components: 1) encoding visual data, 2) assigning $k$ different vectors in the so-called \textit{embedding space}, and 3) decoding the learnt features. While images are often represented in RGB colour space, the specific organisation of colours in other spaces also offer interesting features, e.g. CIE L*a*b* decorrelates chromaticity into opponent axes. In this article, we propose colour space conversion---a simple quasi-unsupervised task---to enforce a network learning structured representations. To this end, we trained several instances of VQ-VAE whose input is an image in one colour space, and its output in another, e.g. from RGB to CIE L*a*b* (in total five colour spaces were considered). We examined the finite embedding space of trained networks in order to disentangle the colour representation in VQ-VAE models. Our analysis suggests that certain vectors encode hue and others luminance information. We further evaluated the quality of reconstructed images at low-level using pixel-wise colour metrics, and at high-level by inputting them to image classification and scene segmentation networks. We conducted experiments in three benchmark datasets: ImageNet, COCO and CelebA. Our results show, with respect to the baseline network (whose input and output are RGB), colour conversion to decorrelated spaces obtains 1-2$\Delta E$ lower colour difference and 5-10\% higher classification accuracy. We also observed that the learnt embedding space is easier to interpret in colour opponent models.
\end{abstract}

\section{Introduction}

Colour is an inseparable component of our conscious visual perception and its objective utility spans over a large set of tasks such as object recognition and scene segmentation \cite{chirimuuta2015uses,Gegenfurtner&Rieger2000, WichmannSharpe&Gegenfurtner2002}. The colour vision sensory system begins in our retina with three types of cone photoreceptors whose sensitivity functions strongly correlate \cite{Buchsbaum&Gottschalk1983, Gegenfurtner&Kiper2003}. Horizontal and ganglion cells decorrelate this signal into what is known as colour-opponency before its transmission to the visual cortex \cite{schiller1977properties,derrington1984chromatic,Gegenfurtner&Kiper2003}. This colour space transformation (from cone excitation to opponency) increases the efficiency of the system in information coding \cite{Buchsbaum&Gottschalk1983, Ruderman1998,lee2001color}. Correspondingly, opponent colour spaces---like CIE L*a*b*---proved themselves effective in several classical computer vision algorithms \cite{gevers2004robust, stokman2007selection, creusen2010color}. This has been less explored in the modern era of deep-learning except in applications such as style transfer \cite{luan2017deep, gatys2017controlling} and picture colourisation \cite{cheng2015deep, larsson2016learning}. 
Nevertheless, colour opponency appears to be an inherent in deep learning models. Networks trained for high-level visual tasks learn to decorrelate their inputs \cite{rafegas2018color, flachot2018processing, harris2019spatial}.

A profound grasp on colour neural encoding in deep networks is of great interest, at least, from two standpoints. From a practical perspective, the ``ground-truth'' of many applications, such as colour transfer \cite{rodriguez2020color}, colour constancy \cite{chakrabarti2015color}, style transfer \cite{luan2017deep}, computer graphics \cite{Bratkova2009}, image denoising \cite{Dabov2007} and quality assessment \cite{Preiss2014}, is human colour perception. Advancement in these lines requires a better understanding of colour representation in deep networks. From a theoretical point of view, the neural basis of some aspects of our colour vision, such as unique hues and colour categories, remains to be explained \cite{parraga2016nice,witzel2018color, siuda2019biological}. Insights on whether or how these phenomena emerge in artificial networks can shed light on some of these open questions. Motivated by this, we designed a novel task---colour conversion---to be learnt by Vector Quantised Variational Autoencoders (VQ-VAE) \cite{van2017neural}. This simple task helps comprehension of the network's colour representation by means of comparing the learnt embedding spaces across different instances and colour spaces. We hypothesised a network that, for instance, converts the visual signal from RGB (a space with correlated colour coordinates) to CIE L*a*b (a perceptual colour space with decorrelated coordinates) would exhibit a twofold functionality: better capturing human colour perception and more efficiently conveying colour information in its compressing bottleneck.

The quantification of the latter is more straight forward \cite{schwarz1987experimental}. We measured the efficiency of colour converting networks in two ways: (i) low-level pixel-wise colour difference to the expected target and (ii) high-level global utility to a visual task. The entanglement of the former to the subjective human perception is best qualitatively analysed. We formalised this by (i) computing the transformations each embedding vector encodes and (ii) measuring its impact on luminance and chromaticity axes of the colour space. Grounded on these analyses, we demonstrate decorrelating colour spaces is beneficial for deep networks in line with the \textit{efficient coding} theory \cite{Barlow1961, Zhaoping2006}. We present an explanation at the level of embedding vectors according to the \textit{histogram equalisation} technique \cite{pratt2007digital}. Finally, we suggest a novel approach to conceptualise the learnt features of those vectors by a parsimonious linear transformation matrix.

\subsection{Related Work}

Colour spaces have been investigated in a few empirical studies of deep neural networks (DNNs). Information fusion over several colour spaces improves retinal medical imaging \cite{fu2019evaluation}. A similar strategy enhances the robustness of face \cite{li2014deepreid, larbi2018deepcolorfasd} and traffic light recognition \cite{cirecsan2012multi, kim2018efficient}. This has also been explored in predicting eye fixation \cite{shen2015predicting}. Nevertheless, the great majority of the deep learning vision models work with input RGB images.


The theoretical perspective of neural information processing predicts that, because of the constraints on the visual system (i.e. the number of neurons and the metabolic cost of neural activity), visual processing must have an efficient strategy for transmitting information (i.e. \textit{efficient coding} \cite{Barlow1961, Zhaoping2006}). The idea is to relay receptorial input by encoding it in an efficient and compressed manner to maximise neural resources. An example from the evolution of colour vision is the orthogonalisation  of cone responses  which yield colour opponent signals \cite{Zaidi1997,Ruderman1998,Buchsbaum&Gottschalk1983, Gegenfurtner&Kiper2003}. Crucially, efficient coding should match the statistics of the input signals \cite{Ming&Holt2009}. For instance, colour properties and spatial structures of cells receptive fields in the early visual cortex of the rhesus monkey match very well to the cone excitation from natural scenes \cite{lee2001color, van1998independent, hyvarinen2000emergence}. An analogy of \textit{efficient coding} appears in autoencoders due to the narrow nature of their bottlenecks. Due to this constraint, these artificial visual systems must incorporate an efficient strategy to transmit and represent information.



\section{Colour Conversion Autoencoders}

The ultimate question we would like to answer is how colour gets encoded in deep neural networks (DNNs). At this point, the aim is towards a more generic colour representation rather than describing colour contribution to a specific task. To this end, autoencoders are a suitable tool since their objective is to learn an efficient representation for a set of data in an unsupervised manner \cite{kramer1991nonlinear}. We studied a particular class of these networks---Vector Quantised Variational Autoencoder (VQ-VAE) \cite{van2017neural}---due to the discrete nature of its latent embedding space that facilitates the analysis and interpretability of the learnt features. This distinguishes VQ-VAE from other groups of variational encoders \cite{kingma2013auto}.

To better address our research question, we propose a novel quasi-unsupervised task of colour conversion: the network's output colour space is distinct from its input (see Figure~\ref{fig:flow}). A colour space is an arbitrary definition of colours' organisation in the space \cite{koenderink2003perspectives}. Hence, this simple extension offers the opportunity to learn more about the fundamentals of networks' internal colour space and whether it underlines a specific set of features. Eventually, we are interested to relate this to human-defined colour spaces with simple mathematical transformation matrices to enhance the interpretability of networks' internal representation.

\subsection{Networks}

\begin{figure}[t]
    \centering
    \includegraphics[width=\linewidth]{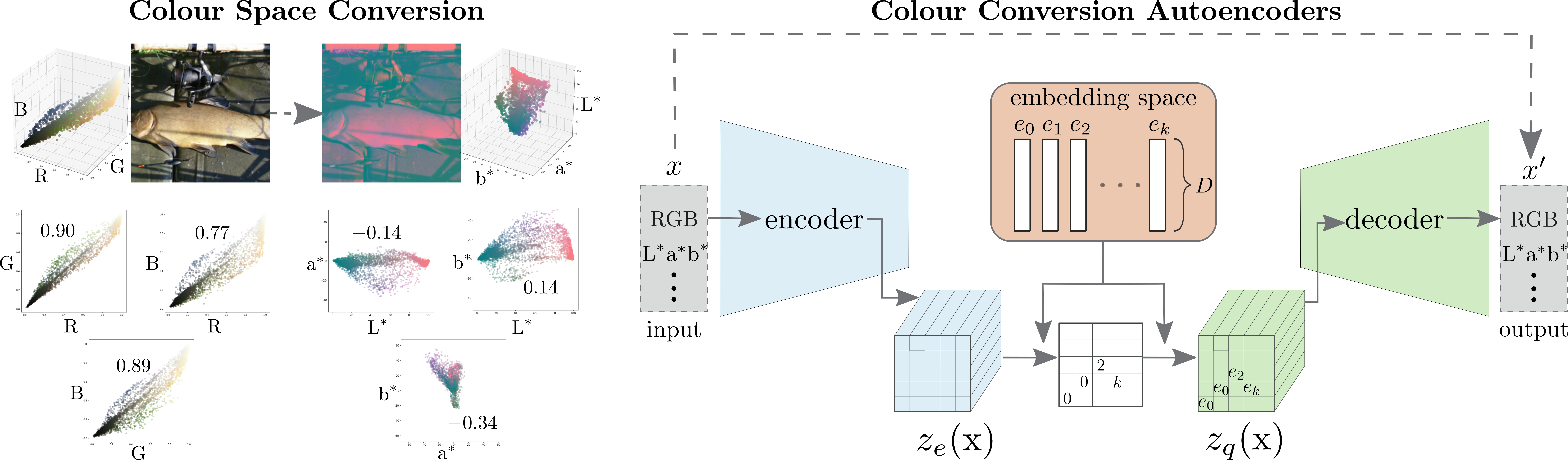}
    \caption{On the left: an exemplary conversion from RGB to CIE L*a*b* colour space. Each image is shown together with its 3D representation of their colours in different axis. On the right: the representation of colour converting VQ-VAE.}
    \label{fig:flow}
\end{figure}

We focused on Vector Quantised Variational Autoencoder (VQ-VAE) \cite{van2017neural}. These networks consist of three main blocks: 1) an encoder that processes the input data $x$ to $z_e(x)$; 2) a latent embedding space $\{e\} \in \mathbb{R}^{K \times D}$, with $K$ vectors of dimensionality $D$, that maps $z_e(x)$ onto $z_q(x)$ by estimating the nearest vector $e_i$ to $z_e(x)$; 3) a decoder that reconstructs the final output $x'$ with a distribution $p(x|z_q(x))$ over the input data (see the right panel in Figure~\ref{fig:flow}). The loss function is defined as follows,
\begin{equation}\label{loss}
    L = \log p(x|z_q(x)) + \|sg[z_e(x)] - e \|_{2}^{2} + \beta \| z_e(x) - sg[e]\|_{2}^{2},
\end{equation}
where $sg$ denotes the stop gradient computation that is defined as the identity during the forward-propagation, and with zero partial derivatives during the back-propagation to refrain its update. The first term in Eq.~\ref{loss} corresponds to the reconstruction loss incorporating both encoder and decoder; the second term updates the embedding vectors; and the third term, referred to as the commitment loss, harmonising the encoder and embedding vectors. The parameter $\beta \in \mathbb{R}$ is set to $0.5$ in all our experiments. 


\subsection{Colour Spaces}

We studied five three-dimensional colour spaces: RGB, LMS, CIE L*a*b*, DKL and HSV. The RGB colour space is a cubic representation of colours by three additive primaries. RGB is the standard in electronic imaging. The LMS colour space corresponds to the response of human cones (long-, middle-, and short-wavelengths) \cite{Gegenfurtner&Sharpe1999}. The CIE L*a*b* colour space (lightness, red-green and yellow-blue axes) is designed to approximately capture equal perceptual changes \cite{CIELAB}. The DKL colour space (Derrington-Krauskopf-Lennie \cite{derrington1984chromatic}) models the colour opponent responses of rhesus monkeys in the early visual system. The HSV colour space (hue, saturation, value) is a cylindrical representation of RGB cube designed by computer graphics. The intra-axes correlation for RGB and LMS is very high for natural images, hence referred to as \textit{correlated} colour spaces. On the contrary, intra-axes correlations for CIE L*a*b* and DKL are very low, hence referred to as \textit{decorrelated} colour spaces.\footnote{We computed these correlations $r$ in all images of ImageNet dataset (hundred-random pixels per image). RGB: $r^{RG}=0.90$, $r^{RB}=0.77$, $r^{GB}=0.89$; LMS: $r^{LM}=1.00$, $r^{LS}=0.93$, $r^{MS}=0.93$;  L*a*b*: $r^{L*a*}=-0.14$, $r^{L*b*}=0.13$, $r^{a*b*}=-0.34$; DKL: $r^{DK}=0.01$, $r^{DL}=0.14$, $r^{KL}=0.61$.}

The input-output to our networks can be in any combination of these colour spaces. Effectively, our VQ-VAE models, in addition to learning efficient representation, must learn the transformation function from their input to output colour space. It is worth considering that the original images in explored datasets are in the RGB format. Therefore, one might expect a slight positive bias towards this colour space given its gamut defines the limits of other colour spaces.




\section{Experiments}

We trained several instances of VQ-VAEs with distinct sizes of embedding space $\{e\} \in \mathbb{R}^{K \times D}$. The training procedure was identical for all networks: trained with Adam optimiser \cite{kingma2014adam} ($lr=2\times10^{-4}$) for 90 epochs. We used ImageNet dataset \cite{deng2009imagenet} for training. This is a visual database of object recognition in real-world images, divided into one thousand categories. The training set contains 1.3 million images. At every epoch, we exposed the network to 100K images of size $ 224\times 224$ of three colour channels.

To increase the generalisation power of our findings, we evaluated all networks on the validation-set of three benchmark datasets: ImageNet (50K images), COCO (5K images), and CelebA (\texttildelow 20K images). COCO is a large-scale object detection and segmentation dataset \cite{lin2014microsoft}. CelebA (CelebFaces Attributes Dataset) \cite{liu2015deep} contains facial attributes of celebrities. We relied on two classes of evaluation\footnote{For reproduction, the source code and all experimental data are available in our GitHub: \url{https://github.com/ArashAkbarinia/DecomposeNet} .}: low-level \cite{theis2015note}, to account for the model's reliability in capturing the local statistics of an image; high-level \cite{borji2019pros}, to assess the model's capacity in reproducing the global content of an image.

\begin{figure}[h]
    \centering
    \begin{tabular}{@{\hskip0pt}c@{\hskip2pt}c@{\hskip2pt}c@{\hskip0pt}}
         ImageNet & COCO & CelebA \\
         \includegraphics[width=0.38\linewidth]{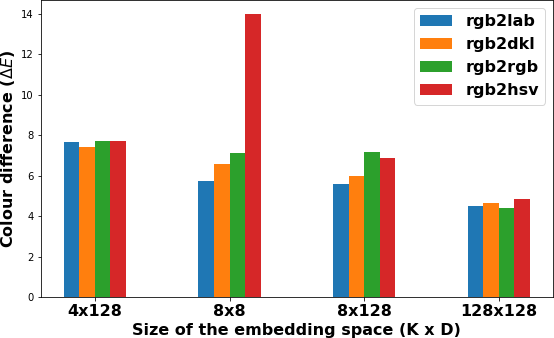} & \includegraphics[width=0.3\linewidth]{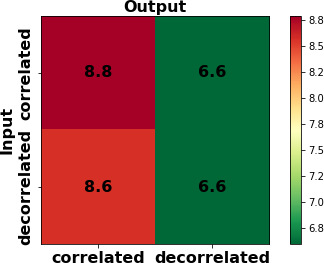} & \includegraphics[width=0.3\linewidth]{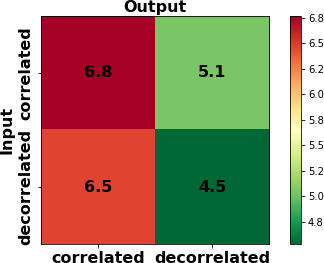} \\
    \end{tabular}
    \caption{The colour difference (CIE $\Delta E$-2000) between the reconstructed and the original image (the lower the better). Left: results of various sizes of embedding space. Middle and right: pairwise comparison of two groups of input-output colour spaces for the network with $\{e\} \in \mathbb{R}^{8 \times 128}$.}
    \label{fig:coco_imagenet_deltae}
\end{figure}

\textbf{Low-level evaluation} -- We used the standard colour difference metric CIE $\Delta E$-2000 \cite{sharma2005ciede2000} to measure the pixel-wise performance of networks. The summary of this evaluation is reported in Figure \ref{fig:coco_imagenet_deltae}. The left panel compares the performance across embedding spaces for RGB-input networks. The high $\Delta E$ of \textit{rgb2hsv} pops up at low-dimensionality of the embedding vector ($D=8$) that might be due to the circular nature of hue. For the smallest and the largest embedding space, we observe no significant differences between the four networks. However, for embedding spaces of $8 \times 8$ and $8 \times 128$ an advantage appears for networks whose outputs are opponent colour spaces (DKL and CIE L*a*b).

The middle and right panels of Figure \ref{fig:coco_imagenet_deltae} groups the results into correlated (RGB and LMS) versus decorrelated (DKL and CIE L*a*b) colour spaces for both inputs and outputs. HSV is excluded in these analysis due to the aforementioned reason. There is a clear tendency of lower $\Delta E$ values for networks whose output colour space is decorrelated. Compare the right columns of each confusion matrix. Overall, the decorrelating networks have an advantage of 2 JND (just-noticeable difference).

\begin{figure}[h]
    \centering
    \begin{tabular}{@{\hskip0pt}cc@{\hskip0pt}}
         ImageNet & COCO \\
         \includegraphics[width=0.47\linewidth]{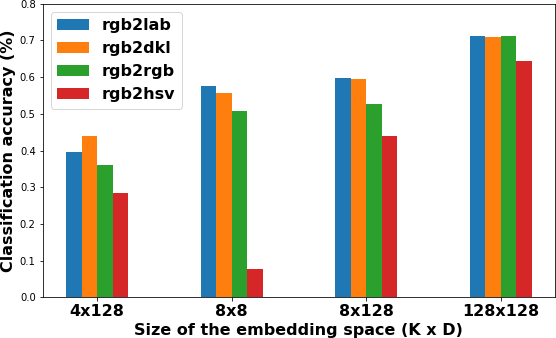} & \includegraphics[width=0.47\linewidth]{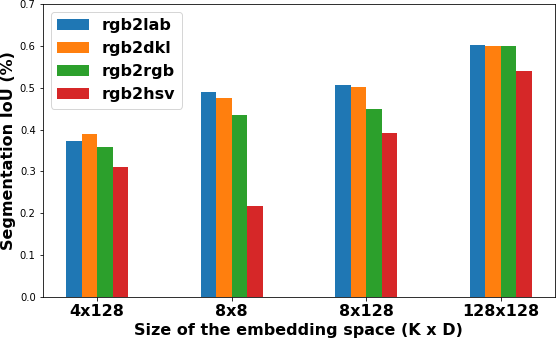} \\
    \end{tabular}
    \caption{High-level visual task evaluation. Left: ResNet50's classification accuracy on reconstructed images of ImageNet. Right: FPNS's segmentation IoU on reconstructed images of COCO.}
    \label{fig:coco_imagenet_acc}
\end{figure}

\textbf{High-level evaluation} -- Although informative, pixel-wise measures are unable to capture the global content of an image and whether semantic information remains perceptually intact. To account for this limitation, the utility of reconstructed images in a high-level visual task must be measured. We performed a procedure similar to the standard Inception Score \cite{salimans2016improved, borji2019pros}. We fed the reconstructed images to two other networks performing the task of object classification (ResNet50 \cite{he2016deep}) and scene segmentation (Feature Pyramid Network---FPN \cite{kirillov2019panoptic}). We simply used the pretrained-models of ResNet50 and FPN without any fine-tuning. The evaluation for ResNet50 is the classification accuracy on ImageNet dataset. The evaluation for FPN is the intersection over union (IoU) on COCO dataset. The results for various sizes of embedding space are reported in Figure \ref{fig:coco_imagenet_acc}.

The overall trend is much alike for both high-level tasks. The lowest performance occurs for \textit{rgb2hsv} across all embedding spaces. Networks whose output is an opponent colour space systematically perform better than \textit{rgb2rgb}, with an exception for the largest embedding space ($128 \times 128$). The comparison of the smallest embedding space ($4 \times 128$) across Figures \ref{fig:coco_imagenet_deltae} and \ref{fig:coco_imagenet_acc} demonstrates the importance of high-level evaluation. Although no difference emerges for the measure of $\Delta E$, the classification and segmentation metrics are substantially influenced by the quality of the reconstructed images in those four VQ-VAEs.

\begin{figure}[h]
    \centering \renewcommand{\arraystretch}{1.5}
    \begin{tabular}{@{\hskip0pt}c@{\hskip2pt}c@{\hskip2pt}c@{\hskip2pt}c@{\hskip0pt}}
        \multicolumn{2}{c}{ImageNet} & \multicolumn{2}{c}{COCO} \\
         $\{e\} \in \mathbb{R}^{8 \times 8}$ & $\{e\} \in \mathbb{R}^{8 \times 128}$ & $\{e\} \in \mathbb{R}^{8 \times 8}$ & $\{e\} \in \mathbb{R}^{8 \times 128}$ \\ 
         \includegraphics[width=0.23\linewidth]{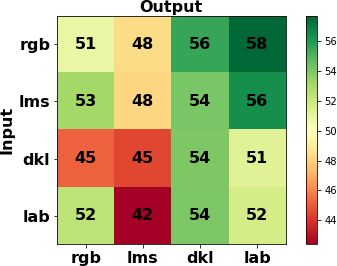} & \includegraphics[width=0.23\linewidth]{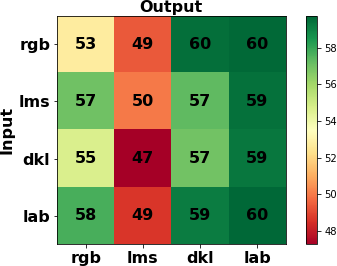} & \includegraphics[width=0.23\linewidth]{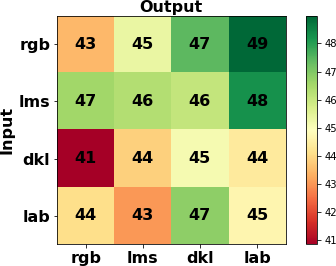} & \includegraphics[width=0.23\linewidth]{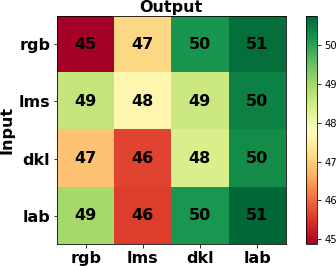} \\ 
         \includegraphics[width=0.23\linewidth]{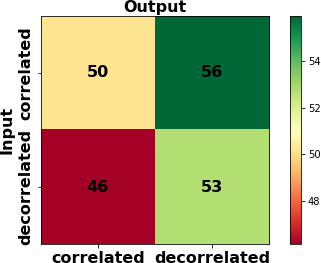} & \includegraphics[width=0.23\linewidth]{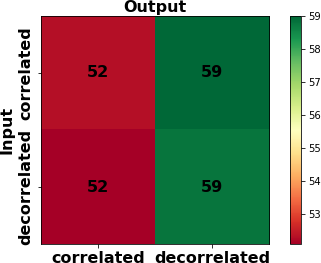} & \includegraphics[width=0.23\linewidth]{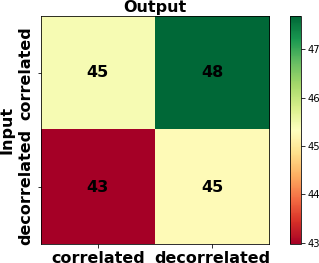} & \includegraphics[width=0.23\linewidth]{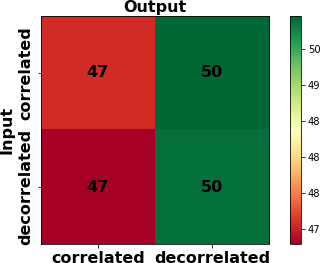} \\ 
    \end{tabular}
    \caption{Pairwise comparison of correlated (RGB and LMS) versus decorrelated (DKL and L*a*b) input-output colour spaces of VQ-VAEs. ImageNet metric: accuracy; COCO: IoU.}
    \label{fig:all_pairs_acc}
\end{figure}

To thoroughly examine the effect observed above, networks whose output is an opponent colour space perform better than the baseline \textit{rgb2rgb} models, we performed a pairwise comparison of all combinations of input and output colour spaces. The results are reported in Figure \ref{fig:all_pairs_acc}. The model executing the \textit{rgb2lab} colour conversion obtains the highest performance across both datasets and embedding space capacity (the top row). Overall, we observe a clear advantage for networks whose output is a decorrelated colour space (the bottom row). This suggests the neural information processing is optimised by opponency, in line with the \textit{efficient coding} theory \cite{Barlow1961, Zhaoping2006}.

\subsection{Explaining the Performance Advantage}

\begin{figure}[h]
    \centering
    \begin{tabular}{@{\hskip0pt}cc@{\hskip0pt}}
         ImageNet & COCO \\
         \includegraphics[width=0.47\linewidth]{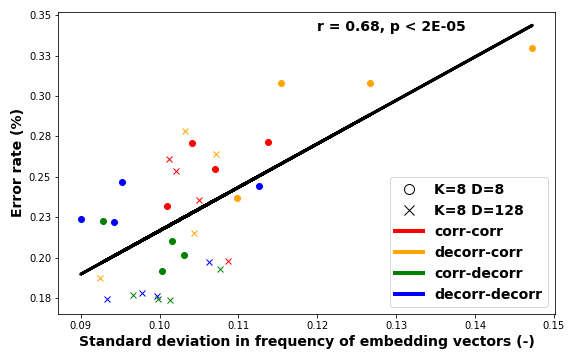} & \includegraphics[width=0.47\linewidth]{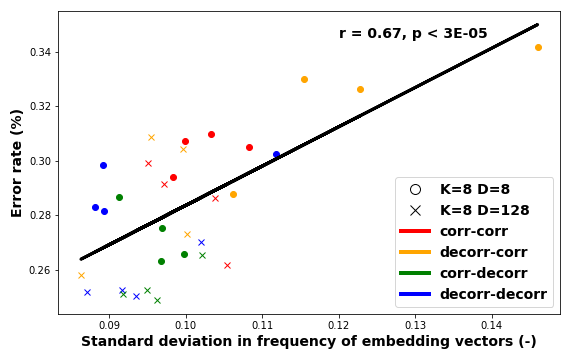} \\
    \end{tabular}
    \caption{Error rate as a function of the difference in frequency of selected vectors in the embedding space. A value of zero in the $x$-axis indicates all embedding vectors are equally used by the model. Higher values of $x$ indicate that the model relies heavily on certain vectors.}
    \label{fig:coco_imagenet_per}
\end{figure}

Networks whose output is a decorrelated colour space are more efficient. Understanding its cause is of great interest \cite{lillicrap2019does}. We hypothesised that the \textit{orthogonality} \cite{Buchsbaum&Gottschalk1983} of networks' embedding vectors would explain this effect. Nevertheless, no significant correlation emerged in this respect. We further explored the frequency of each embedding vector being used in the latent representation of images. Our hypothesis is: an efficient system distributes its encoding across all resources instead of heavily relying on a few components \cite{laughlin1981simple}. We measured this by computing the histogram of embedding vectors across all images ImageNet (50K) and COCO (5K). A zero standard deviation in frequency of selected vectors means an equal utilisation of all embedding vectors. In Figure \ref{fig:coco_imagenet_per}, we have plotted the error rate as a function of this measure. A significant correlation emerges in both datasets, suggesting a more uniform contribution of embedding vectors in VQ-VAEs enhances their visual information representation. This matches the neural model of histogram equalisation \cite{pratt2007digital} and is consistent with the \textit{efficient coding} theory for the biological visual system. In fact, to overcome the neurons restricted range of responses, \textit{efficient coding} could be achieved by encoding the input so that all response levels are used with equal frequency. If so, the encoded information is maximised because the information channel (i.e. the neuron) achieves its maximum entropy. Such coding can be achieved by computing responses from the cumulative probability function for the input distribution, so that equal output excursions correspond to equal probabilities of input. This mapping was found to take place in biological neural systems like neurons in the fly's compound eye \cite{laughlin1981simple}.

\section{Interpreting the Embedding Space}

\begin{figure}[h]
    \centering
    \includegraphics[width=1.\linewidth]{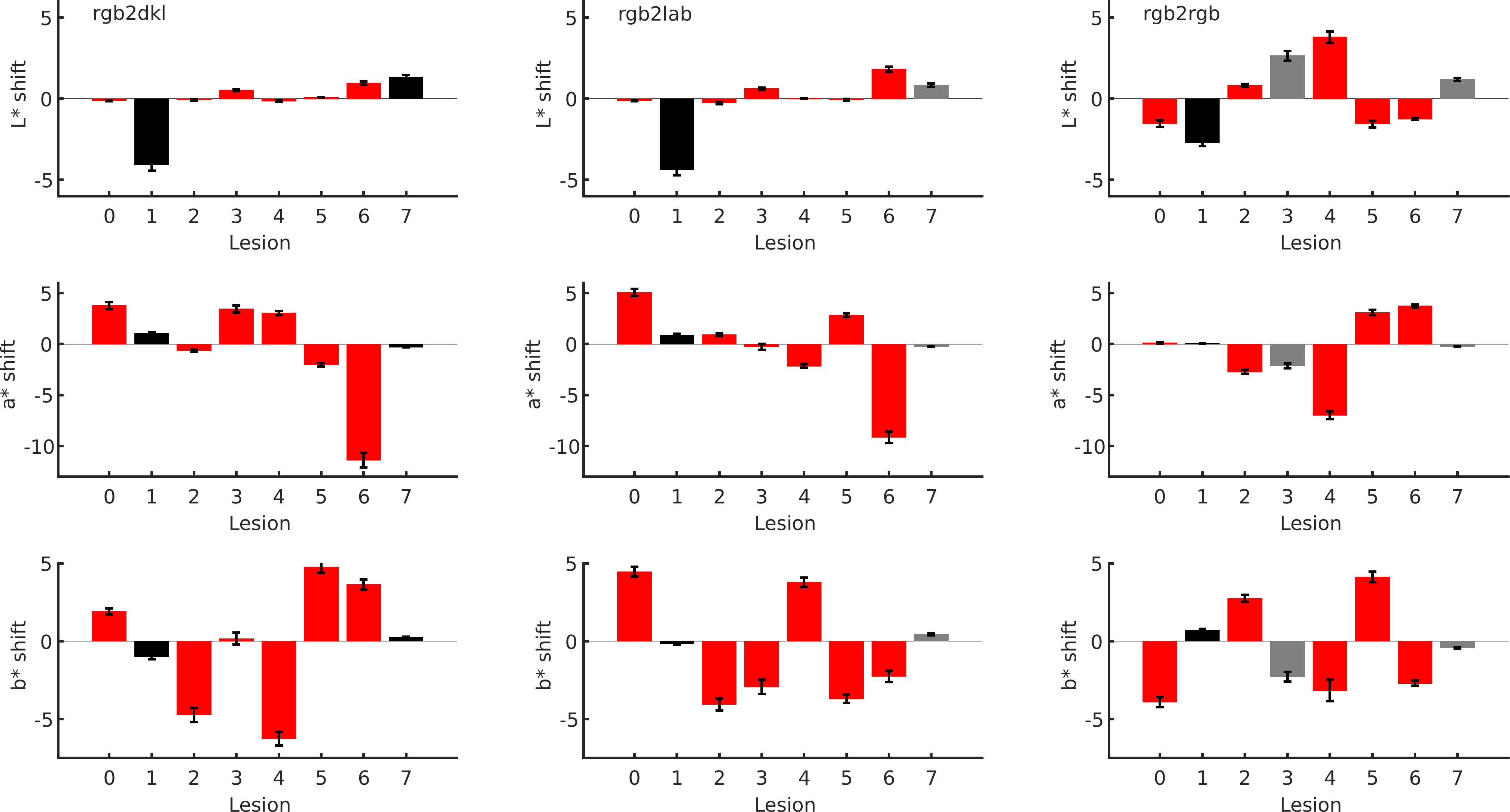}
    \caption{Average chromatic shifts in CIE L*a*b* colour space for each vector lesion. Numbers in the x-axis denote the indices of the embedding vectors. All networks are of $\{e\} \in \mathbb{R}^{8 \times 128}$. Grey bars indicate those vectors that predominantly impact the luminance channel.}
    \label{fig:hue_shift}
\end{figure}

Comprehension of the features learnt by a DNN remains a great challenge to the entire community \cite{lillicrap2019does}. Generative models and in particular variational autoencoders are no exceptions. Many works have focused on visual exploration of DNN's latent spaces. Interpolation or arithmetic operations on learnt latent features can reveal some meaningful and interpretable structure of these spaces \cite{radford2015unsupervised, bojanowski2017optimizing, kim2018efficient}. In practice, however, these approaches require explicit human supervision, a cumbersome task due to the often large dimensionality of the latent space. Here, we borrowed the ``lesion'' technique, commonly practised in the neuroscience community \cite{vaidya2019lesion}, and applied it to the embedding space by silencing one vector at a time (i.e. setting its weights to zero). A similar approach has revealed contrast kernels in DNNs \cite{akbarinia2020deciphering}. To measure the consequences of vectors' lesion, we quantified the chromatic shifts (in CIE L*a*b*) between the reconstructed image of full embedding space and lesion one. The differences computed for all pixels over a hundred random images from the COCO dataset are
illustrated in Figure \ref{fig:hue_shift}.

\begin{figure}[h]
    \centering
    \begin{tabular}{@{\hskip0pt}ccc@{\hskip0pt}}
          \includegraphics[width=0.3\linewidth]{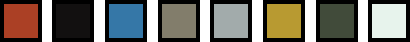} & \includegraphics[width=0.3\linewidth]{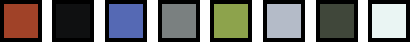} & \includegraphics[width=0.3\linewidth]{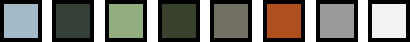} \\
         \textit{rgb2dkl} & \textit{rgb2lab} & \textit{rgb2rgb}
    \end{tabular}
    \caption{The reconstruction output by selecting a single vector of the entire embedding space. All models are of $\{e\} \in \mathbb{R}^{8 \times 128}$.}
    \label{fig:unique_hues}
\end{figure}

\begin{figure}
    \centering
    \begin{tabular}{@{\hskip0pt} p{0.2cm} @{\hskip1pt} >{\centering\arraybackslash}m{0.2cm} @{\hskip5pt} >{\centering\arraybackslash}m{.13\linewidth} @{\hskip5pt} >{\centering\arraybackslash}m{.13\linewidth} @{\hskip5pt} >{\centering\arraybackslash}m{.13\linewidth} @{\hskip8pt} p{0.2cm} @{\hskip5pt} >{\centering\arraybackslash}m{.47\linewidth} @{\hskip0pt}}
         & & \multicolumn{3}{c}{\textbf{Reconstructed Image}} & & \textbf{CIE L*a*b* Planes} \\
         \raisebox{3\height}{\textbf{A}} & \raisebox{0.5\height}{\rotatebox[origin=c]{90}{\scriptsize Full model}} & \includegraphics[width=\linewidth]{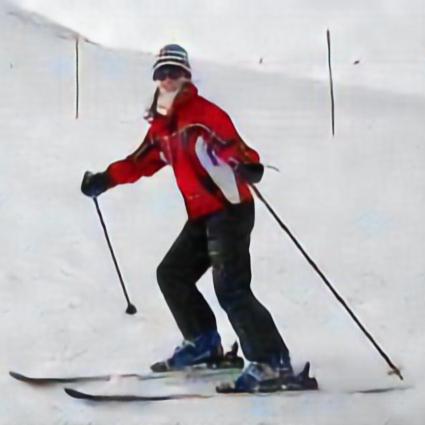} & \includegraphics[width=\linewidth]{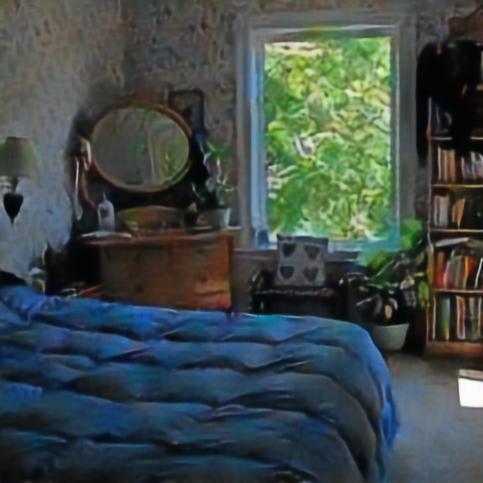} & \includegraphics[width=\linewidth]{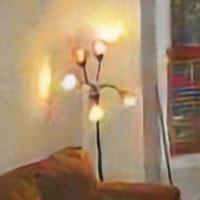} & \raisebox{3\height}{\textbf{E}} & \includegraphics[width=\linewidth]{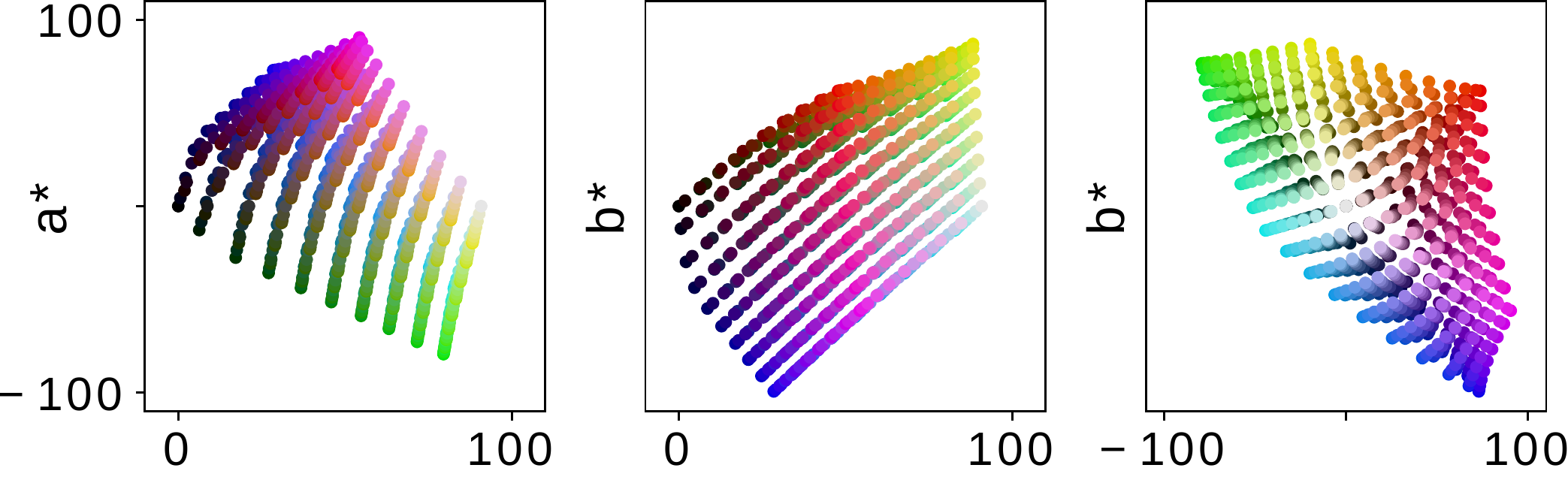} \\
         \raisebox{3\height}{\textbf{B}} & \raisebox{0.6\height}{\rotatebox[origin=c]{90}{\scriptsize Lesion $e_7$}} & \includegraphics[width=\linewidth]{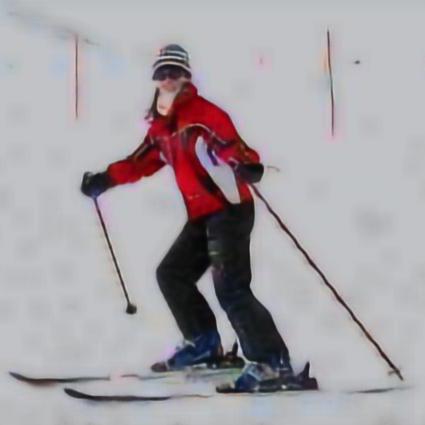} & \includegraphics[width=\linewidth]{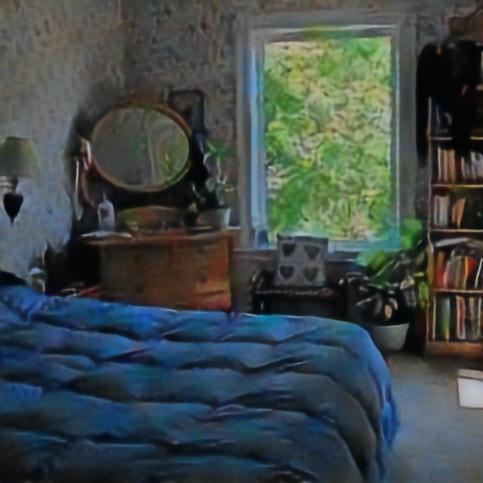} & \includegraphics[width=\linewidth]{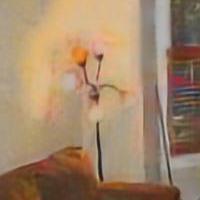} & \raisebox{3\height}{\textbf{F}} & \vspace{5pt}\begin{overpic}[width=\linewidth]{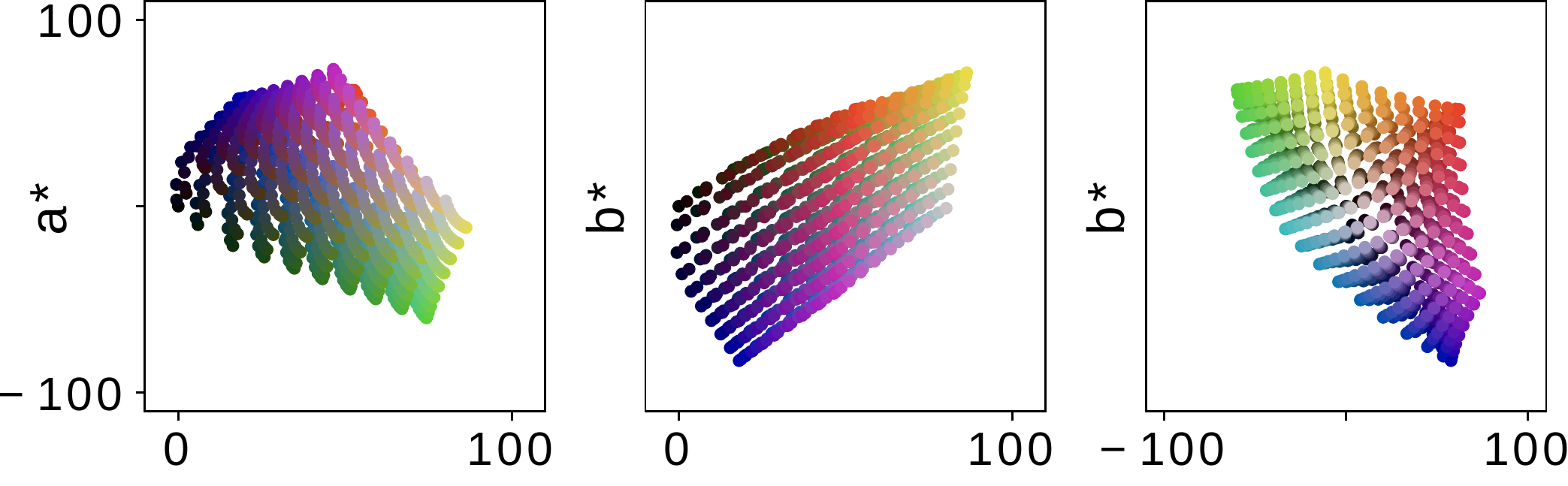}
         \put(32,32){{\parbox[c]{\linewidth}{\scriptsize{$R^2 = 0.99 \; SI = 0.09$}}}}
         \end{overpic} \\
         \raisebox{3\height}{\textbf{C}} & \raisebox{0.6\height}{\rotatebox[origin=c]{90}{\scriptsize Lesion $e_0$}} & \includegraphics[width=\linewidth]{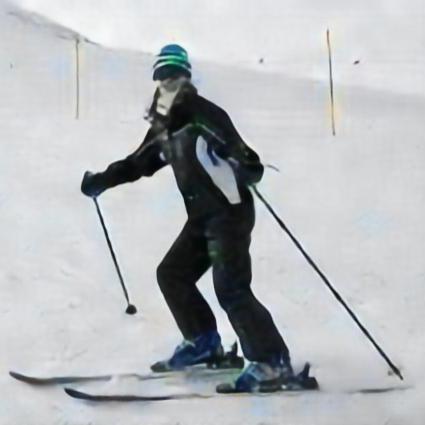} & \includegraphics[width=\linewidth]{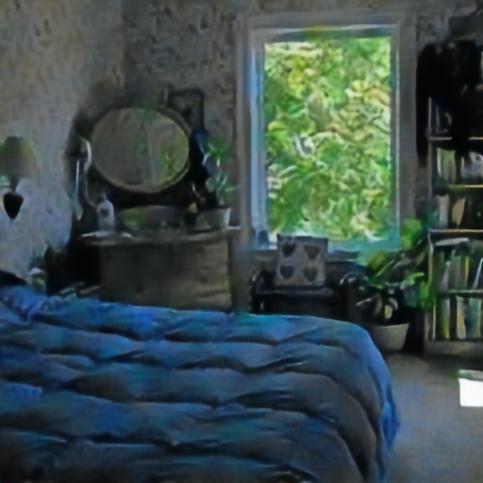} & \includegraphics[width=\linewidth]{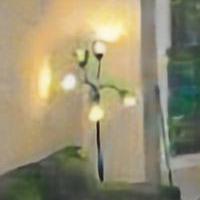} & \raisebox{3\height}{\textbf{G}} & \vspace{5pt}\begin{overpic}[width=\linewidth]{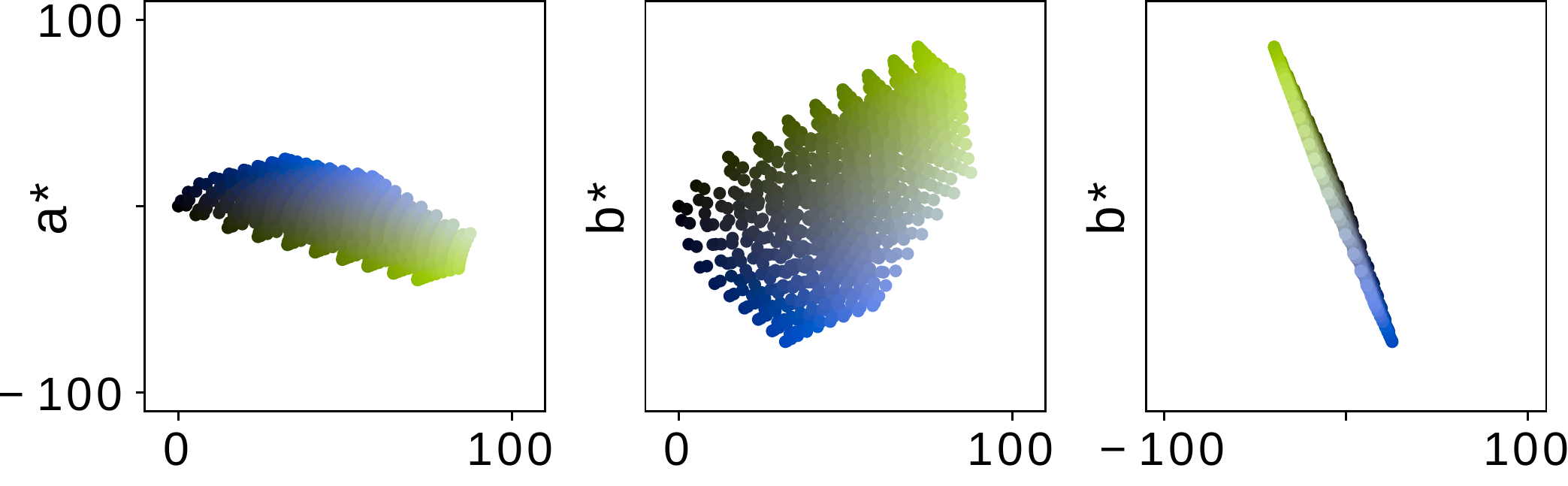}
         \put(32,32){{\parbox[c]{\linewidth}{\scriptsize{$R^2 = 0.99 \; SI = 0.99$}}}}
         \end{overpic} \\
         \raisebox{3\height}{\textbf{D}} & \raisebox{0.6\height}{\rotatebox[origin=c]{90}{\scriptsize Lesion $e_2$}} &  \includegraphics[width=\linewidth]{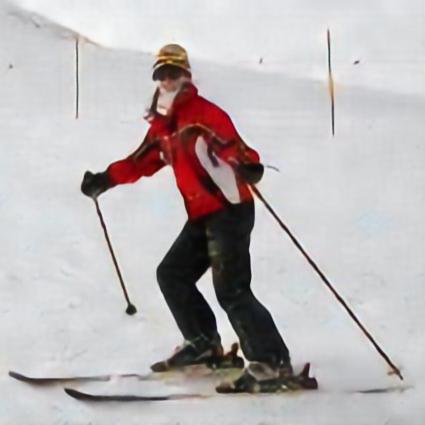} & \includegraphics[width=\linewidth]{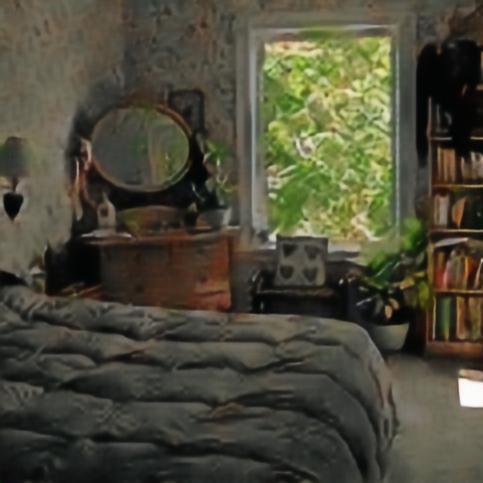} & \includegraphics[width=\linewidth]{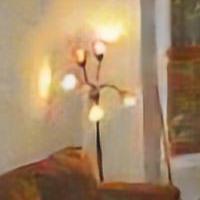} & \raisebox{3\height}{\textbf{H}} & \vspace{5pt}\begin{overpic}[width=\linewidth]{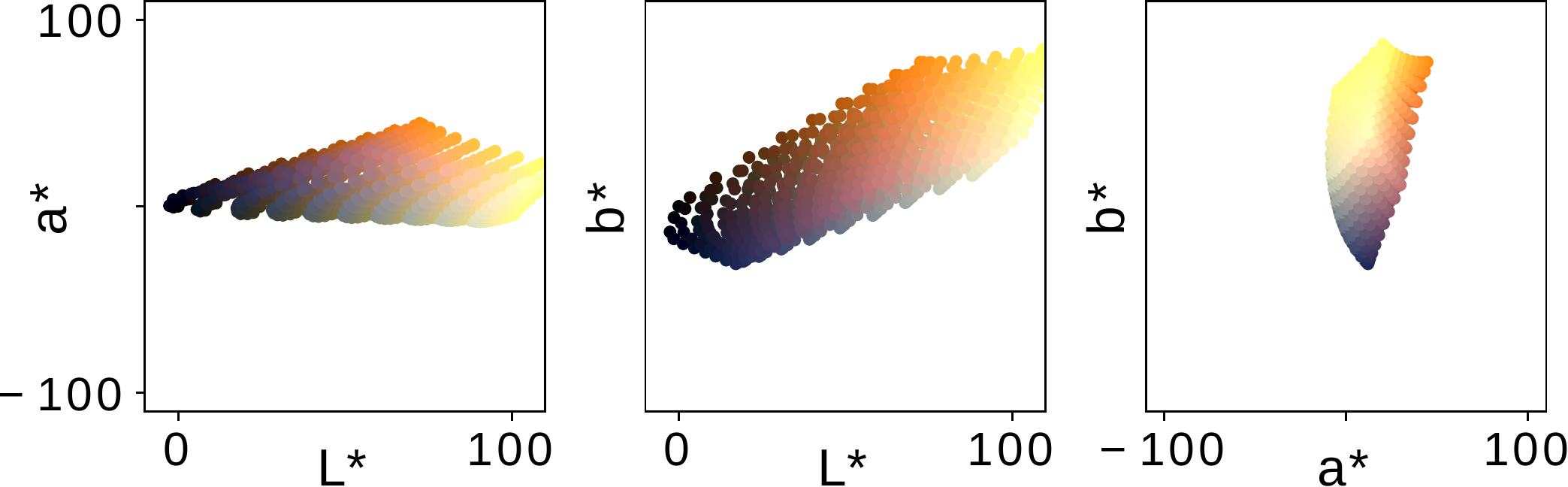}
         \put(32,32){{\parbox[c]{\linewidth}{\scriptsize{$R^2 = 0.99 \; SI = 0.93$}}}}
         \end{overpic} \\
    \end{tabular}
    \caption{Visualisation of the lesion effect for the \textit{rgb2dkl} VQ-VAE $\{e\} \in \mathbb{R}^{8 \times 128}$. On the left, images reconstructed by \textbf{A:} the full model; \textbf{B-D:} the lesion embedding space. On the right, scatter plots in CIE L*a*b* coordinates of \textbf{E:} the entire RGB cube; \textbf{F-H}: the RGB cube after applying the linear transformations modelled by each lesion.}
    \label{fig:lesion}
\end{figure}

This analysis provides an intuitive interpretation of some of the embedding vectors. For instance, the $e_0$ lesion in the \textit{rgb2dkl} network only impacts the chromatic axes. The direction of its shift implies pixels in the first quadrant of the chromaticity plane are effected (i.e. red pixels, see a*b* plane in panel E of Figure \ref{fig:lesion}). The $e_1$ vector of the same network appears to largely encode the luminance information. The $e_2$ vector predominantly influences the blue pixels (negative direction in the b* axis). Intrigued by this interpretation, we sampled from the embedding space an example where all pixels are of the same vector index. An RGB visualisation of this technique's output is illustrated in Figure \ref{fig:unique_hues}. Naturally, each vector encodes more than this simple reconstructed hues. However, the colour of these hues match the direction of shifts in Figure \ref{fig:hue_shift} and both together offer a perspective on colour representation in these VQ-VAEs.

To better conceptualise the effect of vector lesion, we have illustrated the \textit{rgb2dkl}'s reconstructions for three exemplary images in Figure \ref{fig:lesion}. Panel A corresponds to the full embedding space and panels B--D show examples of reconstructions with distinct vector lesions causing clear perceptual effects. In B, the lightness of very bright pixels is reduced (attend pixels outside the window and around light bulbs) while changes of chromaticity are unnoticeable. Contrary to this, in C \& D, colour is drastically modified with reddish and blueish pixels, respectively, turning achromatic. 

We hypothesised that the changes induced by a lesion, presented in Figure \ref{fig:hue_shift}, could be approximated by a linear transformation mapping the pixel distribution of the full reconstruction onto the lesion image. To compute these transformations, we used a multi-linear regression to find the best linear fit for the 1\% of pixels most affected with a lesion. The resulting $3\times3$ matrix is a linear transformation in the three-dimensional CIE L*a*b colour space (see the right side of Figure \ref{fig:lesion}). Lesions are excellently approximated by a linear transformation: on average accounting for 97\% of the total variance in the lesion effect (the lowest bound was 86\%). 

We diagonalised those resulting matrices to simplify their interpretations. 
The relative norms of the eigenvalues are indicative of the geometrical properties of the transformation matrices. For instance, the presence of at least one eigenvalue equal to zero specifies the extreme case of a singular matrix, corresponding to a linear transformation projecting a three-dimensional space onto lower dimensions. 
We quantified this by defining a singularity index \cite{philipona2006color}. Consider a transformation matrix $T$ approximating the lesion effect on the image colour distribution. Let $\lambda_1$, $\lambda_2$ and $\lambda_3$ be the three eigenvalues of $T$, such that $ \lVert \lambda_1 \rVert > \lVert \lambda_2 \rVert > \lVert \lambda_3 \rVert$. The singularity index is defined as: $SI = 1 - \frac{\lambda_3}{\lambda_1}$.

On the right side of Figure \ref{fig:lesion}, we have illustrated the result of applying these linear transformations in CIE L*a*b* coordinates. Panel E corresponds to the full RGB cube (essentially the L*a*b* planes limited by RGB gamut). In panels F--H the very same points are plotted transformed by the matrix modelled from the lesion of that vector.
This visualisation offers an intuitive interpretation of the embedding space. In the images of the second row (panel B), contrast in bright pixels is reduced and colour is little modified. We can observe this in its corresponding L*a*b* planes (e.g. attend the a*b* plane in F where the overall chromaticity structure is retained). In C, red pixels turn grey also evident in its corresponding L*a*b* planes (panel F) where red coordinates are collapsed. These features qualitatively match well to those reported in Figures \ref{fig:unique_hues} and \ref{fig:hue_shift}.
The singularity index also adequately captures the essence of these transformations. On the one hand, the low value of $SI$ in F suggests the global shape of colour space is retained while its volume is reduced. On the other hand, high values of $SI$ in panels G and H indicate the near collapse of a dimension.

The average $SI$ is very similar for \textit{rgb2dkl}, \textit{rgb2lab} and \textit{rgb2rgb}, 0.75, 0.73 and 0.72 respectively. 
Threshholding these indices distinguishes two family of transformations matrices $T$. The \textit{singular} group with $SI \ge 0.90$ indicating a lesion impact of \textit{at least} ten times larger along one dimension of the colour space. The  \textit{homogeneous} group with $SI \le 0.10$ indicating a lesion of uniform impact across all colour dimension with modulation in their contrast. Following these criteria, three of eight transformation matrices classify as \textit{singular} in all three colour converting networks. We observed only one \textit{homogeneous} matrix in the \textit{rgb2dkl} network (i.e. $e_7$ depicted in the second row of Figure \ref{fig:lesion}).

To summarise, performing the lesion technique on vectors of the VQ-VAE's embedding space presents interesting insights. Colour changes induced by these lesions can be faithfully modelled with parsimonious linear transformations. Studying these simple matrices offers an explicit understanding of lesions effect. In turn, this can shed light on colour representation in VQ-VAE models and open a promising line of investigation to decipher how visual information is encoded in generative models.

\section{Conclusion}

In this article, we introduced a simple quasi-unsupervised task---\textit{colour conversion}--for variational autoencoders. The comparison across several colour spaces suggests a network with a decorrelated colour space as its output exhibits an advantage in terms of capturing local and global features. We offered an explanation for this within the framework of \textit{efficient coding} \cite{Barlow1961, Zhaoping2006} and \textit{histogram equalisation} \cite{pratt2007digital}. These principles are not specific to autoencoders and could be applicable to a larger family of deep learning models.
We proposed a set of methods to interpret the embedding space of VQ-VAEs. Many of the constituent vectors manifest a clear effect along one colour direction.

We showed the vectors' influence could be modelled by \textit{parsimonious linear transformations}. This is a powerful tool with great potentials to enhance the interpretability of latent spaces in general.


\section*{Acknowledgement}
This study was funded by Deutsche Forschungsgemeinschaft SFB/TRR 135.

\bibliographystyle{plain}
\bibliography{references}

\appendix

\section{Qualitative Comparison}

\begin{figure}[h]
    \centering \renewcommand{\arraystretch}{1.5}
    \begin{tabular}{@{\hskip0pt} >{\centering\arraybackslash}m{.22\linewidth} @{\hskip8pt} >{\centering\arraybackslash}m{.22\linewidth} @{\hskip8pt} >{\centering\arraybackslash}m{.22\linewidth} @{\hskip8pt} >{\centering\arraybackslash}m{.22\linewidth} @{\hskip0pt}}
        Original & rgb2rgb & rgb2dkl & rgb2lab \\
        \includegraphics[width=\linewidth]{000000000139_32.jpg} & \includegraphics[width=\linewidth]{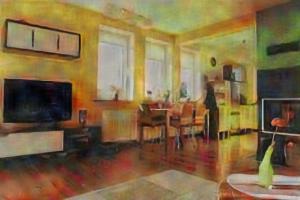} & \includegraphics[width=\linewidth]{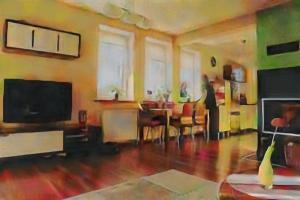} & \includegraphics[width=\linewidth]{lab_ful_01111111_000000000139.jpg} \\ \includegraphics[width=\linewidth]{000000002157_32.jpg} & \includegraphics[width=\linewidth]{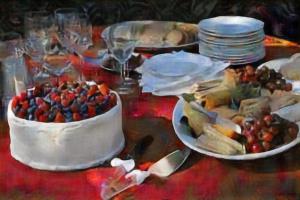} & \includegraphics[width=\linewidth]{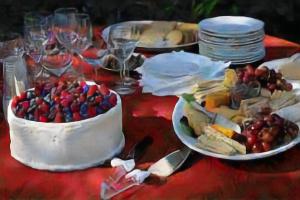} & \includegraphics[width=\linewidth]{lab_ful_01111111_000000002157.jpg} \\ \includegraphics[width=\linewidth]{000000007784_32.jpg} & \includegraphics[width=\linewidth]{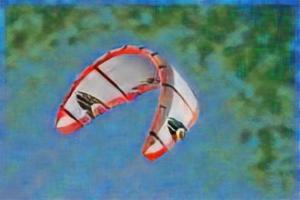} & \includegraphics[width=\linewidth]{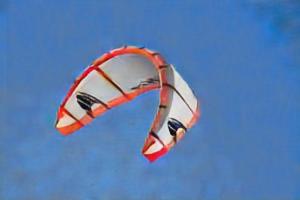} & \includegraphics[width=\linewidth]{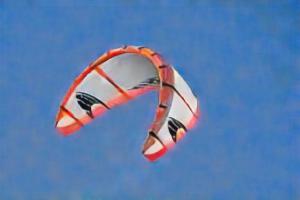} \\ \includegraphics[width=\linewidth]{000000577182_32.jpg} &  \includegraphics[width=\linewidth]{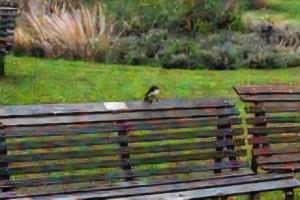} &  \includegraphics[width=\linewidth]{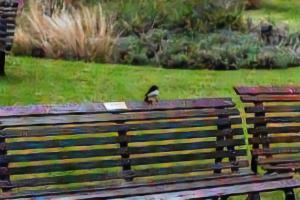} & \includegraphics[width=\linewidth]{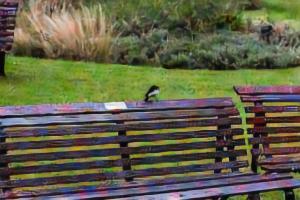} \\
    \end{tabular}
    \caption{Qualitative comparison of three VQ-VAEs of \textbf{K=8} and \textbf{D=128}.}
    \label{fig:examples}
\end{figure}

Along with the quantitative evaluations reported in the manuscript, the benefits of utilising a decorrelated colour space for the network's output can be appreciated qualitatively (see Figure \ref{fig:examples}). These are representative samples from the COCO dataset \cite{lin2014microsoft}. The Jupyter-Notebook scripts our GitHub provide more examples\footnote{The weights of all trained networks and image outputs of lesion study are publicly available for interested readers in our GitHub: \url{https://github.com/ArashAkbarinia/DecomposeNet} .}. Overall, the \textit{rgb2dkl} and \textit{rgb2lab} VQ-VAEs generate more coherent images. For instance, in the first row of Figure \ref{fig:examples}, the \textit{rgb2rgb} output contains a large amount of artefacts on walls and ceilings. In contrast, the output of \textit{rgb2dkl} and \textit{rgb2lab} are sharper.

\section{Quantitative Comparison}

The qualitative intuitions from the pictures above match very well the computed quantitative metrics. 

\subsection{Low-level Pixel-wise}

\begin{figure}[h]
    \centering
    \begin{tabular}{@{\hskip0pt}cc@{\hskip0pt}}
         ImageNet & COCO \\
         \includegraphics[width=0.47\linewidth]{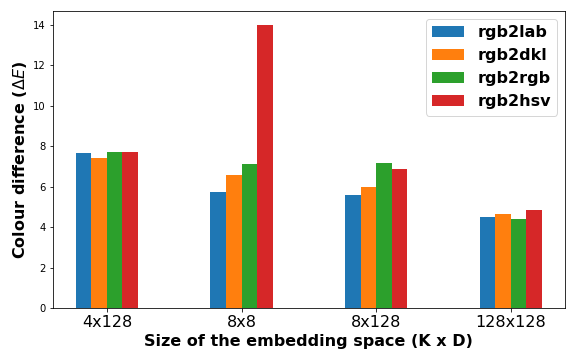} & \includegraphics[width=0.47\linewidth]{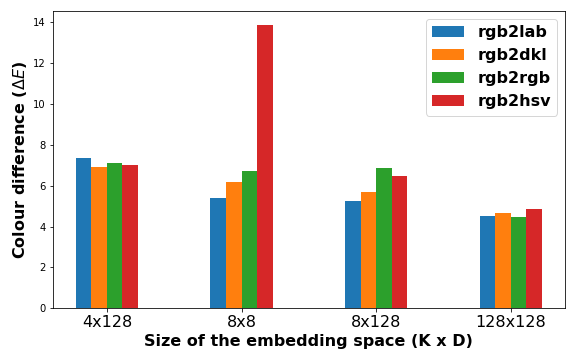} \\
    \end{tabular}
    \caption{The colour difference (CIE $\Delta E$-2000) between the reconstructed and the original image (the lower the better).}
    \label{fig:coco_imagenet_deltae_sup}
\end{figure}

Figure \ref{fig:coco_imagenet_deltae_sup} compares the colour difference metric of four colours converting VQ-VAEs with various sizes of embedding space. The results for ImageNet and COCO are very similar. The benefit of using DKL or CIE L*a*b* colour spaces for the network's output appears when the embedding space contains eight vectors (i.e. $8\times 8$ and $8 \times 128$). In high-level visual task evaluation (Figure \ref{fig:coco_imagenet_acc_sup}) the benefit of these two colour spaces is also evident for the $4 \times 128$ network. This effect disappears for the largest networks ($128 \times 128$). Perhaps, the benefit of decorrelated colour space vanishes with abundant of computation resources. To resolve this, we propose a psychophysical experiment in which participants must judge the quality of generated images for these networks.

\begin{figure}[h]
    \centering \renewcommand{\arraystretch}{1.5}
    \begin{tabular}{@{\hskip0pt}c@{\hskip2pt}c@{\hskip2pt}c@{\hskip0pt}}
         ImageNet & COCO & CelebA \\
         \includegraphics[width=0.31\linewidth]{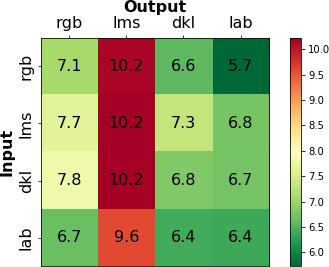} & \includegraphics[width=0.31\linewidth]{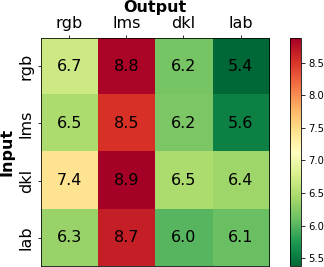} & \includegraphics[width=0.31\linewidth]{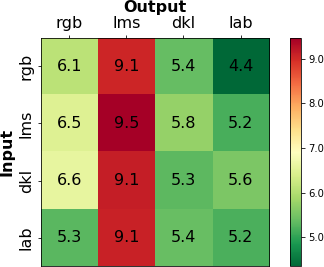} \\
         \includegraphics[width=0.31\linewidth]{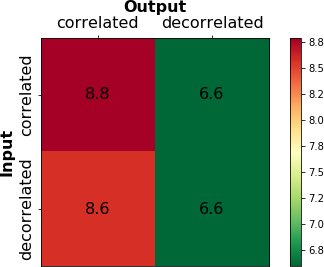} & \includegraphics[width=0.31\linewidth]{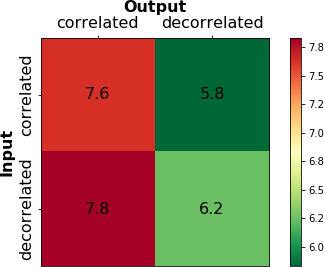} & \includegraphics[width=0.31\linewidth]{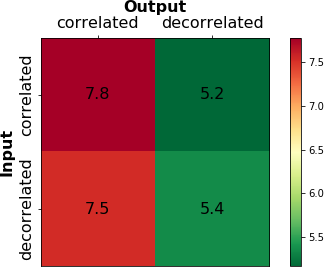} \\
    \end{tabular}
    \caption{CIE $\Delta E$-2000 pairwise comparison of correlated (RGB and LMS) versus decorrelated (DKL and L*a*b) input-output colour spaces for VQ-VAEs of \textbf{K=8} and \textbf{D=8}.}
    \label{fig:all_pairs_des_k8d8}
\end{figure}

\begin{figure}[h]
    \centering \renewcommand{\arraystretch}{1.5}
    \begin{tabular}{@{\hskip0pt}c@{\hskip2pt}c@{\hskip2pt}c@{\hskip0pt}}
         ImageNet & COCO & CelebA \\
         \includegraphics[width=0.31\linewidth]{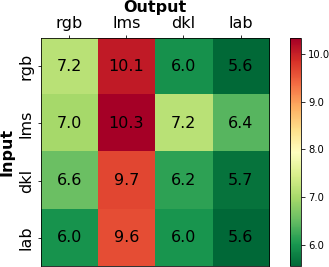} & \includegraphics[width=0.31\linewidth]{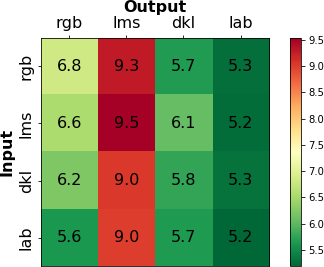} & \includegraphics[width=0.31\linewidth]{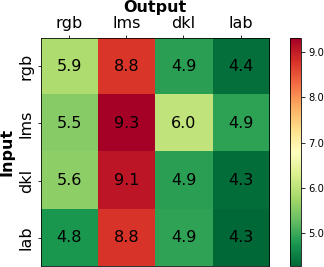} \\
         \includegraphics[width=0.31\linewidth]{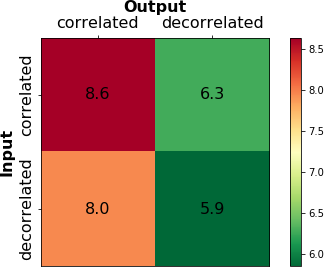} & \includegraphics[width=0.31\linewidth]{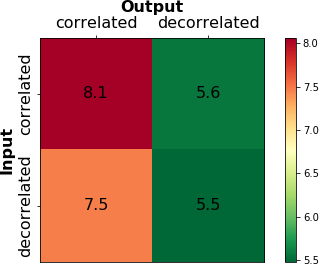} & \includegraphics[width=0.31\linewidth]{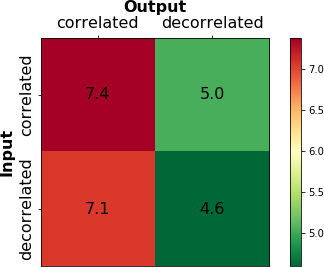} \\
    \end{tabular}
    \caption{CIE $\Delta E$-2000 pairwise comparison of correlated (RGB and LMS) versus decorrelated (DKL and L*a*b) input-output colour spaces for VQ-VAEs of \textbf{K=8} and \textbf{D=128}.}
    \label{fig:all_pairs_des_k8d128}
\end{figure}

The first rows of Figures \ref{fig:all_pairs_des_k8d8} and \ref{fig:all_pairs_des_k8d128} illustrate a pair-wise comparison of conversion among four colour spaces and the second rows group them into correlated and decorrelated colour spaces. Networks whose outputs are a decorrelated colour space obtain lower colour difference (more than $2 \Delta E$).

\subsection{High-level Visual Tasks}

\begin{figure}[h]
    \centering
    \begin{tabular}{@{\hskip0pt}cc@{\hskip0pt}}
         ImageNet & COCO \\
         \includegraphics[width=0.47\linewidth]{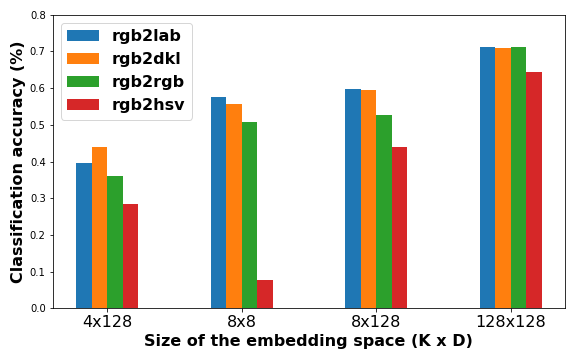} & \includegraphics[width=0.47\linewidth]{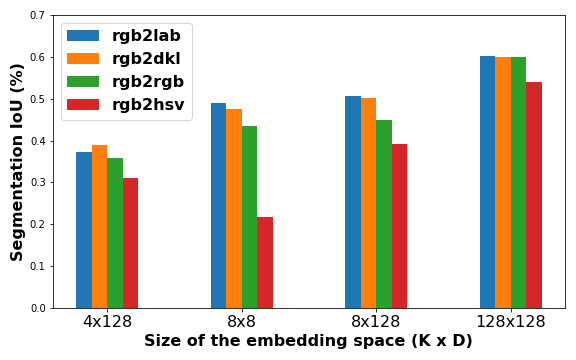} \\
    \end{tabular}
    \caption{High-level visual task evaluation. Left: ResNet50's classification accuracy on reconstructed images of ImageNet. Right: FPNS's segmentation IoU on reconstructed images of COCO.}
    \label{fig:coco_imagenet_acc_sup}
\end{figure}

Figure \ref{fig:coco_imagenet_acc_sup} compares high-level metrics of four colour converting VQ-VAEs with various sizes of embedding space. The results for ImageNet and COCO are very similar. The benefit of using DKL or CIE L*a*b* colour spaces for the network's output appears in all except for the largest embedding space ($128\times 128$). This suggests that colour opponency might be related to the scarcity of the computational resources a visual system faces, in line with \textit{efficient coding} \cite{Barlow1961, Zhaoping2006}.

\begin{figure}[h]
    \centering \renewcommand{\arraystretch}{1.5}
    \begin{tabular}{@{\hskip0pt}c@{\hskip2pt}c@{\hskip2pt}c@{\hskip2pt}c@{\hskip0pt}}
        \multicolumn{2}{c}{ImageNet} & \multicolumn{2}{c}{COCO} \\
         $\{e\} \in \mathbb{R}^{8 \times 8}$ & $\{e\} \in \mathbb{R}^{8 \times 128}$ & $\{e\} \in \mathbb{R}^{8 \times 8}$ & $\{e\} \in \mathbb{R}^{8 \times 128}$ \\ 
         \includegraphics[width=0.23\linewidth]{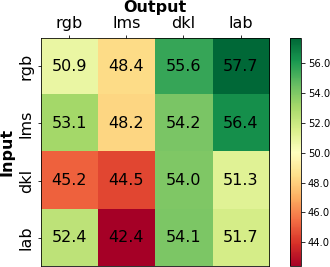} & \includegraphics[width=0.23\linewidth]{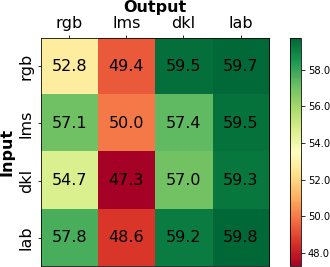} &    \includegraphics[width=0.23\linewidth]{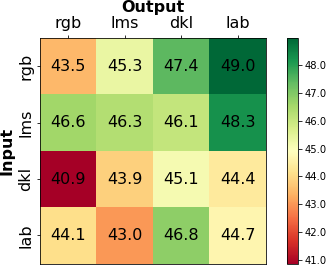} & \includegraphics[width=0.23\linewidth]{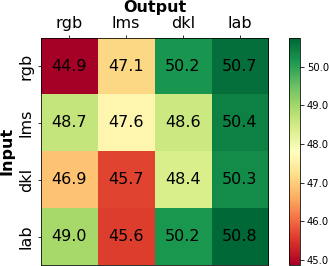}  \\
         \includegraphics[width=0.23\linewidth]{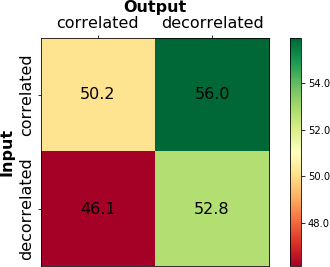} & \includegraphics[width=0.23\linewidth]{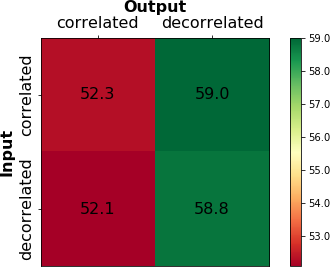} &    \includegraphics[width=0.23\linewidth]{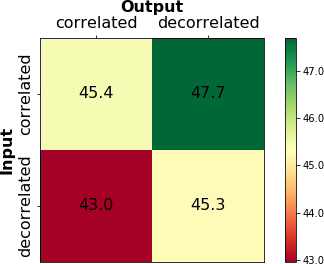} & \includegraphics[width=0.23\linewidth]{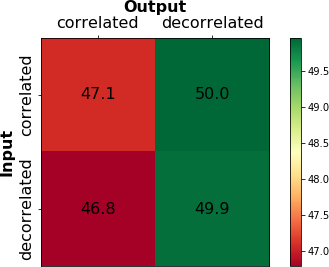}  \\
    \end{tabular}
    \caption{Pairwise comparison of correlated (RGB and LMS) versus decorrelated (DKL and L*a*b) input-output colour spaces of VQ-VAEs. Metric is accuracy for ImageNet and IoU for COCO.}
    \label{fig:all_pairs_acc_sup}
\end{figure}

The first rows of Figure \ref{fig:all_pairs_acc_sup} illustrates a pair-wise comparison of conversion among four colour spaces and the second row group them into correlated and decorrelated colour spaces. Networks whose outputs are a decorrelated colour space on average obtain higher classification accuracy and segmentation IoU (about 5\%). Within this group the \textit{rgb2lab} tends to systematically achieves higher performance in comparison to the \textit{rgb2dkl}. Although the difference is quite small this is interesting to explore the reason behind. Perhaps the nonlinear operation in the luminance channel of CIE L*a*b* is the root cause of this effect.

\section{Interpreting the Embedding Space}

In order to decipher the role of each embedding vector we explored and developed a set of techniques.

\subsection{Embedding Vector's Hue}

In order to visualise the information each vector encodes, we sampled from the embedding space an example where all pixels are of the same vector index. In particular, we report the results for $\{e\} \in \mathbb{R}^{K \times D}$ with $K=8$ and $D = 8, 128$; in all the 16 possible combinations with input/output \{\textit{rgb}, \textit{lms}, \textit{dkl}, \textit{lab}\}. 

\begin{figure}[h]
    \centering \renewcommand{\arraystretch}{1.5}
    \begin{tabular}{@{\hskip0pt}c@{\hskip10pt}c@{\hskip10pt}c@{\hskip10pt}c@{\hskip0pt}}
        \textit{rgb2rgb} & \textit{rgb2lms} & \textit{rgb2dkl} & \textit{rgb2lab} \\
        \includegraphics[width=0.23\linewidth]{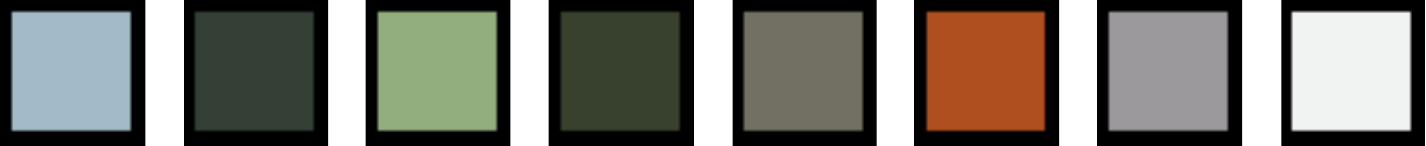} & \includegraphics[width=0.23\linewidth]{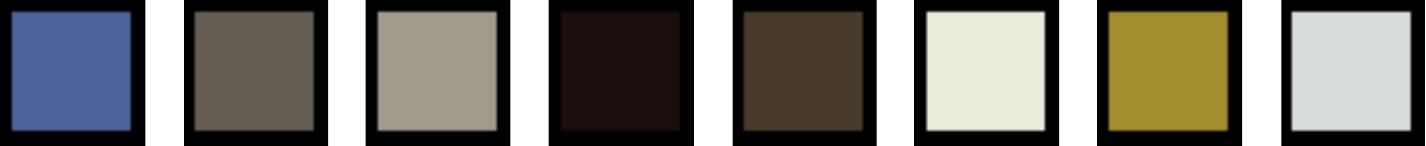} & \includegraphics[width=0.23\linewidth]{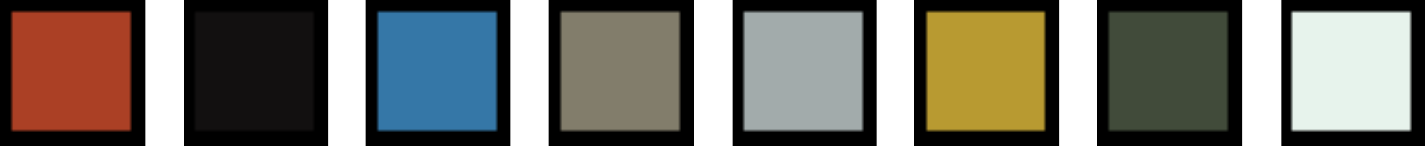} & \includegraphics[width=0.23\linewidth]{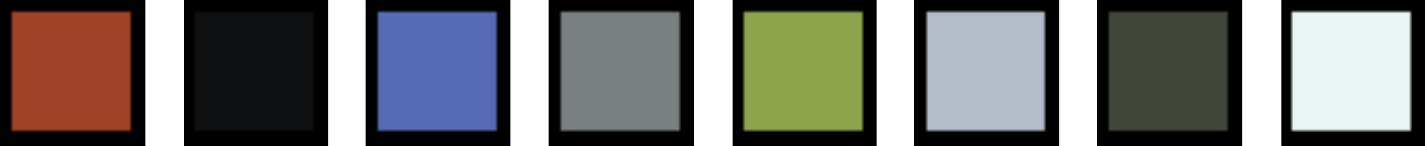} \\
        \textit{lms2rgb} & \textit{lms2lms} & \textit{lms2dkl} & \textit{lms2lab} \\
        \includegraphics[width=0.23\linewidth]{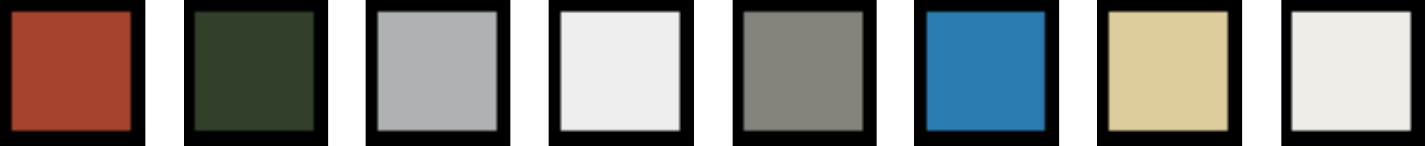} & \includegraphics[width=0.23\linewidth]{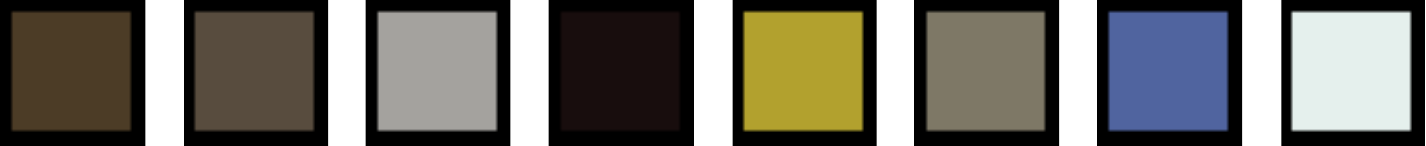} & \includegraphics[width=0.23\linewidth]{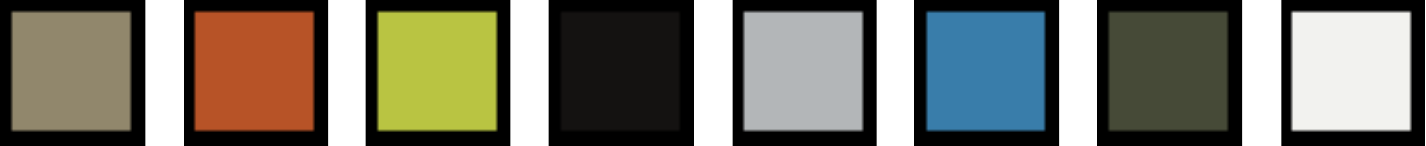} & \includegraphics[width=0.23\linewidth]{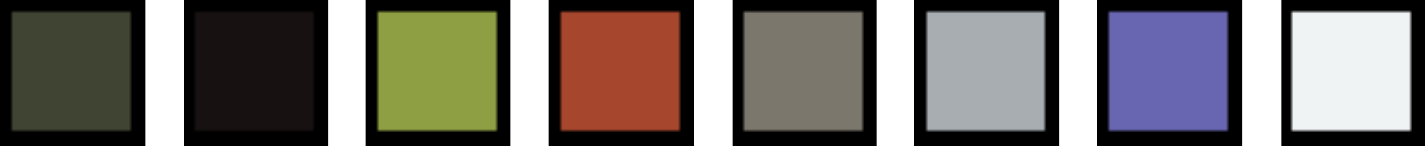} \\
        \textit{dkl2rgb} & \textit{dkl2lms} & \textit{dkl2dkl} & \textit{dkl2lab} \\
        \includegraphics[width=0.23\linewidth]{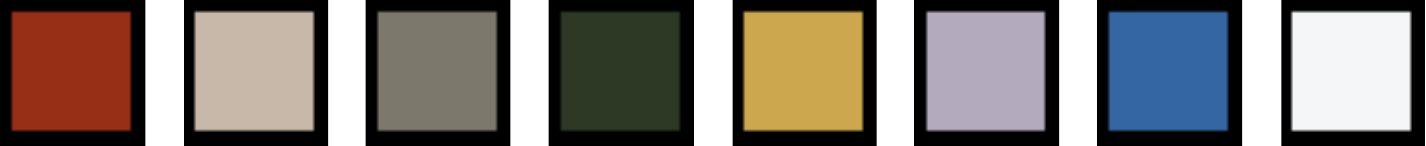} & \includegraphics[width=0.23\linewidth]{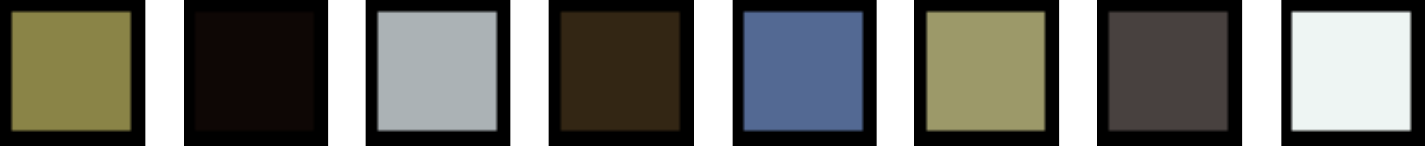} & \includegraphics[width=0.23\linewidth]{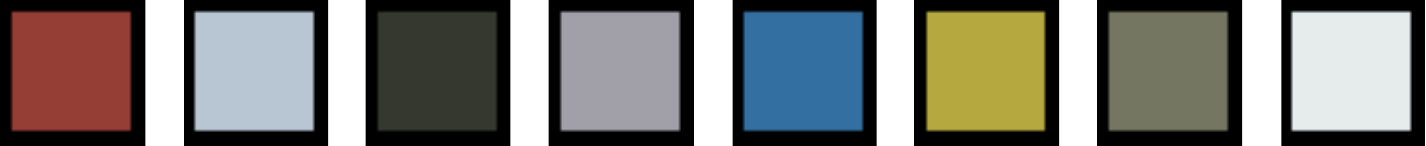} & \includegraphics[width=0.23\linewidth]{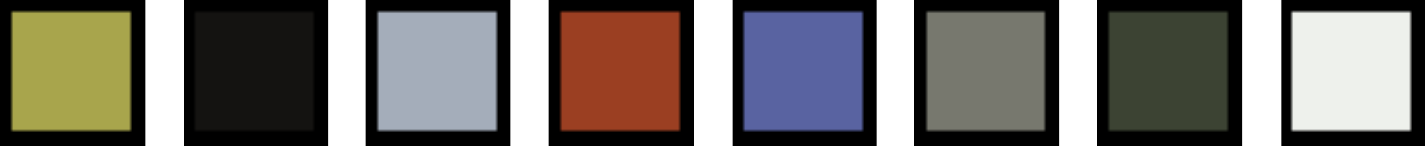} \\
        \textit{lab2rgb} & \textit{lab2lms} & \textit{lab2dkl} & \textit{lab2lab} \\
        \includegraphics[width=0.23\linewidth]{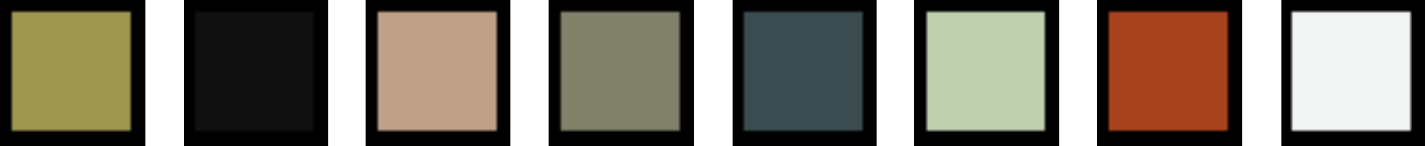} & \includegraphics[width=0.23\linewidth]{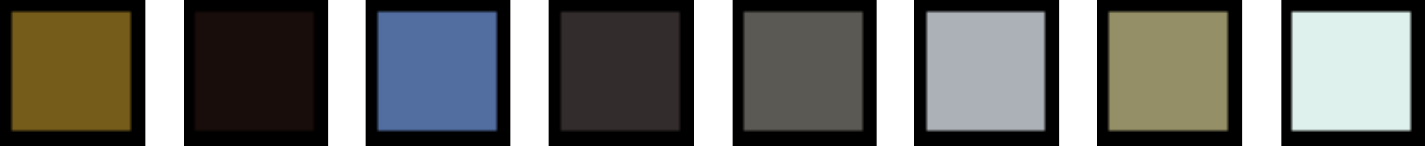} & \includegraphics[width=0.23\linewidth]{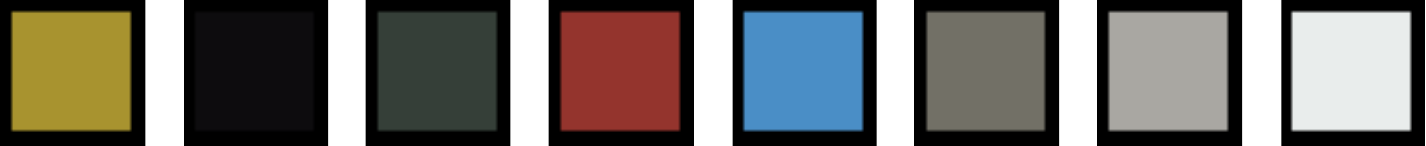} & \includegraphics[width=0.23\linewidth]{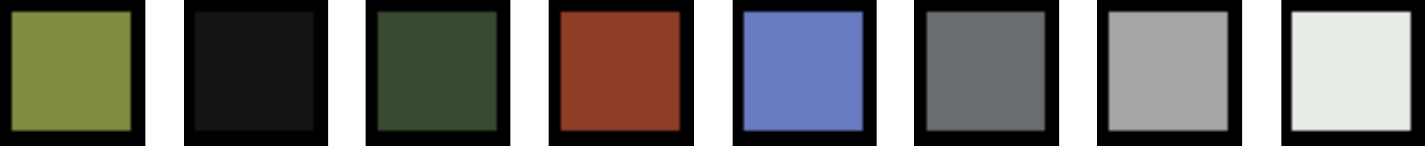} \\
    \end{tabular}
    \caption{The reconstruction output by selecting a single vector of the entire embedding space. All models are VQ-VAE of \textbf{K=8} and \textbf{D=128}.}
    \label{fig:unique_hues8x128}
\end{figure}

\begin{figure}[h]
    \centering \renewcommand{\arraystretch}{1.5}
    \begin{tabular}{@{\hskip0pt}c@{\hskip10pt}c@{\hskip10pt}c@{\hskip10pt}c@{\hskip0pt}}
        \textit{rgb2rgb} & \textit{rgb2lms} & \textit{rgb2dkl} & \textit{rgb2lab} \\
        \includegraphics[width=0.23\linewidth]{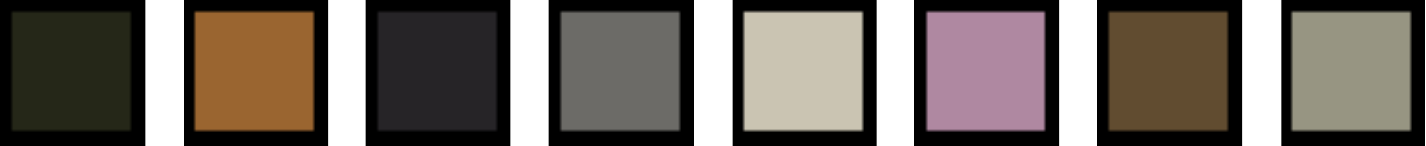} & \includegraphics[width=0.23\linewidth]{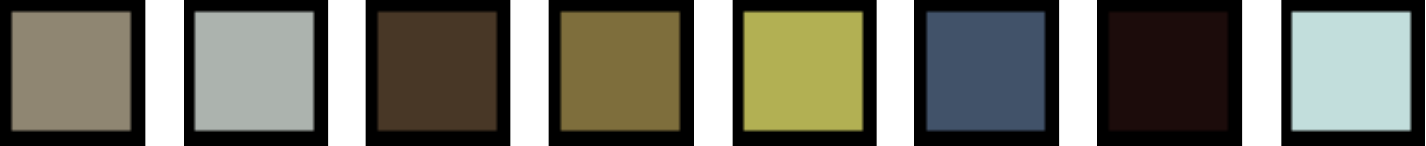} & \includegraphics[width=0.23\linewidth]{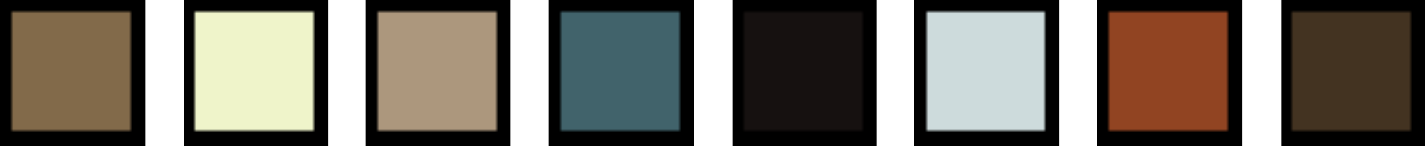} & \includegraphics[width=0.23\linewidth]{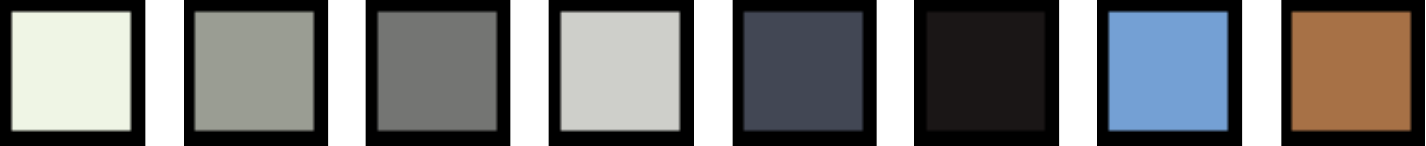} \\
        \textit{lms2rgb} & \textit{lms2lms} & \textit{lms2dkl} & \textit{lms2lab} \\
        \includegraphics[width=0.23\linewidth]{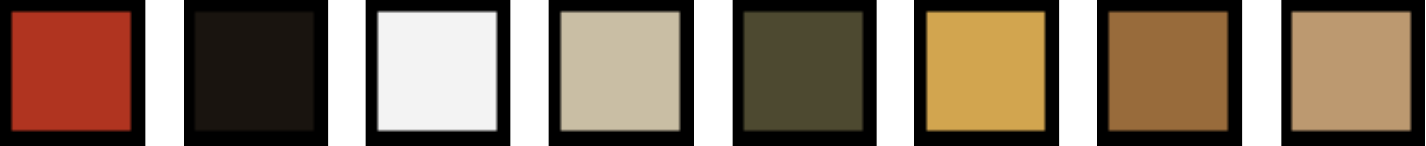} & \includegraphics[width=0.23\linewidth]{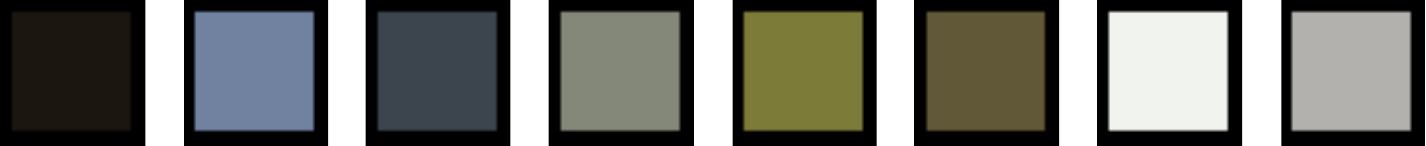} & \includegraphics[width=0.23\linewidth]{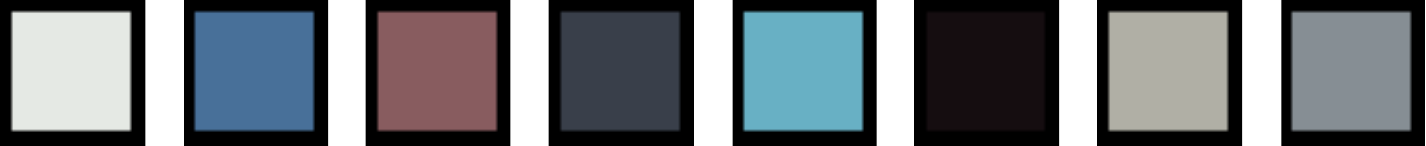} & \includegraphics[width=0.23\linewidth]{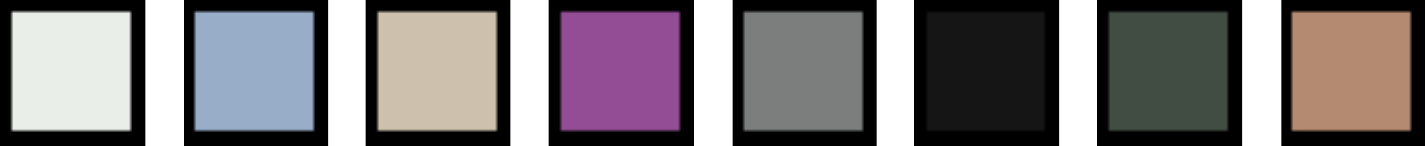} \\
        \textit{dkl2rgb} & \textit{dkl2lms} & \textit{dkl2dkl} & \textit{dkl2lab} \\
        \includegraphics[width=0.23\linewidth]{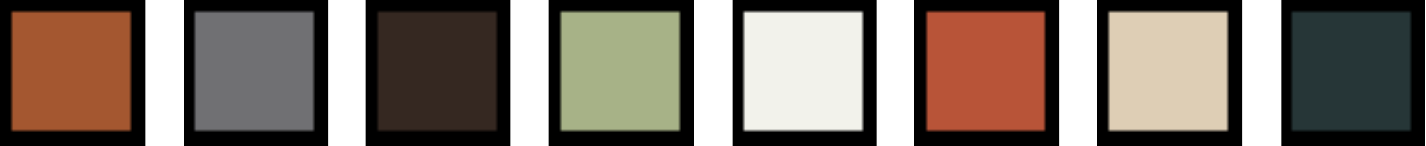} & \includegraphics[width=0.23\linewidth]{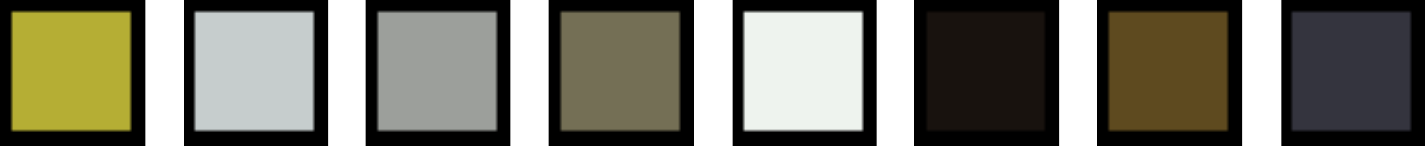} & \includegraphics[width=0.23\linewidth]{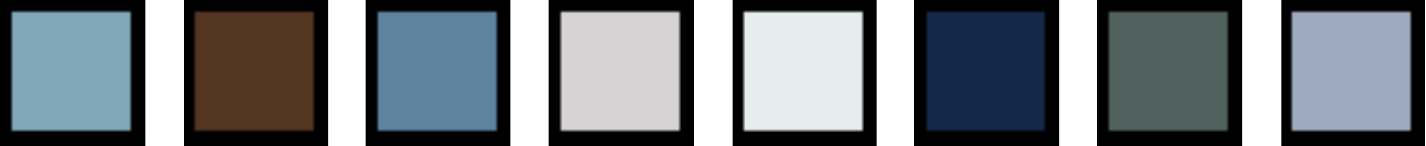} & \includegraphics[width=0.23\linewidth]{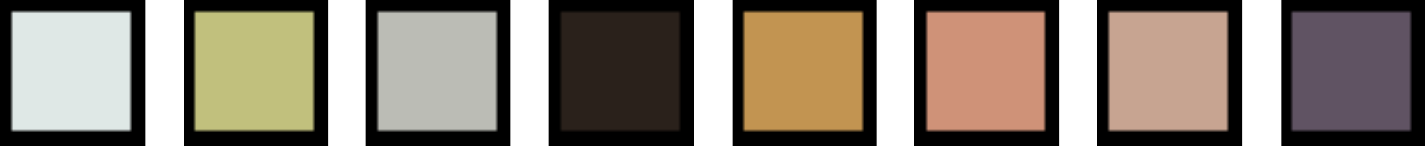} \\
        \textit{lab2rgb} & \textit{lab2lms} & \textit{lab2dkl} & \textit{lab2lab} \\
        \includegraphics[width=0.23\linewidth]{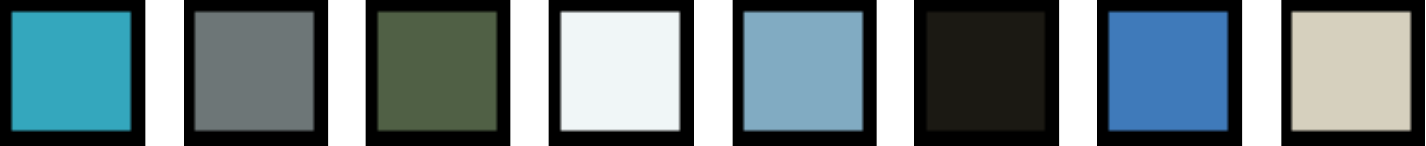} & \includegraphics[width=0.23\linewidth]{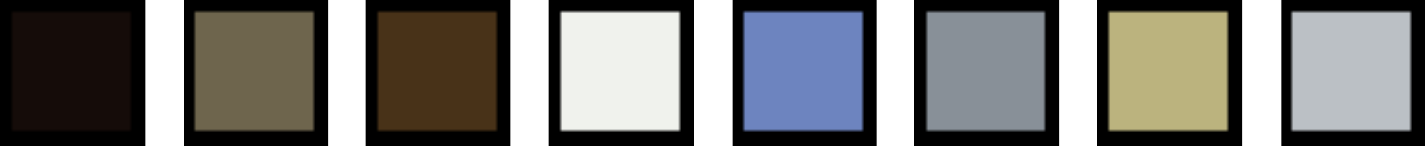} & \includegraphics[width=0.23\linewidth]{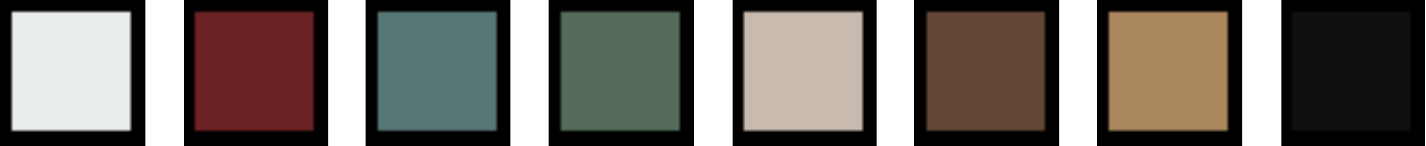} & \includegraphics[width=0.23\linewidth]{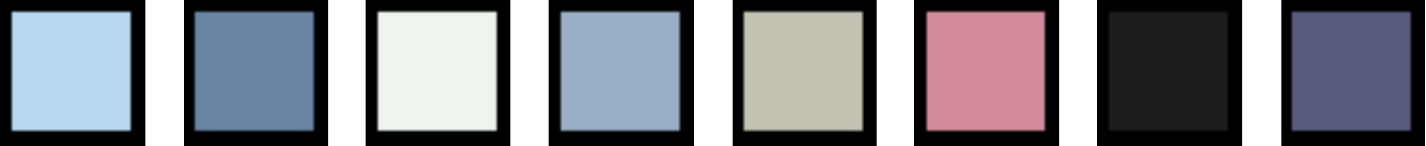} \\
    \end{tabular}
    \caption{The reconstruction output by selecting a single vector of the entire embedding space. All models are VQ-VAE of \textbf{K=8} and \textbf{D=8}.}
    \label{fig:unique_hues8x8}
\end{figure}

Figure~\ref{fig:unique_hues8x128} shows the reconstructed images for all network combinations with embedding space $\{e\} \in \mathbb{R}^{8 \times 128}$. In each row, the input colour space is the same. In each column, the output colour space is the same. An interesting column-wise feature appears. Networks with an identical colour space output share a similar set of hues arranged in different orders. The order within the embedding space of VQ-VAEs is arbitrary and changing it does not impact the network's output. This is an interesting phenomenon suggesting the colour representation in network's embedding space is an attribute of its output colour space. This is less evident in \textit{rgb2rgb} networks and networks of smaller embedding space, $\{e\} \in \mathbb{R}^{8 \times 8}$, presented in Figure~\ref{fig:unique_hues8x8}. This is an exciting line of investigation for feature studies to systematically explore whether the concept of unique hues and colour categories \cite{witzel2018color, siuda2019biological} emerge in machine colour representation.

\subsection{Linear Modelling of Vector Lesions}

\begin{figure}
    \centering \renewcommand{\arraystretch}{1.5}
    \begin{tabular}{@{\hskip0pt} >{\centering\arraybackslash}m{0.2cm} @{\hskip2pt} >{\centering\arraybackslash}m{.15\linewidth} @{\hskip2pt} >{\centering\arraybackslash}m{.15\linewidth} @{\hskip5pt} >{\centering\arraybackslash}m{.15\linewidth} @{\hskip2pt} >{\centering\arraybackslash}m{.15\linewidth} @{\hskip5pt} >{\centering\arraybackslash}m{.15\linewidth} @{\hskip2pt} >{\centering\arraybackslash}m{.15\linewidth} @{\hskip0pt}}
        \raisebox{1.5cm}{\rotatebox[origin=c]{90}{\scriptsize Full model}} & \multicolumn{2}{c}{\includegraphics[width=.15\linewidth]{dkl_ful_01111111_000000000139.jpg}} & \multicolumn{2}{c}{\includegraphics[width=.15\linewidth]{dkl_ful_01111111_000000577182.jpg}} & \multicolumn{2}{c}{\includegraphics[width=.15\linewidth]{dkl_ful_01111111_000000007784.jpg}} \vspace{-0.9cm} \\
         & Lesion output & Linear model & Lesion output & Linear model & Lesion output & Linear model \\
        \raisebox{0.6\height}{\rotatebox[origin=c]{90}{\scriptsize Vector $e_0$}} & \includegraphics[width=\linewidth]{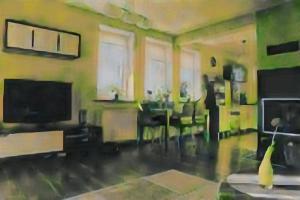} & \includegraphics[width=\linewidth]{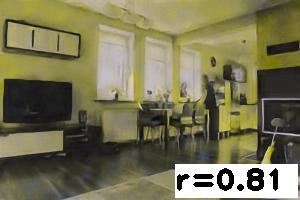} & \includegraphics[width=\linewidth]{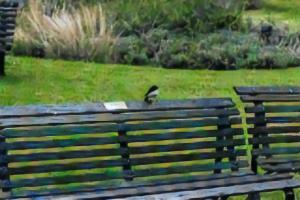} & \includegraphics[width=\linewidth]{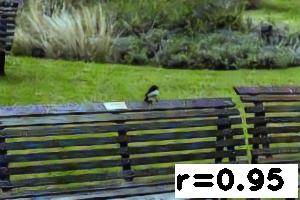} & \includegraphics[width=\linewidth]{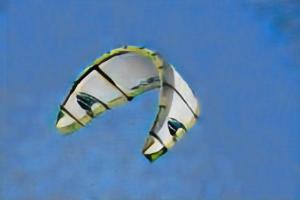} & \includegraphics[width=\linewidth]{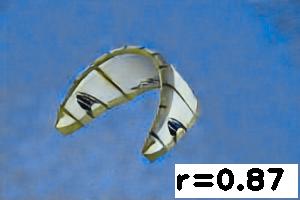} \\
        \raisebox{0.6\height}{\rotatebox[origin=c]{90}{\scriptsize Vector $e_1$}} & \includegraphics[width=\linewidth]{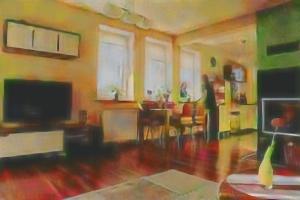} & \includegraphics[width=\linewidth]{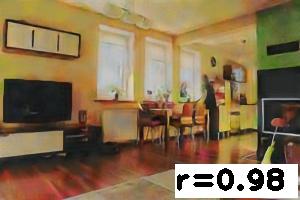} & \includegraphics[width=\linewidth]{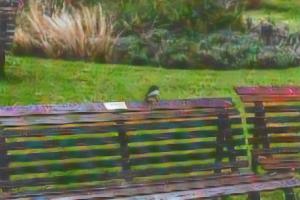} & \includegraphics[width=\linewidth]{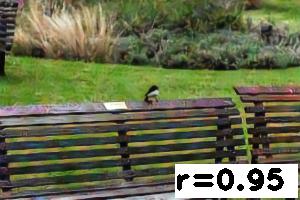} & \includegraphics[width=\linewidth]{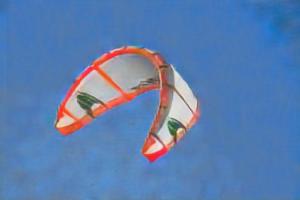} & \includegraphics[width=\linewidth]{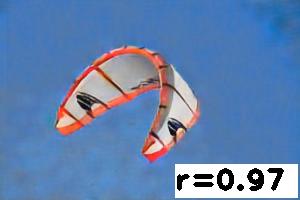} \\
        \raisebox{0.6\height}{\rotatebox[origin=c]{90}{\scriptsize Vector $e_2$}} & \includegraphics[width=\linewidth]{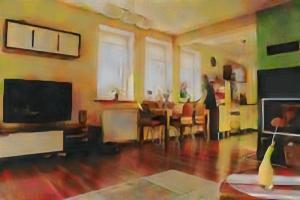} & \includegraphics[width=\linewidth]{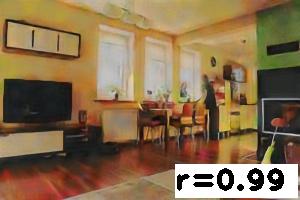} & \includegraphics[width=\linewidth]{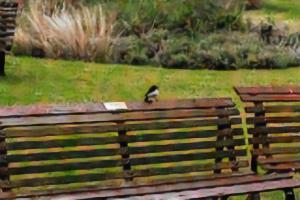} & \includegraphics[width=\linewidth]{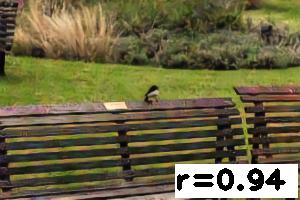} & \includegraphics[width=\linewidth]{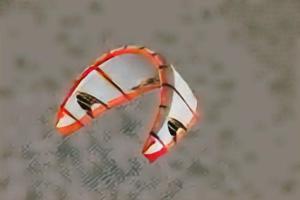} & \includegraphics[width=\linewidth]{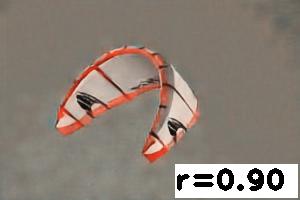} \\
        \raisebox{0.6\height}{\rotatebox[origin=c]{90}{\scriptsize Vector $e_3$}} & \includegraphics[width=\linewidth]{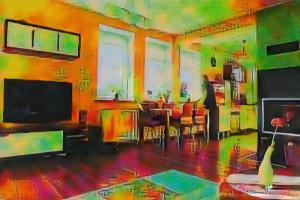} & \includegraphics[width=\linewidth]{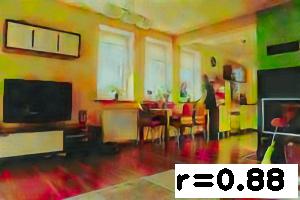} & \includegraphics[width=\linewidth]{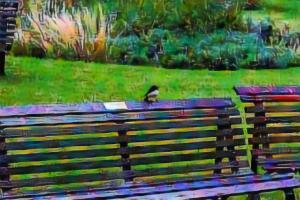} & \includegraphics[width=\linewidth]{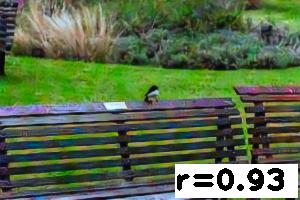} & \includegraphics[width=\linewidth]{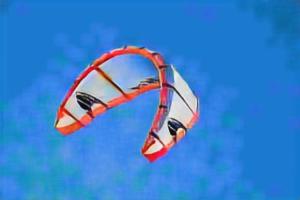} & \includegraphics[width=\linewidth]{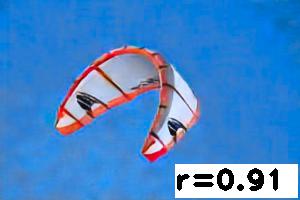} \\
        \raisebox{0.6\height}{\rotatebox[origin=c]{90}{\scriptsize Vector $e_4$}} & \includegraphics[width=\linewidth]{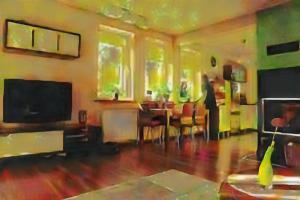} & \includegraphics[width=\linewidth]{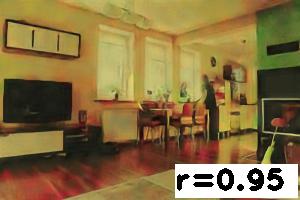} & \includegraphics[width=\linewidth]{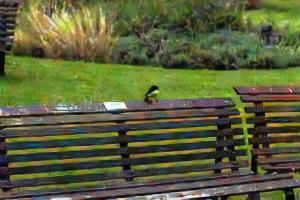} & \includegraphics[width=\linewidth]{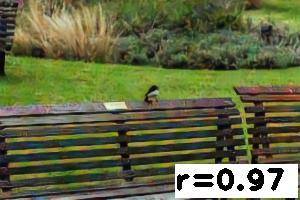} & \includegraphics[width=\linewidth]{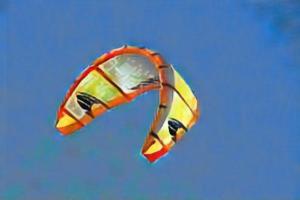} & \includegraphics[width=\linewidth]{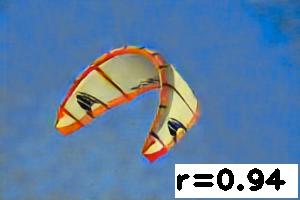} \\
        \raisebox{0.6\height}{\rotatebox[origin=c]{90}{\scriptsize Vector $e_5$}} & \includegraphics[width=\linewidth]{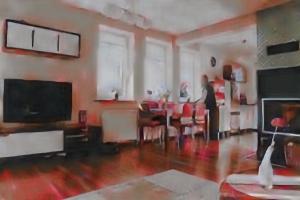} & \includegraphics[width=\linewidth]{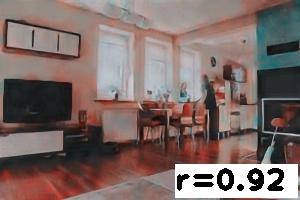} & \includegraphics[width=\linewidth]{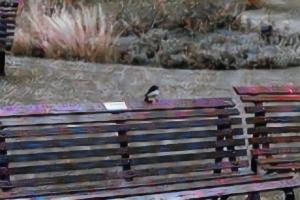} & \includegraphics[width=\linewidth]{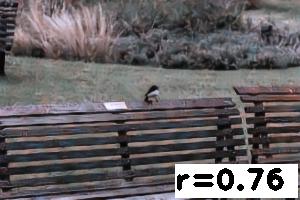} & \includegraphics[width=\linewidth]{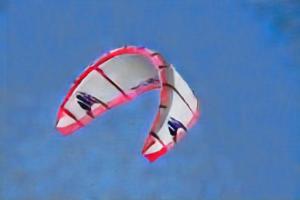} & \includegraphics[width=\linewidth]{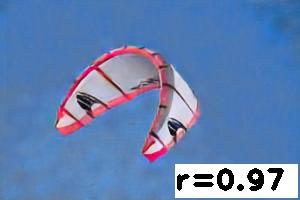} \\
        \raisebox{0.6\height}{\rotatebox[origin=c]{90}{\scriptsize Vector $e_6$}} & \includegraphics[width=\linewidth]{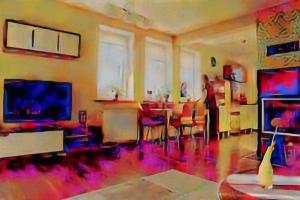} & \includegraphics[width=\linewidth]{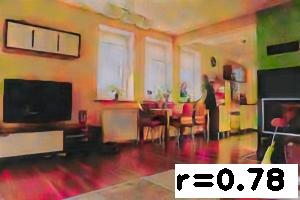} & \includegraphics[width=\linewidth]{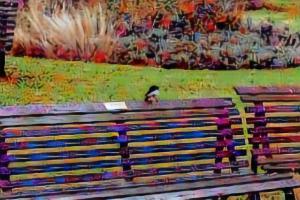} & \includegraphics[width=\linewidth]{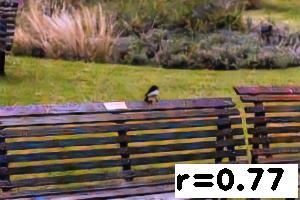} & \includegraphics[width=\linewidth]{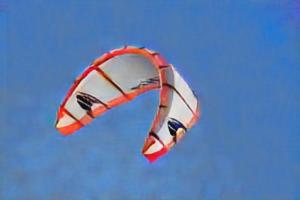} & \includegraphics[width=\linewidth]{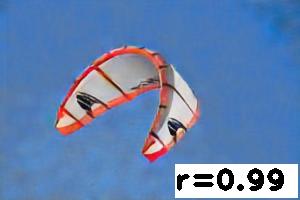} \\
        \raisebox{0.6\height}{\rotatebox[origin=c]{90}{\scriptsize Vector $e_7$}} & \includegraphics[width=\linewidth]{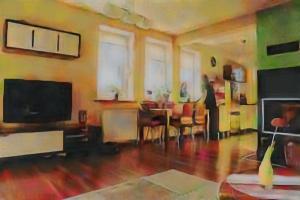} & \includegraphics[width=\linewidth]{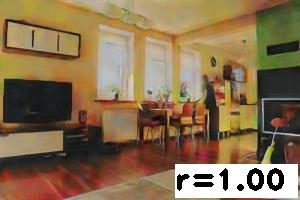} & \includegraphics[width=\linewidth]{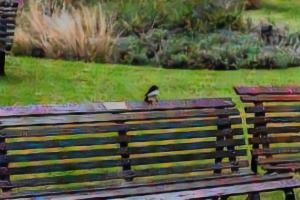} & \includegraphics[width=\linewidth]{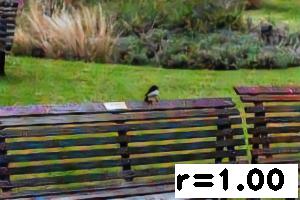} & \includegraphics[width=\linewidth]{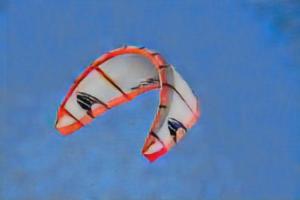} & \includegraphics[width=\linewidth]{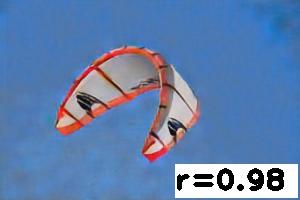} \\
    \end{tabular}
    \caption{The linear modelling of vector lesion for \textbf{\textit{rgb2dkl}} VQ-VAE of \textbf{K=8} and \textbf{D=128}. The denoted $r$ on the bottom right corner of an image is the measure of transformations' fitness.}
    \label{fig:dkl_lesion}
\end{figure}

\begin{figure}
    \centering \renewcommand{\arraystretch}{1.5}
    \begin{tabular}{@{\hskip0pt} >{\centering\arraybackslash}m{0.2cm} @{\hskip2pt} >{\centering\arraybackslash}m{.15\linewidth} @{\hskip2pt} >{\centering\arraybackslash}m{.15\linewidth} @{\hskip5pt} >{\centering\arraybackslash}m{.15\linewidth} @{\hskip2pt} >{\centering\arraybackslash}m{.15\linewidth} @{\hskip5pt} >{\centering\arraybackslash}m{.15\linewidth} @{\hskip2pt} >{\centering\arraybackslash}m{.15\linewidth} @{\hskip0pt}}
        \raisebox{1.5cm}{\rotatebox[origin=c]{90}{\scriptsize Full model}} & \multicolumn{2}{c}{\includegraphics[width=.15\linewidth]{lab_ful_01111111_000000000139.jpg}} & \multicolumn{2}{c}{\includegraphics[width=.15\linewidth]{lab_ful_01111111_000000577182.jpg}} & \multicolumn{2}{c}{\includegraphics[width=.15\linewidth]{lab_ful_01111111_000000007784.jpg}} \vspace{-0.9cm} \\
         & Lesion output & Linear model & Lesion output & Linear model & Lesion output & Linear model \\
        \raisebox{0.6\height}{\rotatebox[origin=c]{90}{\scriptsize Vector $e_0$}} & \includegraphics[width=\linewidth]{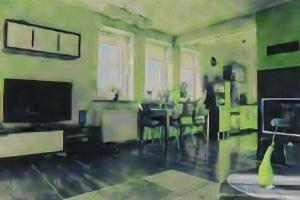} & \includegraphics[width=\linewidth]{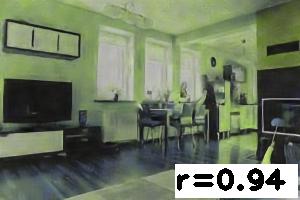} & \includegraphics[width=\linewidth]{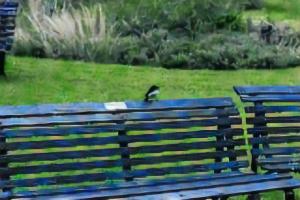} & \includegraphics[width=\linewidth]{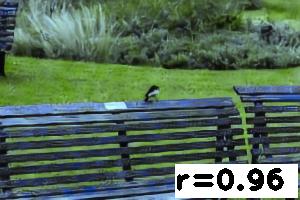} & \includegraphics[width=\linewidth]{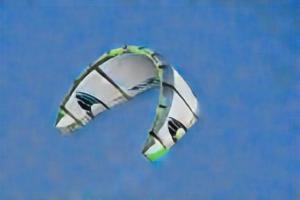} & \includegraphics[width=\linewidth]{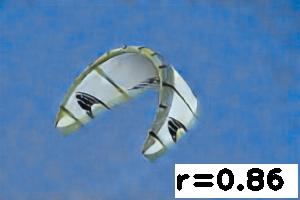} \\
        \raisebox{0.6\height}{\rotatebox[origin=c]{90}{\scriptsize Vector $e_1$}} & \includegraphics[width=\linewidth]{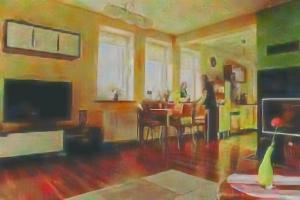} & \includegraphics[width=\linewidth]{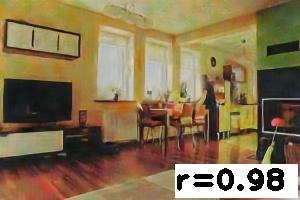} & \includegraphics[width=\linewidth]{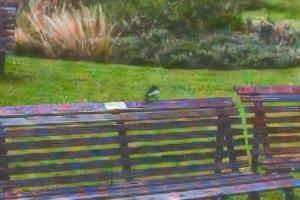} & \includegraphics[width=\linewidth]{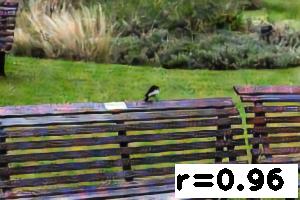} & \includegraphics[width=\linewidth]{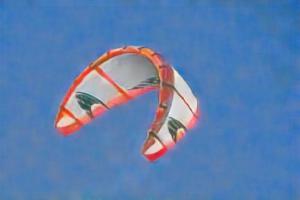} & \includegraphics[width=\linewidth]{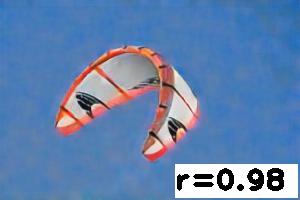} \\
        \raisebox{0.6\height}{\rotatebox[origin=c]{90}{\scriptsize Vector $e_2$}} & \includegraphics[width=\linewidth]{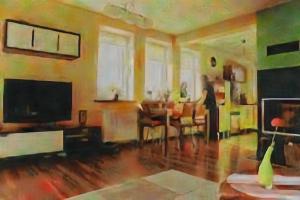} & \includegraphics[width=\linewidth]{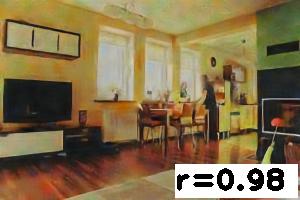} & \includegraphics[width=\linewidth]{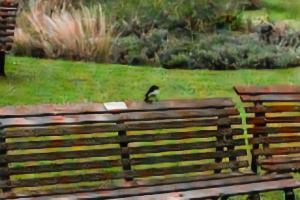} & \includegraphics[width=\linewidth]{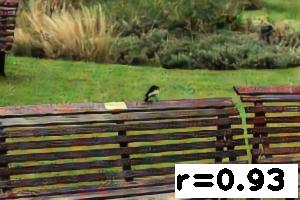} & \includegraphics[width=\linewidth]{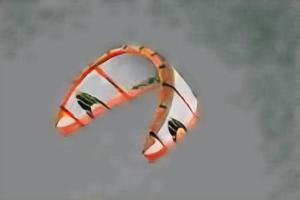} & \includegraphics[width=\linewidth]{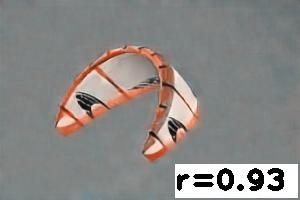} \\
        \raisebox{0.6\height}{\rotatebox[origin=c]{90}{\scriptsize Vector $e_3$}} & \includegraphics[width=\linewidth]{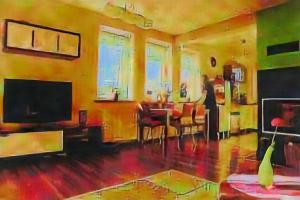} & \includegraphics[width=\linewidth]{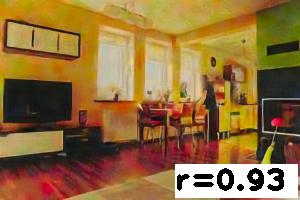} & \includegraphics[width=\linewidth]{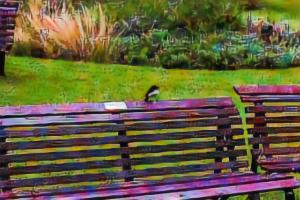} & \includegraphics[width=\linewidth]{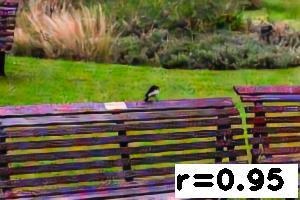} & \includegraphics[width=\linewidth]{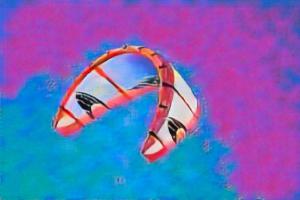} & \includegraphics[width=\linewidth]{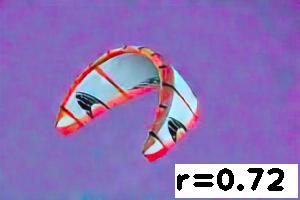} \\
        \raisebox{0.6\height}{\rotatebox[origin=c]{90}{\scriptsize Vector $e_4$}} & \includegraphics[width=\linewidth]{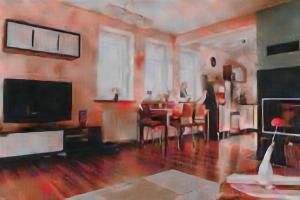} & \includegraphics[width=\linewidth]{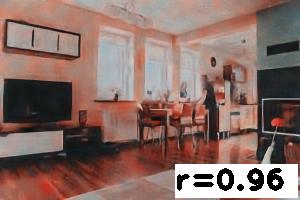} & \includegraphics[width=\linewidth]{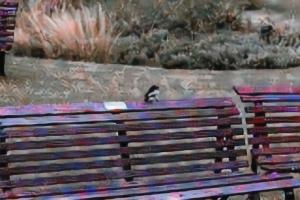} & \includegraphics[width=\linewidth]{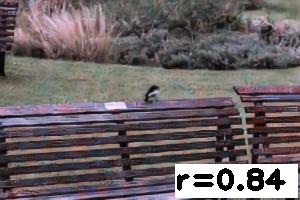} & \includegraphics[width=\linewidth]{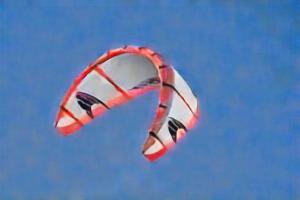} & \includegraphics[width=\linewidth]{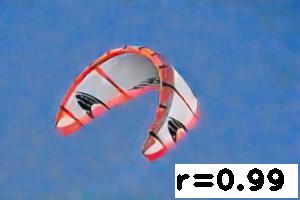} \\
        \raisebox{0.6\height}{\rotatebox[origin=c]{90}{\scriptsize Vector $e_5$}} & \includegraphics[width=\linewidth]{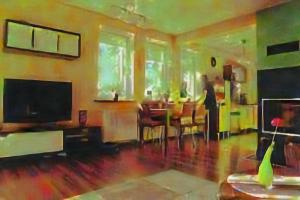} & \includegraphics[width=\linewidth]{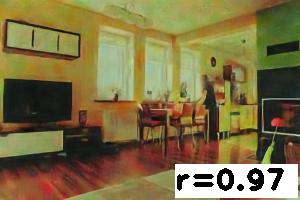} & \includegraphics[width=\linewidth]{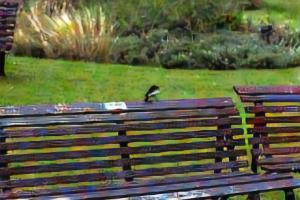} & \includegraphics[width=\linewidth]{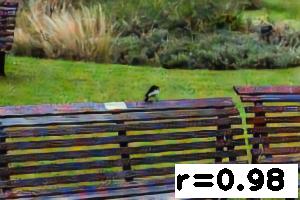} & \includegraphics[width=\linewidth]{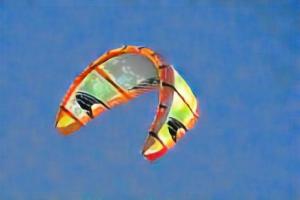} & \includegraphics[width=\linewidth]{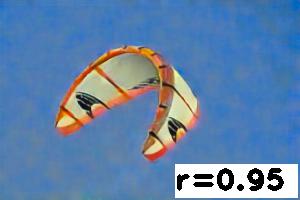} \\
        \raisebox{0.6\height}{\rotatebox[origin=c]{90}{\scriptsize Vector $e_6$}} & \includegraphics[width=\linewidth]{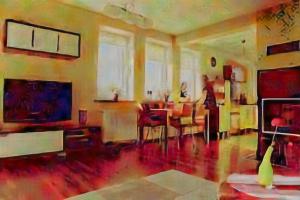} & \includegraphics[width=\linewidth]{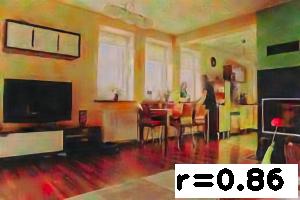} & \includegraphics[width=\linewidth]{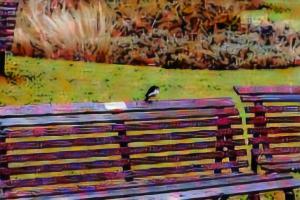} & \includegraphics[width=\linewidth]{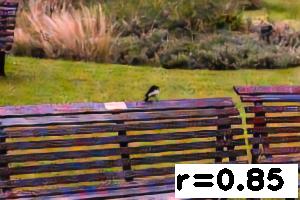} & \includegraphics[width=\linewidth]{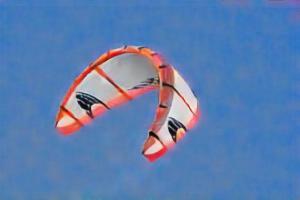} & \includegraphics[width=\linewidth]{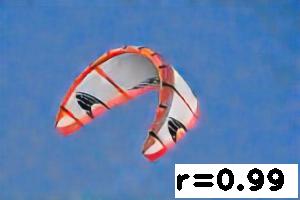} \\
        \raisebox{0.6\height}{\rotatebox[origin=c]{90}{\scriptsize Vector $e_7$}} & \includegraphics[width=\linewidth]{lab_zro_11111110_000000000139.jpg} & \includegraphics[width=\linewidth]{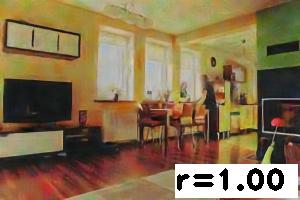} & \includegraphics[width=\linewidth]{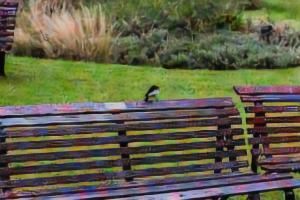} & \includegraphics[width=\linewidth]{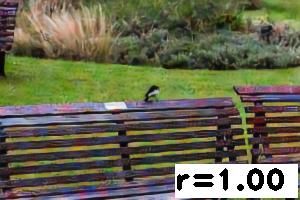} & \includegraphics[width=\linewidth]{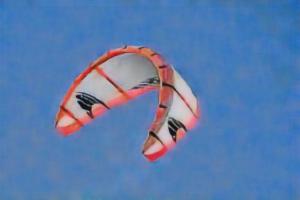} & \includegraphics[width=\linewidth]{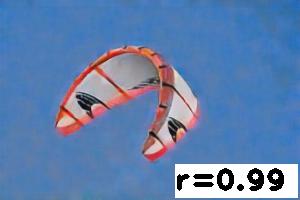} \\
    \end{tabular}
    \caption{The linear modelling of vector lesion for \textbf{\textit{rgb2lab}} VQ-VAE of \textbf{K=8} and \textbf{D=128}. The denoted $r$ on the bottom right corner of an image is the measure of transformations' fitness.}
    \label{fig:lab_lesion}
\end{figure}

\begin{figure}
    \centering \renewcommand{\arraystretch}{1.5}
    \begin{tabular}{@{\hskip0pt} >{\centering\arraybackslash}m{0.2cm} @{\hskip2pt} >{\centering\arraybackslash}m{.15\linewidth} @{\hskip2pt} >{\centering\arraybackslash}m{.15\linewidth} @{\hskip5pt} >{\centering\arraybackslash}m{.15\linewidth} @{\hskip2pt} >{\centering\arraybackslash}m{.15\linewidth} @{\hskip5pt} >{\centering\arraybackslash}m{.15\linewidth} @{\hskip2pt} >{\centering\arraybackslash}m{.15\linewidth} @{\hskip0pt}}
        \raisebox{1.5cm}{\rotatebox[origin=c]{90}{\scriptsize Full model}} & \multicolumn{2}{c}{\includegraphics[width=.15\linewidth]{rgb_ful_01111111_000000000139.jpg}} & \multicolumn{2}{c}{\includegraphics[width=.15\linewidth]{rgb_ful_01111111_000000577182.jpg}} & \multicolumn{2}{c}{\includegraphics[width=.15\linewidth]{rgb_ful_01111111_000000007784.jpg}} \vspace{-0.9cm} \\
         & Lesion output & Linear model & Lesion output & Linear model & Lesion output & Linear model \\
        \raisebox{0.6\height}{\rotatebox[origin=c]{90}{\scriptsize Vector $e_0$}} & \includegraphics[width=\linewidth]{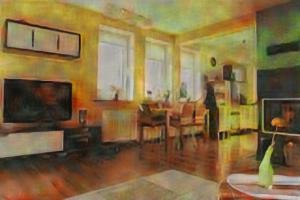} & \includegraphics[width=\linewidth]{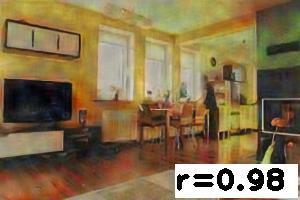} & \includegraphics[width=\linewidth]{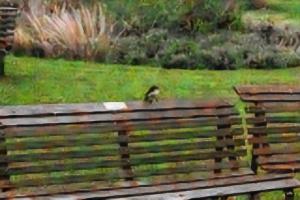} & \includegraphics[width=\linewidth]{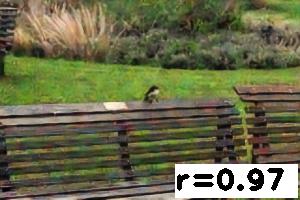} & \includegraphics[width=\linewidth]{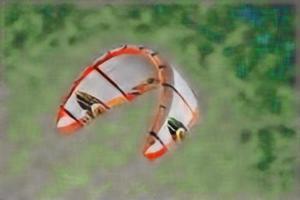} & \includegraphics[width=\linewidth]{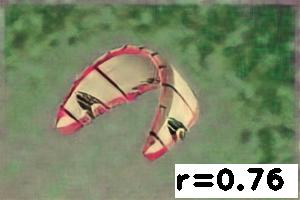} \\
        \raisebox{0.6\height}{\rotatebox[origin=c]{90}{\scriptsize Vector $e_1$}} & \includegraphics[width=\linewidth]{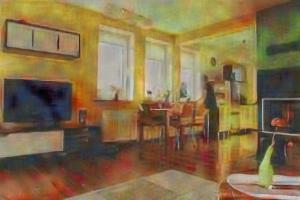} & \includegraphics[width=\linewidth]{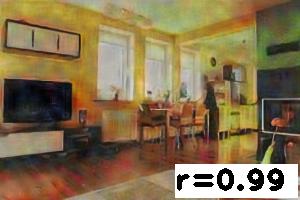} & \includegraphics[width=\linewidth]{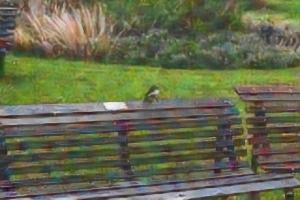} & \includegraphics[width=\linewidth]{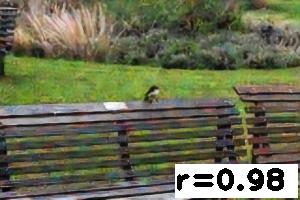} & \includegraphics[width=\linewidth]{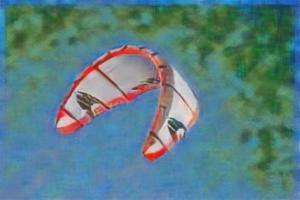} & \includegraphics[width=\linewidth]{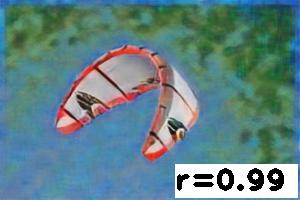} \\
        \raisebox{0.6\height}{\rotatebox[origin=c]{90}{\scriptsize Vector $e_2$}} & \includegraphics[width=\linewidth]{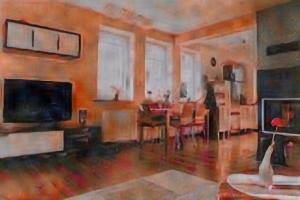} & \includegraphics[width=\linewidth]{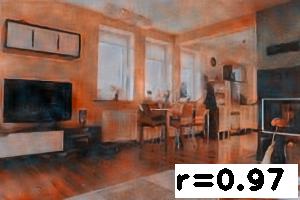} & \includegraphics[width=\linewidth]{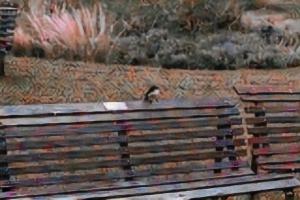} & \includegraphics[width=\linewidth]{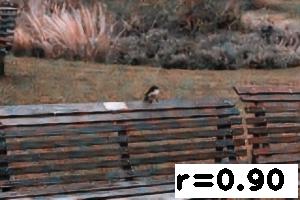} & \includegraphics[width=\linewidth]{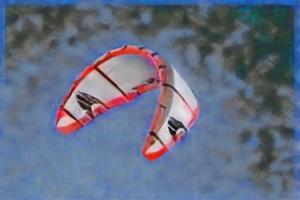} & \includegraphics[width=\linewidth]{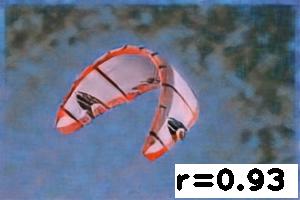} \\
        \raisebox{0.6\height}{\rotatebox[origin=c]{90}{\scriptsize Vector $e_3$}} & \includegraphics[width=\linewidth]{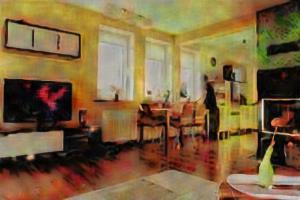} & \includegraphics[width=\linewidth]{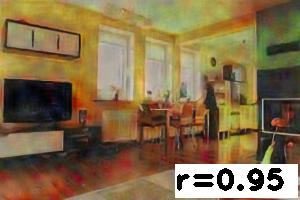} & \includegraphics[width=\linewidth]{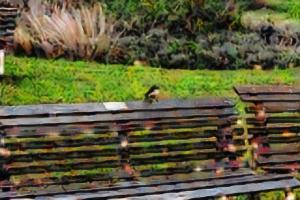} & \includegraphics[width=\linewidth]{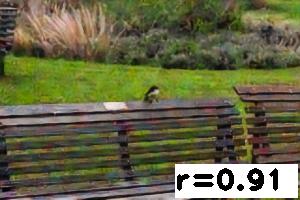} & \includegraphics[width=\linewidth]{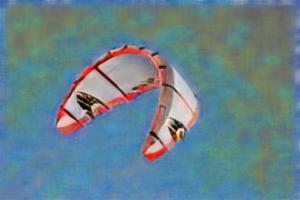} & \includegraphics[width=\linewidth]{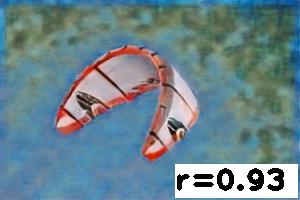} \\
        \raisebox{0.6\height}{\rotatebox[origin=c]{90}{\scriptsize Vector $e_4$}} & \includegraphics[width=\linewidth]{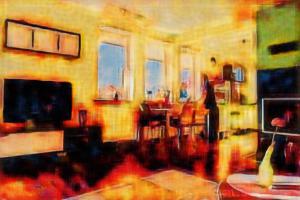} & \includegraphics[width=\linewidth]{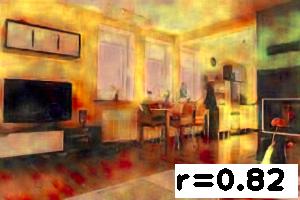} & \includegraphics[width=\linewidth]{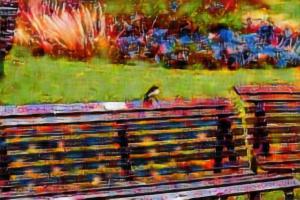} & \includegraphics[width=\linewidth]{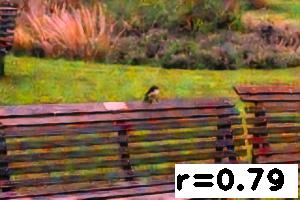} & \includegraphics[width=\linewidth]{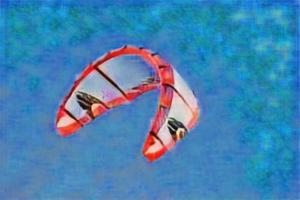} & \includegraphics[width=\linewidth]{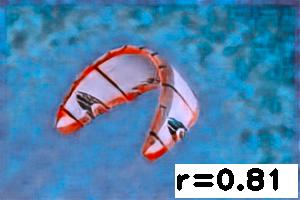} \\
        \raisebox{0.6\height}{\rotatebox[origin=c]{90}{\scriptsize Vector $e_5$}} & \includegraphics[width=\linewidth]{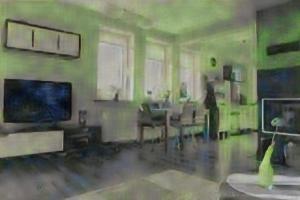} & \includegraphics[width=\linewidth]{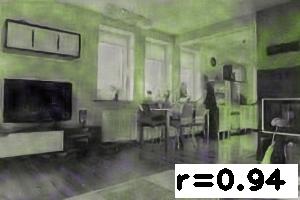} & \includegraphics[width=\linewidth]{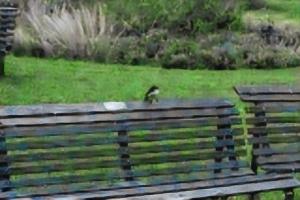} & \includegraphics[width=\linewidth]{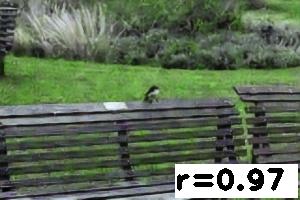} & \includegraphics[width=\linewidth]{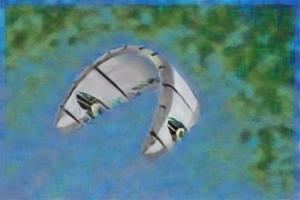} & \includegraphics[width=\linewidth]{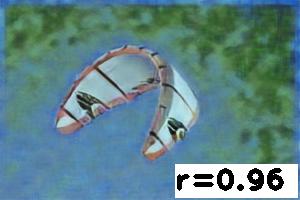} \\
        \raisebox{0.6\height}{\rotatebox[origin=c]{90}{\scriptsize Vector $e_6$}} & \includegraphics[width=\linewidth]{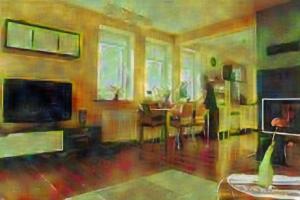} & \includegraphics[width=\linewidth]{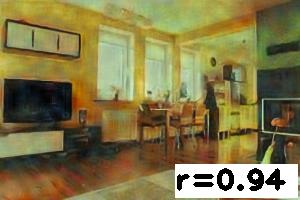} & \includegraphics[width=\linewidth]{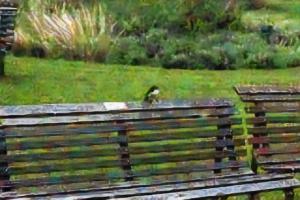} & \includegraphics[width=\linewidth]{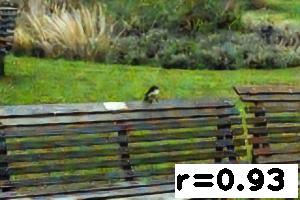} & \includegraphics[width=\linewidth]{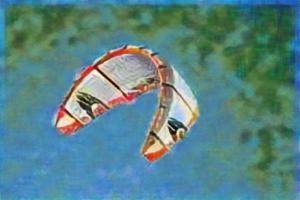} & \includegraphics[width=\linewidth]{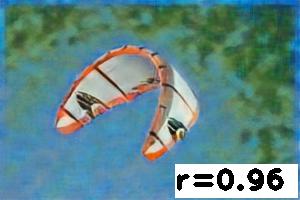} \\
        \raisebox{0.6\height}{\rotatebox[origin=c]{90}{\scriptsize Vector $e_7$}} & \includegraphics[width=\linewidth]{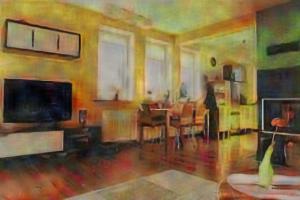} & \includegraphics[width=\linewidth]{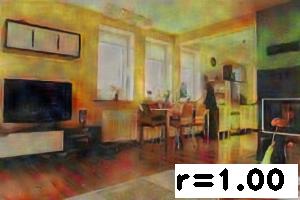} & \includegraphics[width=\linewidth]{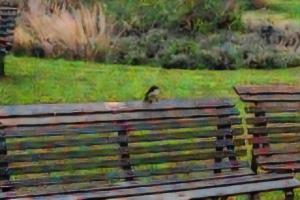} & \includegraphics[width=\linewidth]{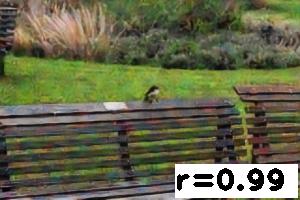} & \includegraphics[width=\linewidth]{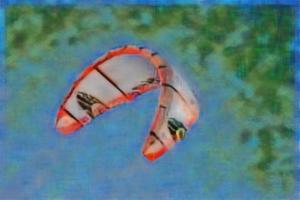} & \includegraphics[width=\linewidth]{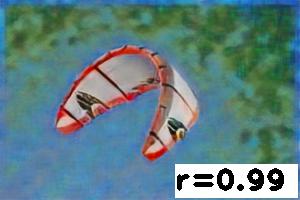} \\
    \end{tabular}
    \caption{The linear modelling of vector lesion for \textbf{\textit{rgb2rgb}} VQ-VAE of \textbf{K=8} and \textbf{D=128}. The denoted $r$ on the bottom right corner of an image is the measure of transformations' fitness.}
    \label{fig:rgb_lesion}
\end{figure}

In order to understand the features learnt by the colour conversion networks, we exercised the ``lesion'' technique. It consists of silencing the embedding space's vectors one at a time. We explored whether a vector lesion can be modelled by a simple linear transformation. We estimated the transformation matrix that maps the pixel distribution of the full reconstruction onto the lesion image (refer to Section 4 in the manuscript). To our surprise, this simple parsimonious modelling can capture a large portion of the vector's encoding. We present qualitative results for VQ-VAEs with $K=8$ and $D = 128$ in the colour conversion networks \textit{rgb2dkl} (Figure \ref{fig:dkl_lesion}), \textit{rgb2lab} (Figure \ref{fig:lab_lesion}) and \textit{rgb2rgb} (Figure \ref{fig:rgb_lesion}).

The top row of Figure~\ref{fig:dkl_lesion} illustrated three examples from the COCO dataset reconstructed with the full model of \textit{rgb2dkl}. The following rows depict the reconstruction output of lesion technique exercised on each embedding vector $e_i$ (the ``Lesion output'' column), alongside with the linear modelling estimation (the ``Linear model'' column) obtained by applying the linear transformation to the full reconstructed image. On the bottom right corner, we have reported the fitness of the mode, correlation in the CIE L*a*b* colour coordinates ($r$) between the colour pixels of the lesion output and the linear model. The overall fitness is very high for such a simple model. Even in cases with lower correlations, we can observe that the model captures well the characteristics of the lesion output. For instance, the indoor scene for $e_0$ obtains $r=0.81$, however, it can be appreciated that the linear model accounts well for the disappearance of red pixels in the lesion output. This is also evident in the kite picture of $e_2$ where blue pixels have vanished or the bench picture of $e_5$ with green pixels.

Naturally, there are limits to this linear modelling. For instance, the excess of chromaticity (pink and blue colours) in the indoor scene of $e_6$ is not fully captured by its linear model. The most extreme can be observed in the kite picture of $e_3$ for the \textit{rgb2lab} model (Figure \ref{fig:lab_lesion}) where the non-linear nature of lesion output is not accounted for in the linear model. Nevertheless, these parsimonious transformations reveal great details about the information encoded by each vector deserving more thorough investigation in future studies.

\begin{figure}
    \centering
    \begin{tabular}{@{\hskip0pt} >{\centering\arraybackslash}m{0.2cm} @{\hskip2pt} >{\centering\arraybackslash}m{.31\linewidth} @{\hskip2pt} >{\centering\arraybackslash}m{.31\linewidth} @{\hskip2pt} >{\centering\arraybackslash}m{.31\linewidth} @{\hskip0pt}}
    & & RGB Cube in CIE L*a*b* & \\
    & & \includegraphics[width=\linewidth]{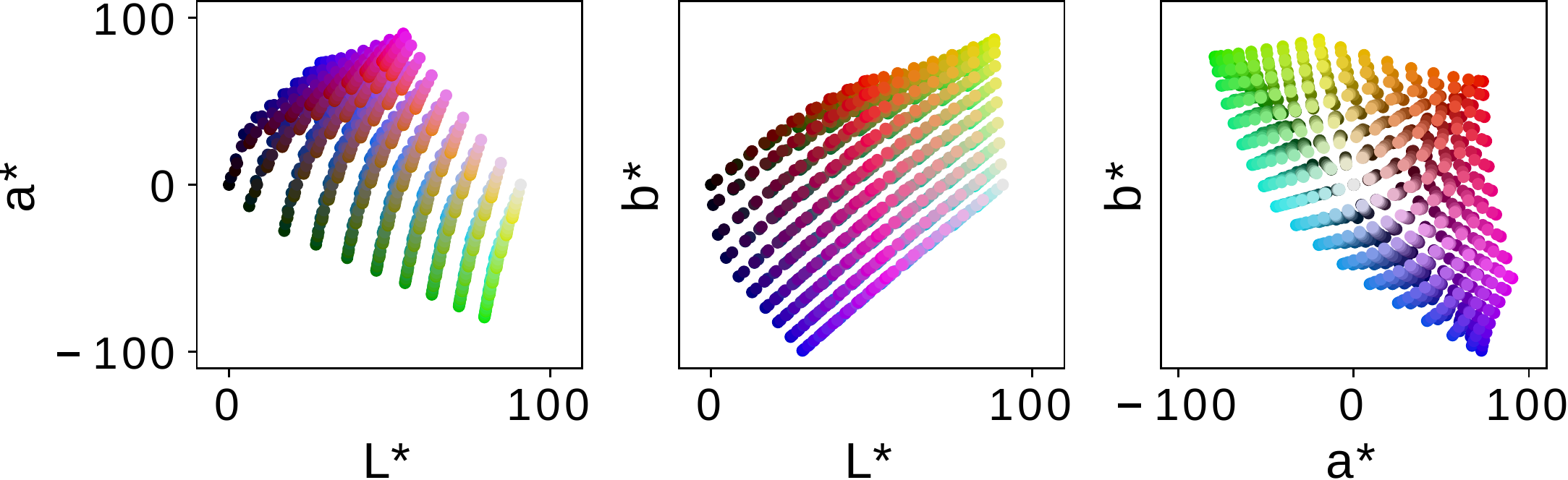} & \\
    & rgb2rgb & rgb2dkl & rgb2lab \\
    \raisebox{1\height}{\rotatebox[origin=c]{90}{\scriptsize Vector $e_0$}} & \includegraphics[width=\linewidth]{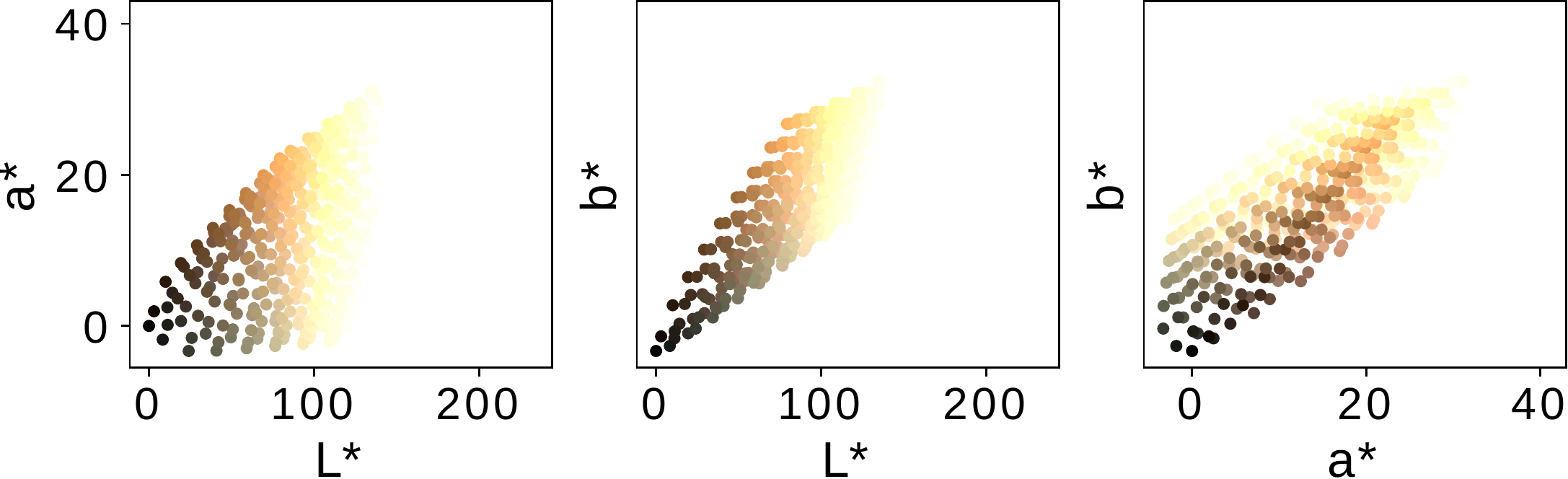} & \includegraphics[width=\linewidth]{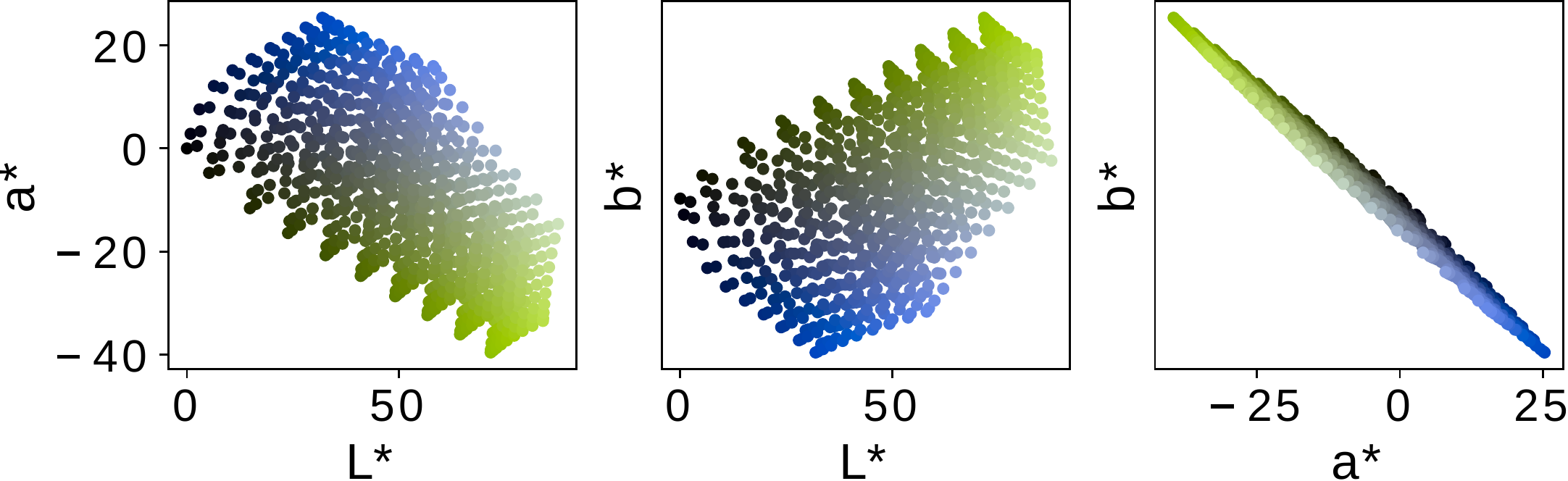} & \includegraphics[width=\linewidth]{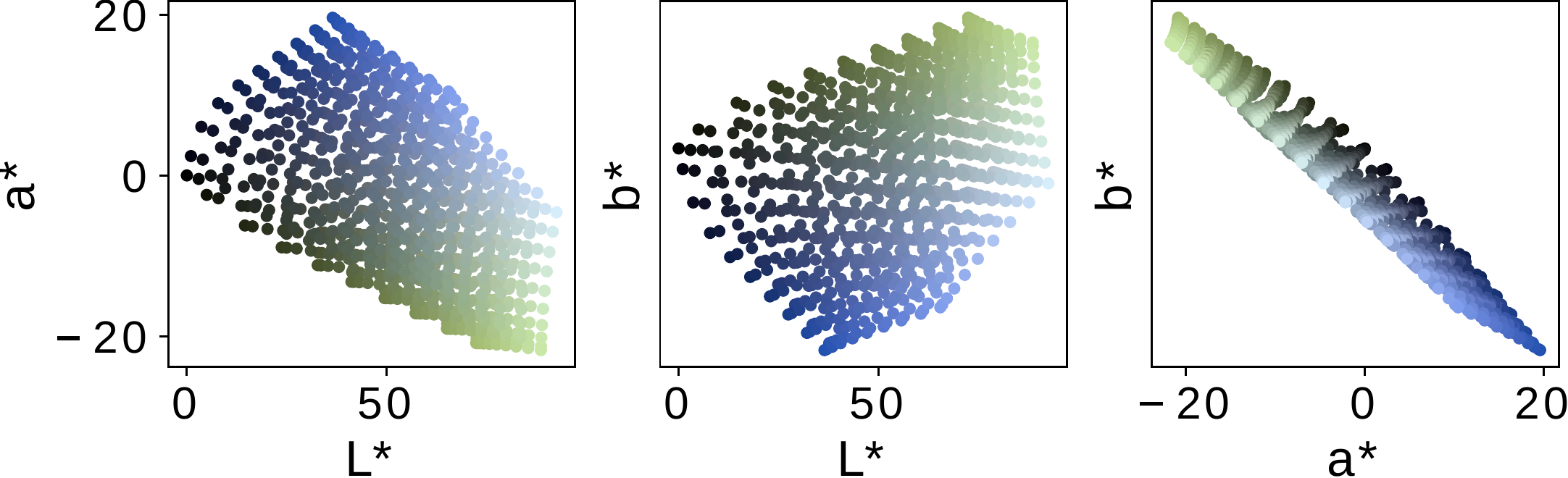} \\
    \raisebox{1\height}{\rotatebox[origin=c]{90}{\scriptsize Vector $e_1$}} & \includegraphics[width=\linewidth]{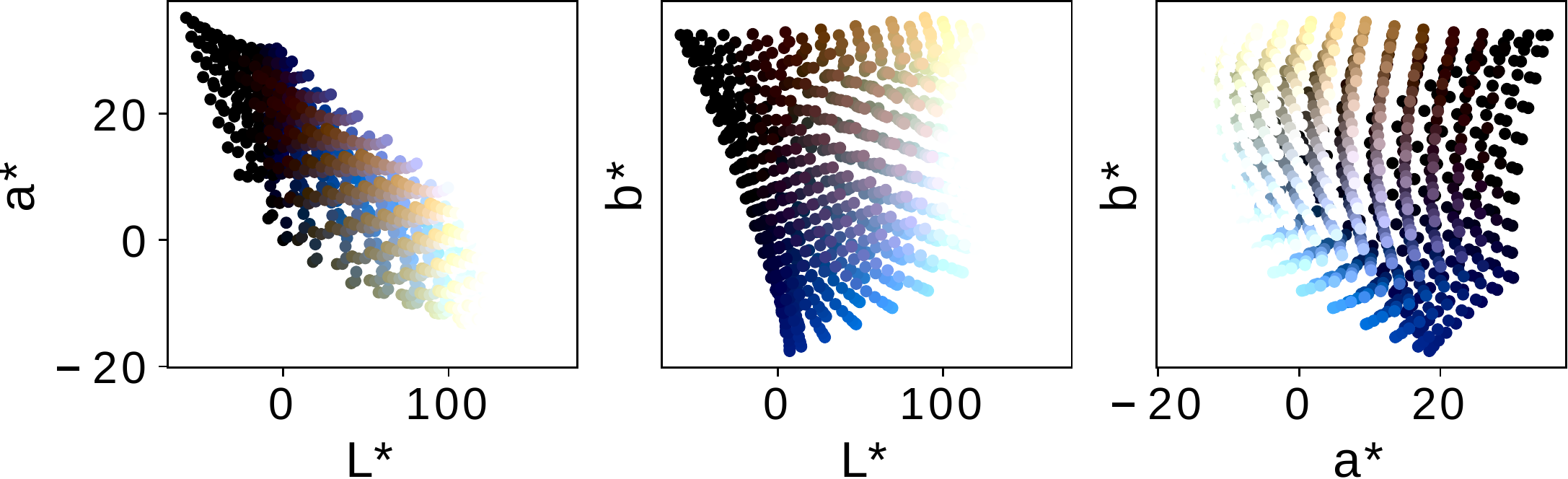} & \includegraphics[width=\linewidth]{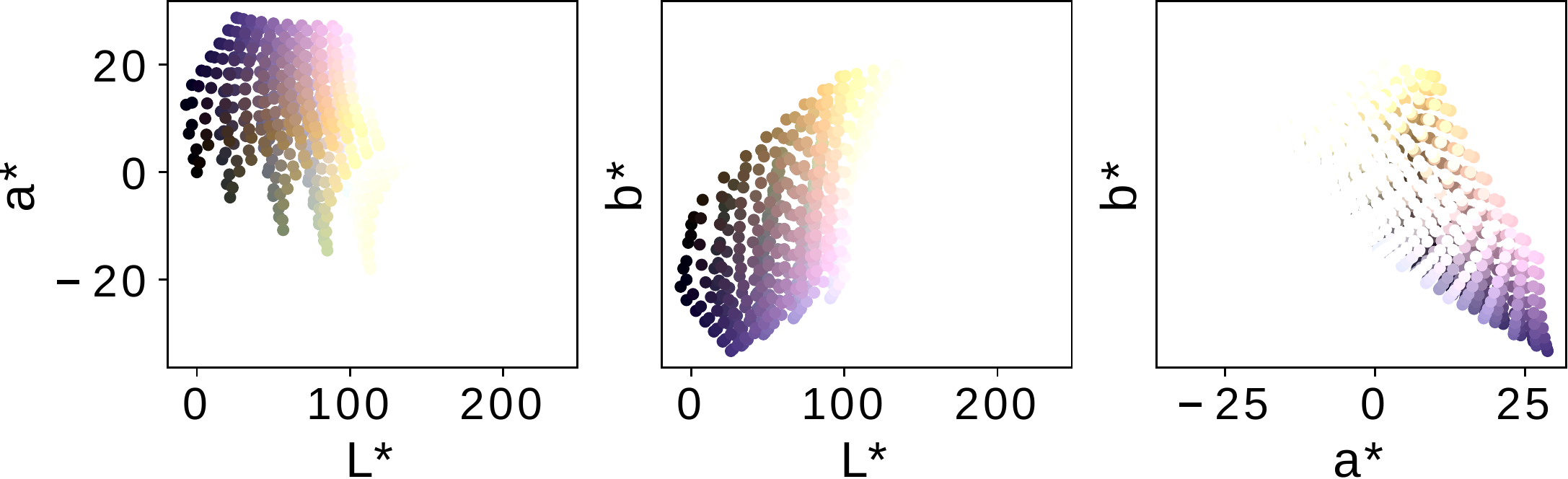} & \includegraphics[width=\linewidth]{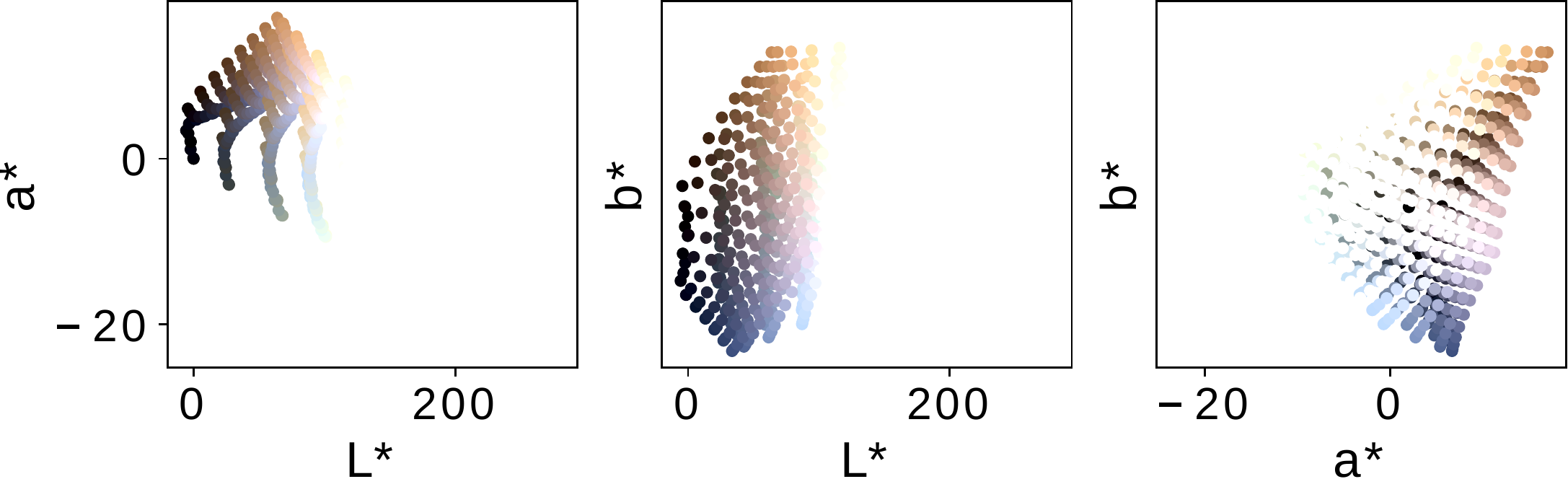} \\
    \raisebox{1\height}{\rotatebox[origin=c]{90}{\scriptsize Vector $e_2$}} & \includegraphics[width=\linewidth]{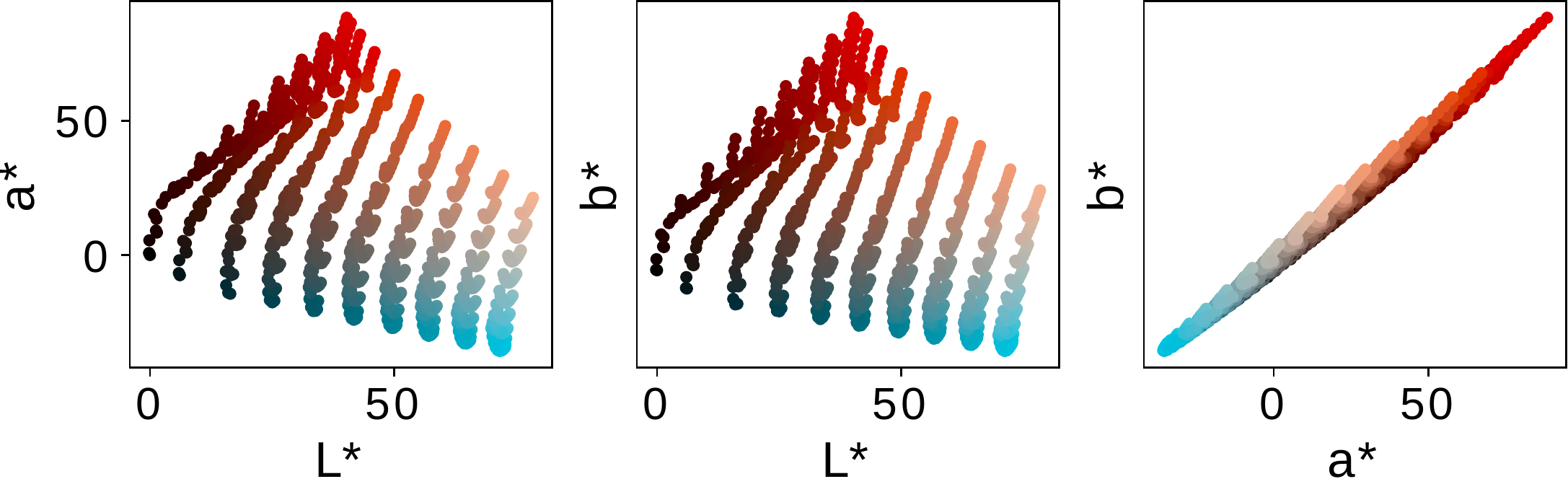} & \includegraphics[width=\linewidth]{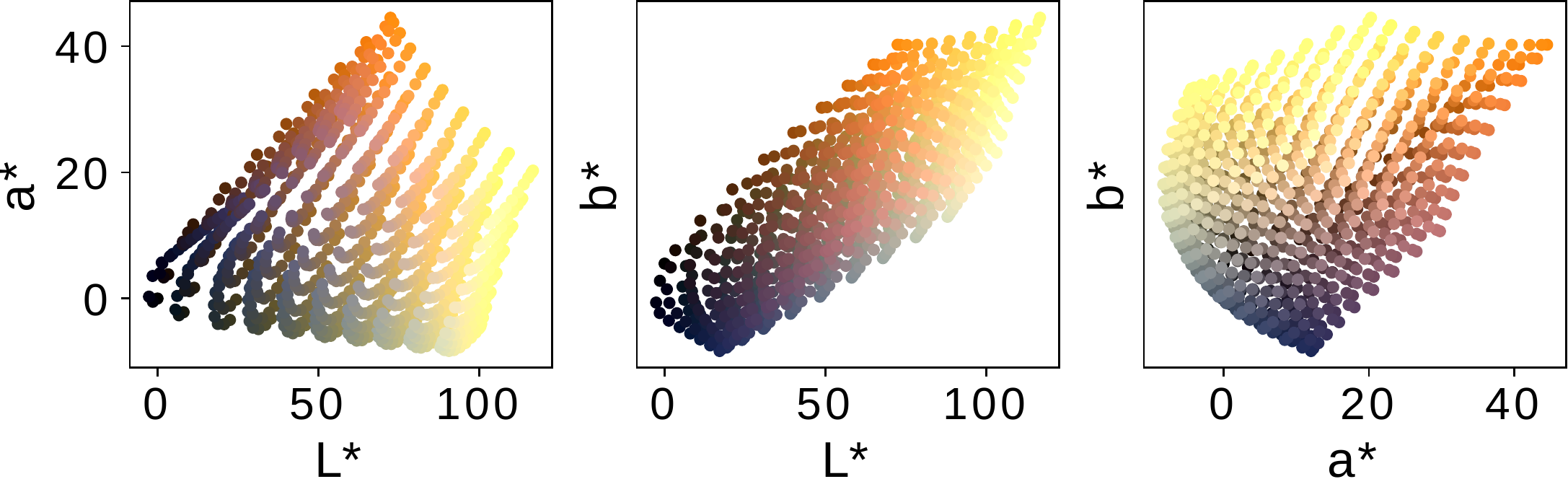} & \includegraphics[width=\linewidth]{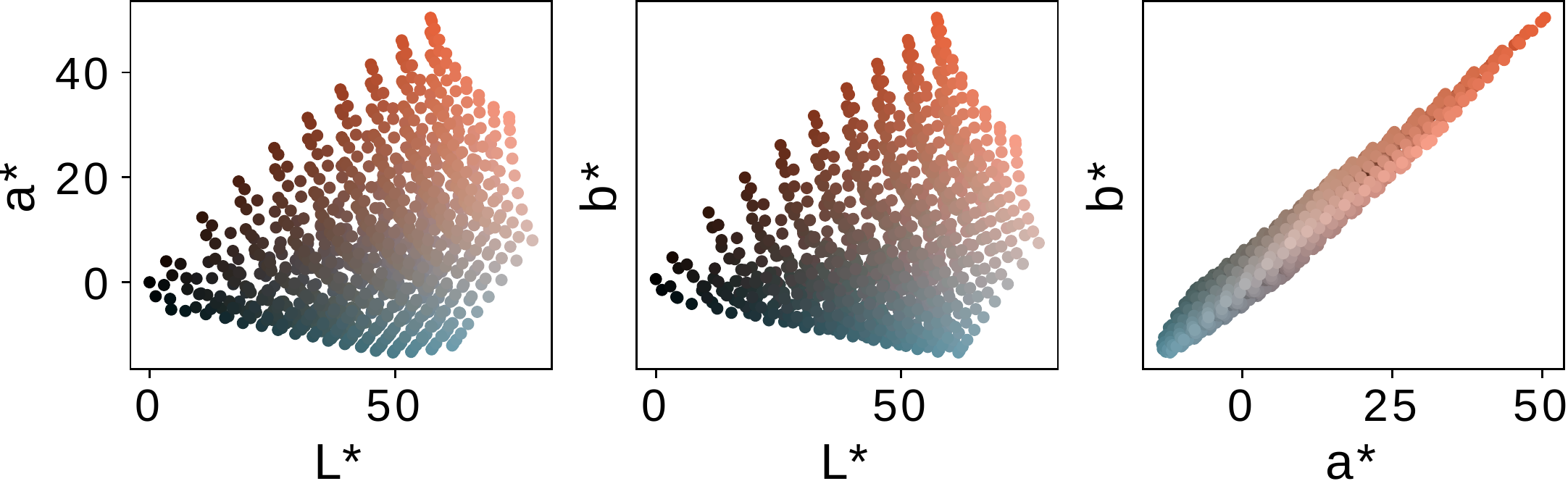} \\
    \raisebox{1\height}{\rotatebox[origin=c]{90}{\scriptsize Vector $e_3$}} & \includegraphics[width=\linewidth]{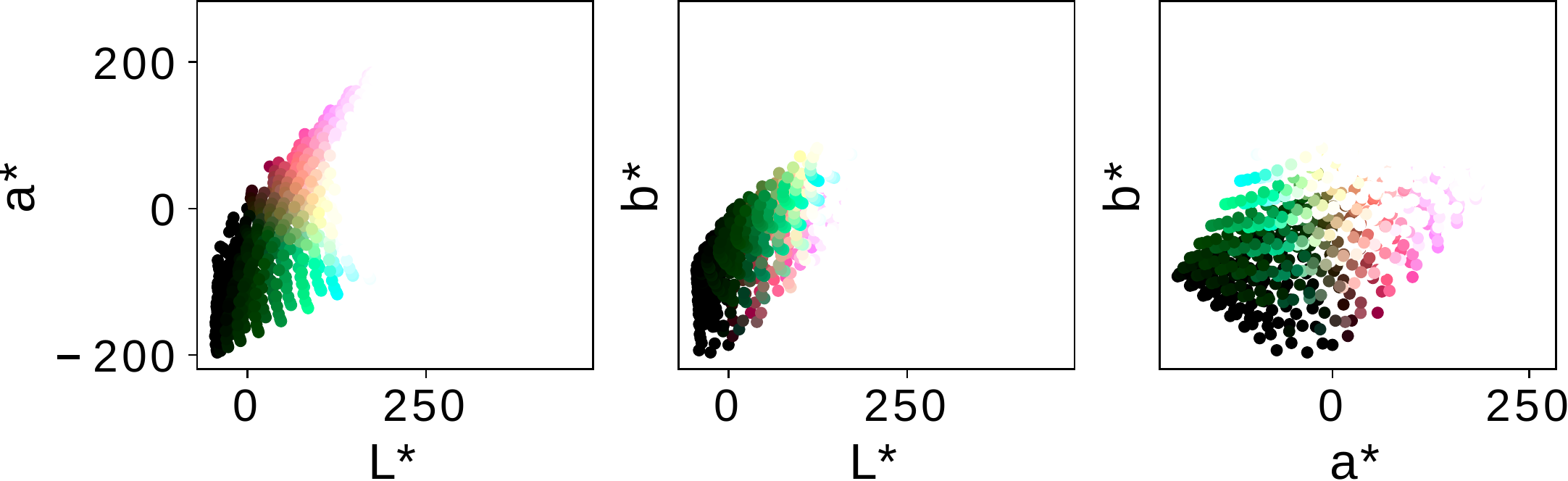} & \includegraphics[width=\linewidth]{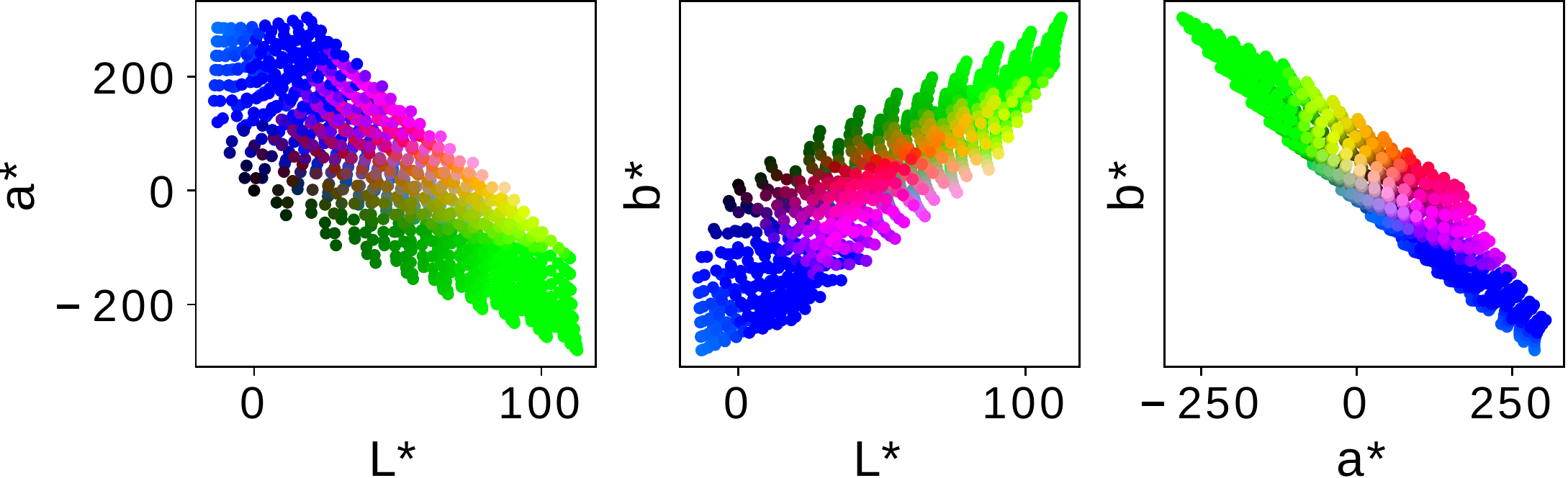} & \includegraphics[width=\linewidth]{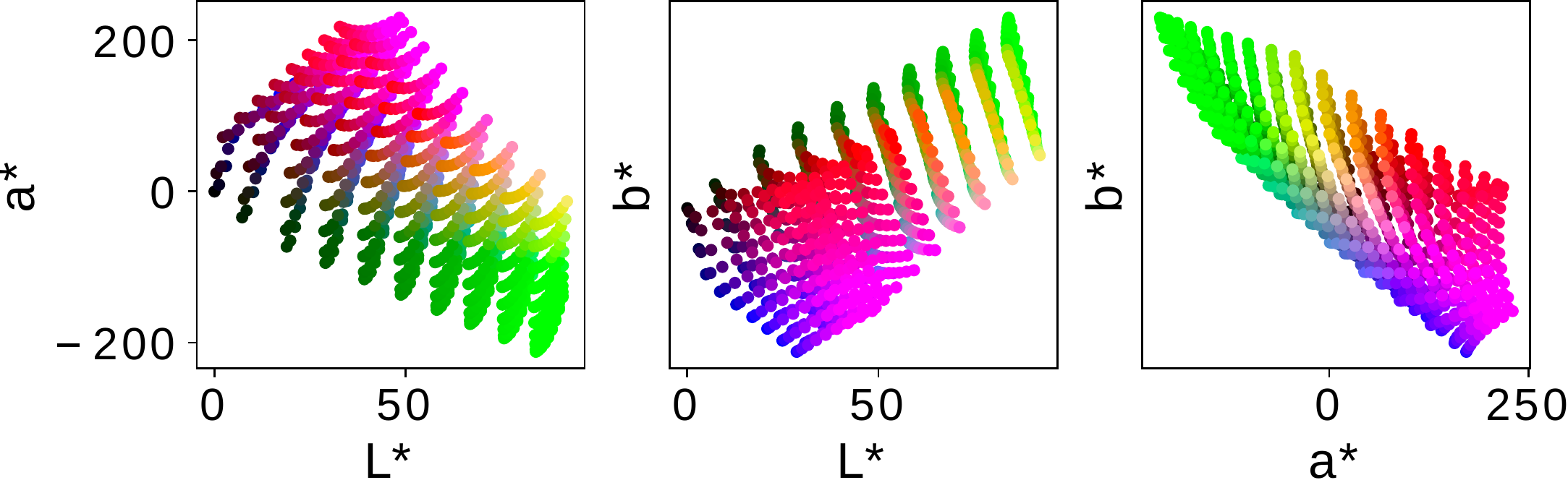} \\
    \raisebox{1\height}{\rotatebox[origin=c]{90}{\scriptsize Vector $e_4$}} & \includegraphics[width=\linewidth]{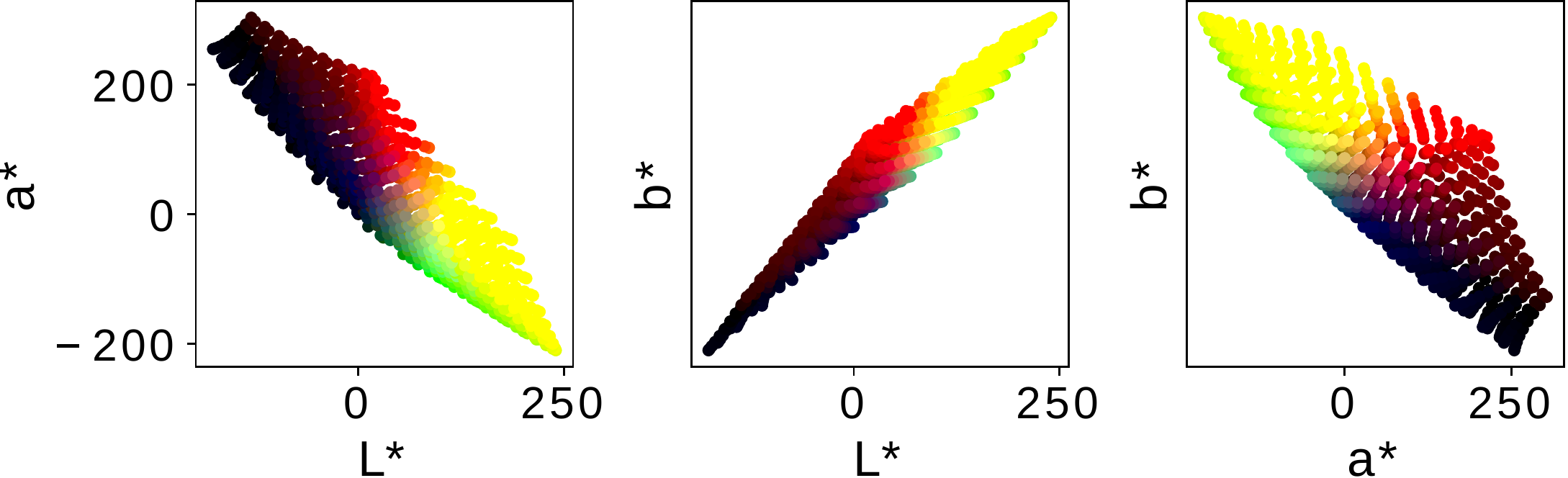} & \includegraphics[width=\linewidth]{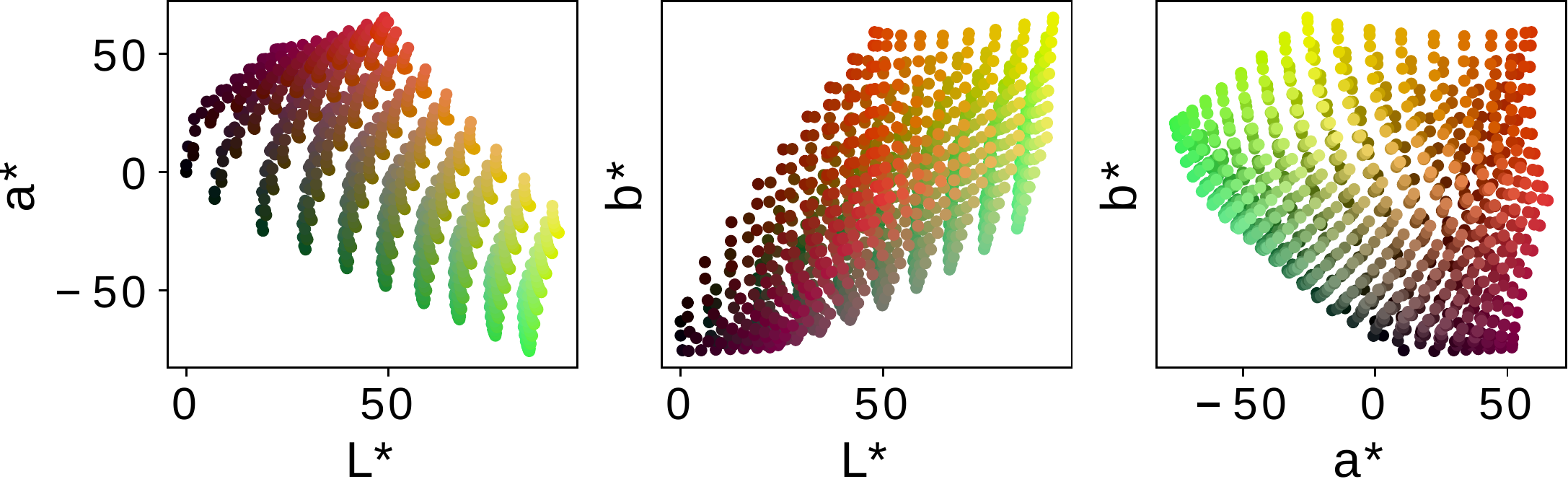} & \includegraphics[width=\linewidth]{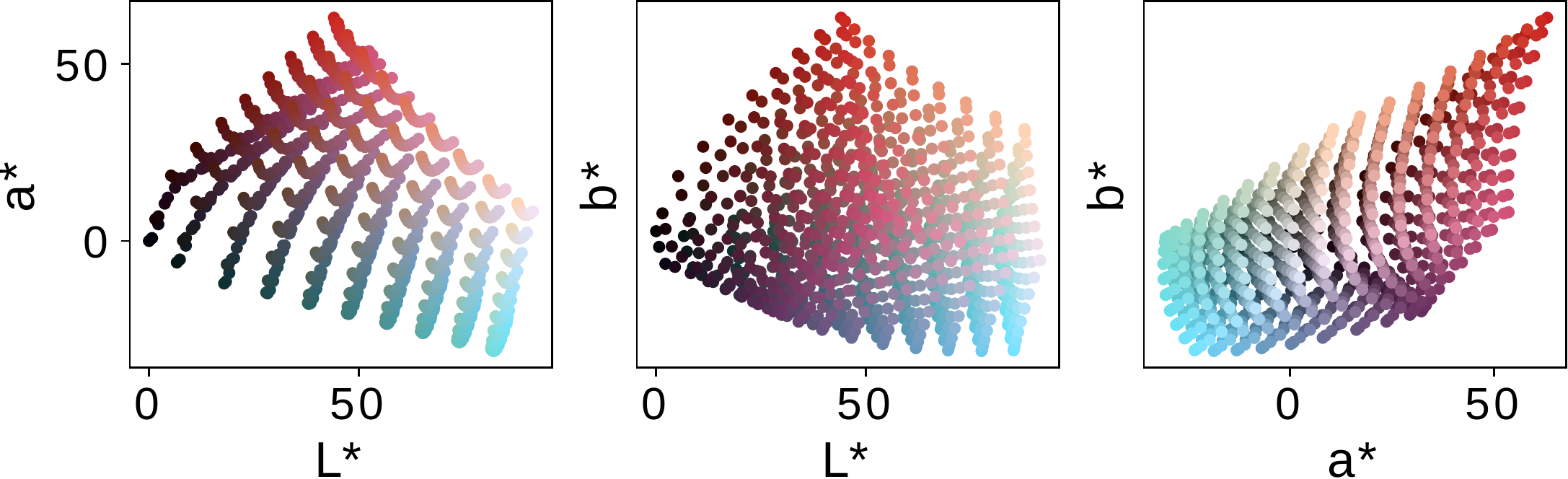} \\
    \raisebox{1\height}{\rotatebox[origin=c]{90}{\scriptsize Vector $e_5$}} & \includegraphics[width=\linewidth]{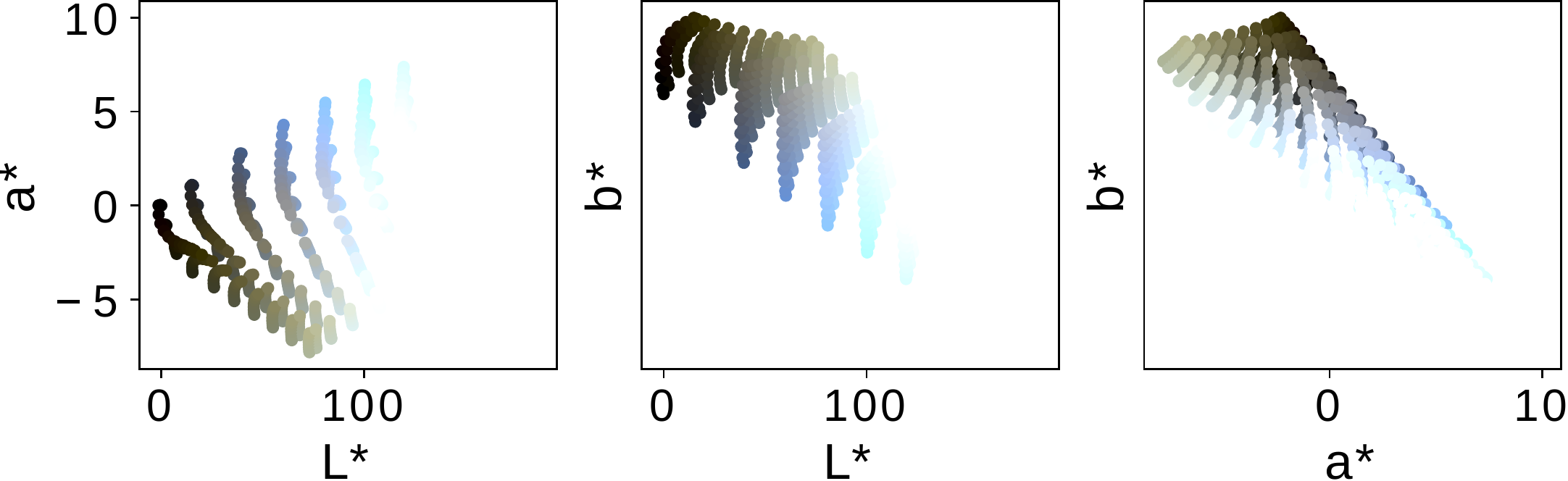} & \includegraphics[width=\linewidth]{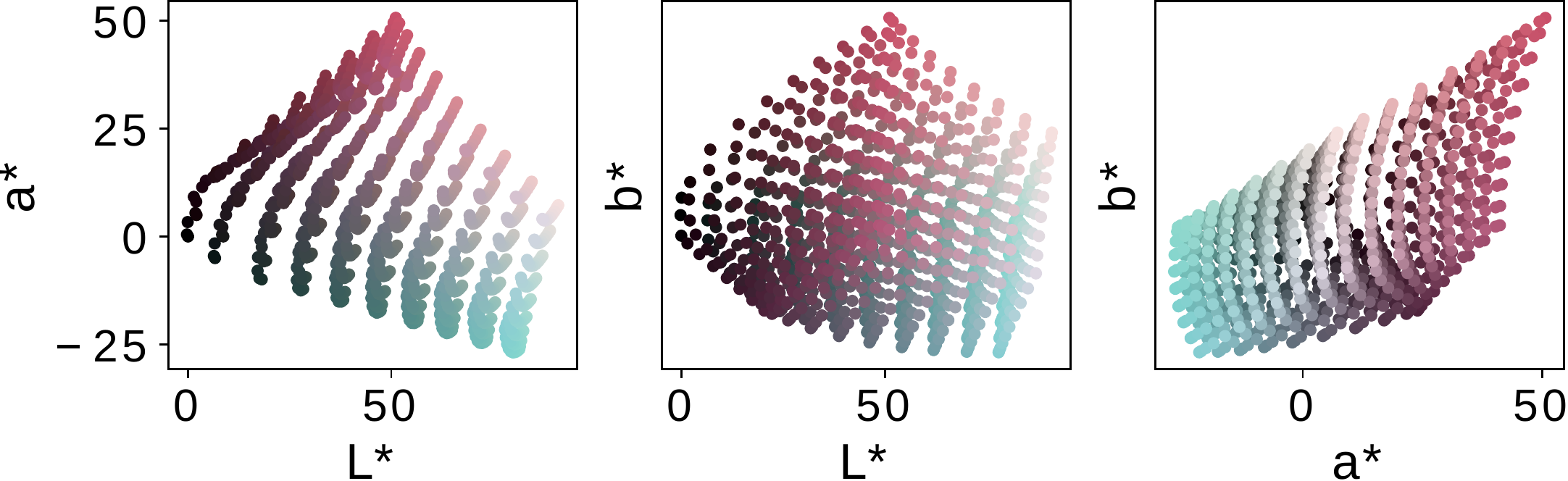} & \includegraphics[width=\linewidth]{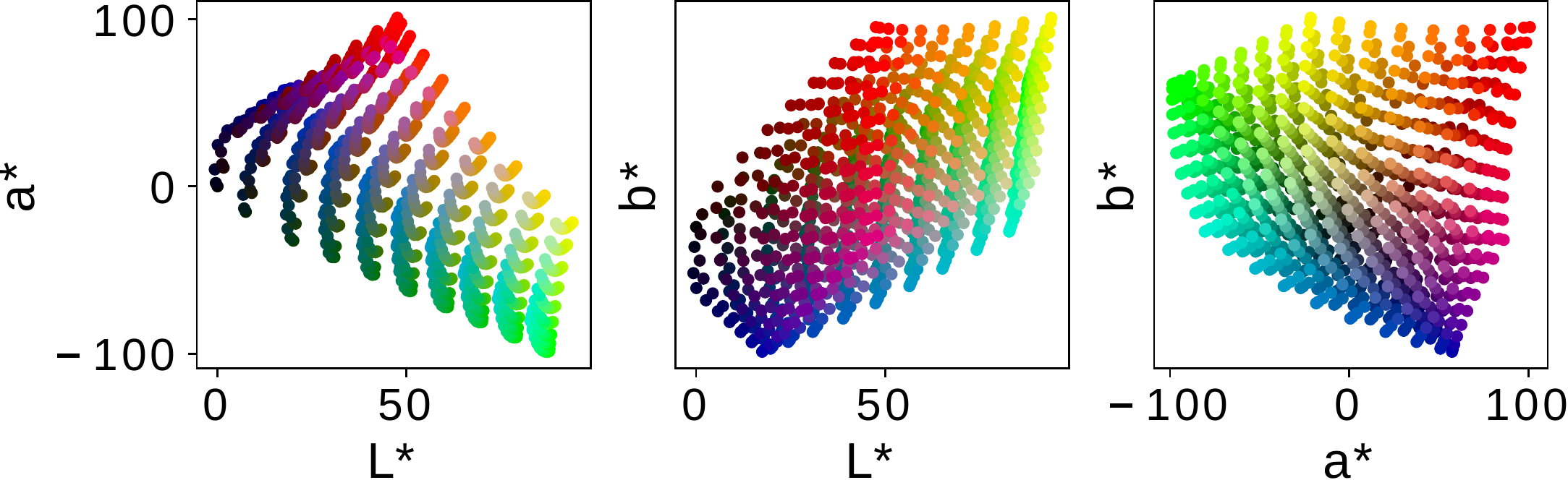} \\
    \raisebox{1\height}{\rotatebox[origin=c]{90}{\scriptsize Vector $e_6$}} & \includegraphics[width=\linewidth]{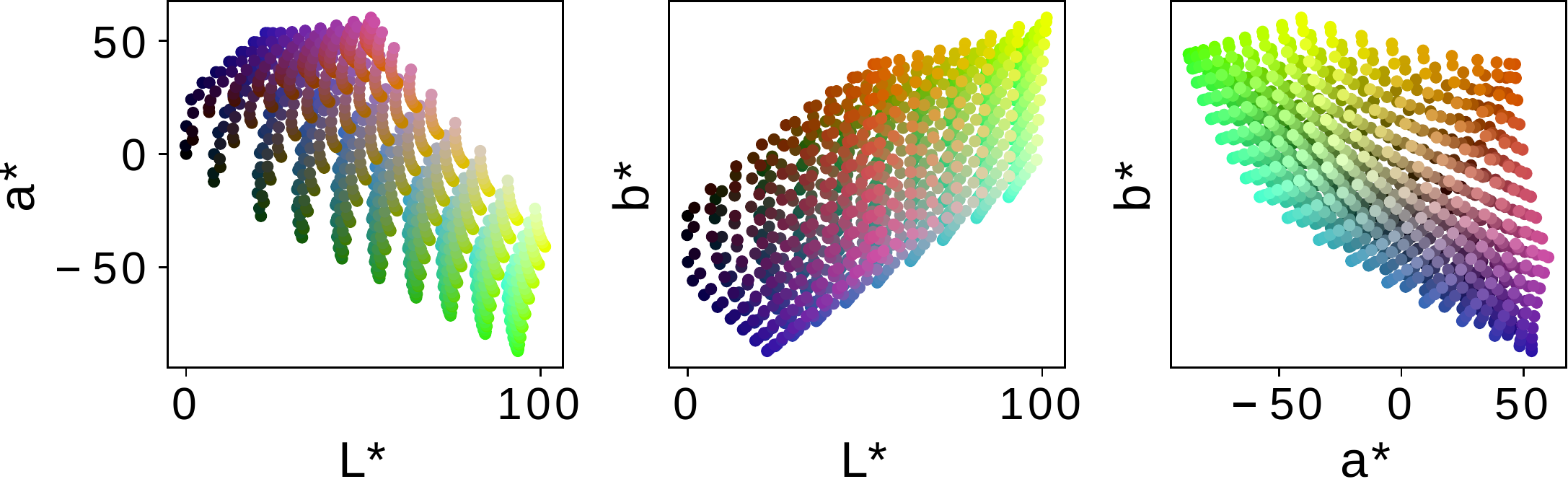} & \includegraphics[width=\linewidth]{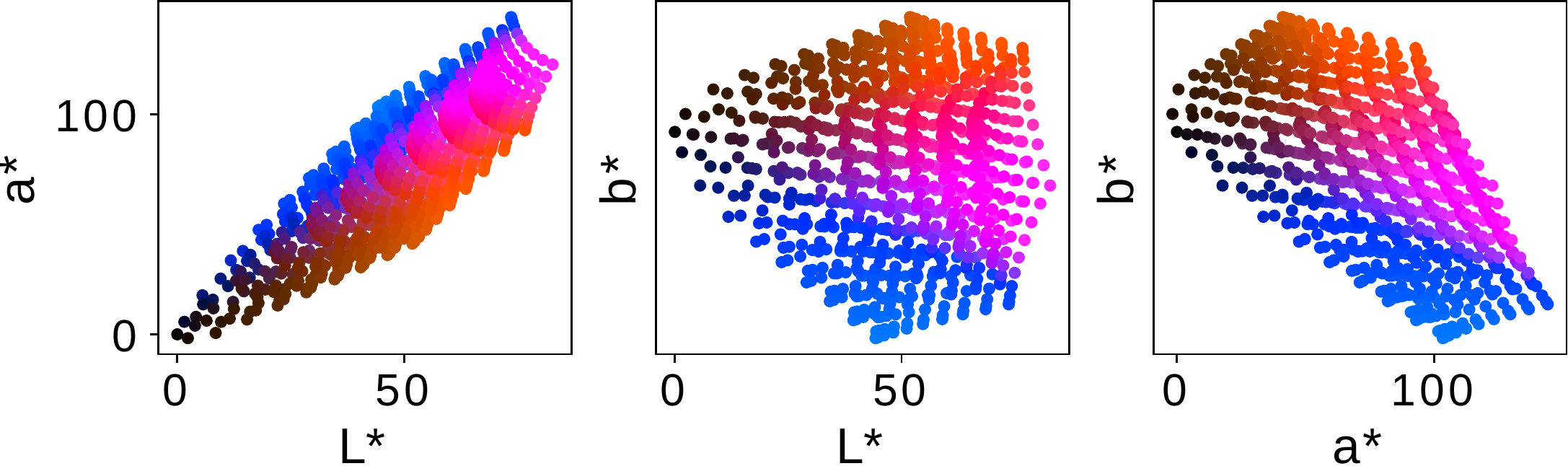} & \includegraphics[width=\linewidth]{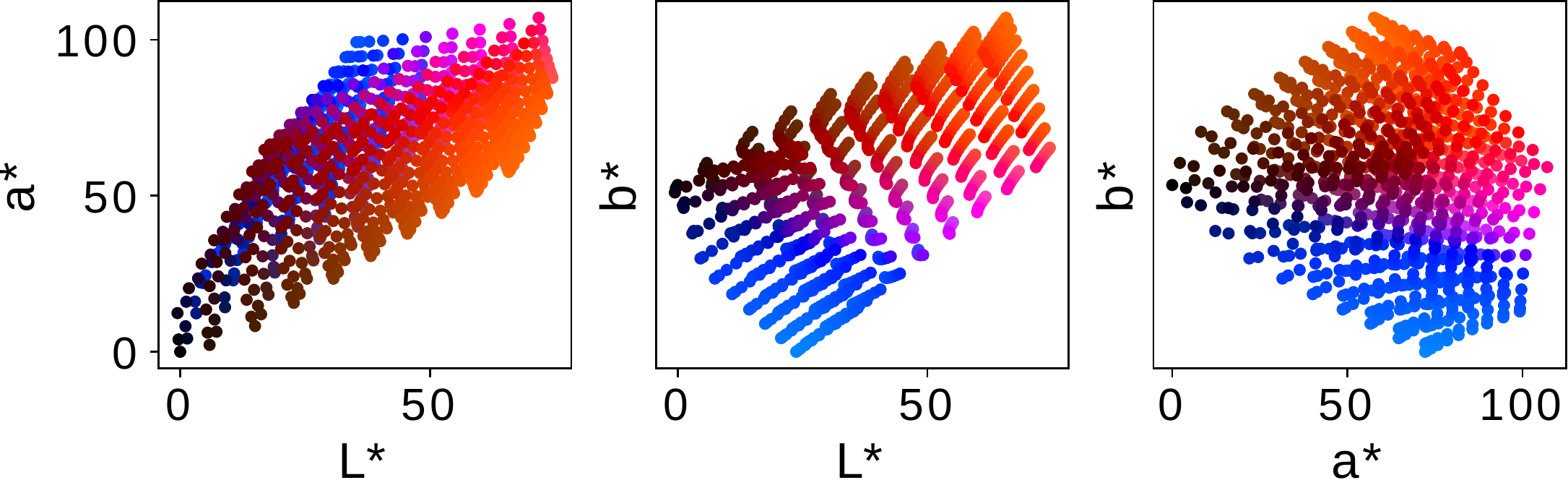} \\
    \raisebox{1\height}{\rotatebox[origin=c]{90}{\scriptsize Vector $e_7$}} & \includegraphics[width=\linewidth]{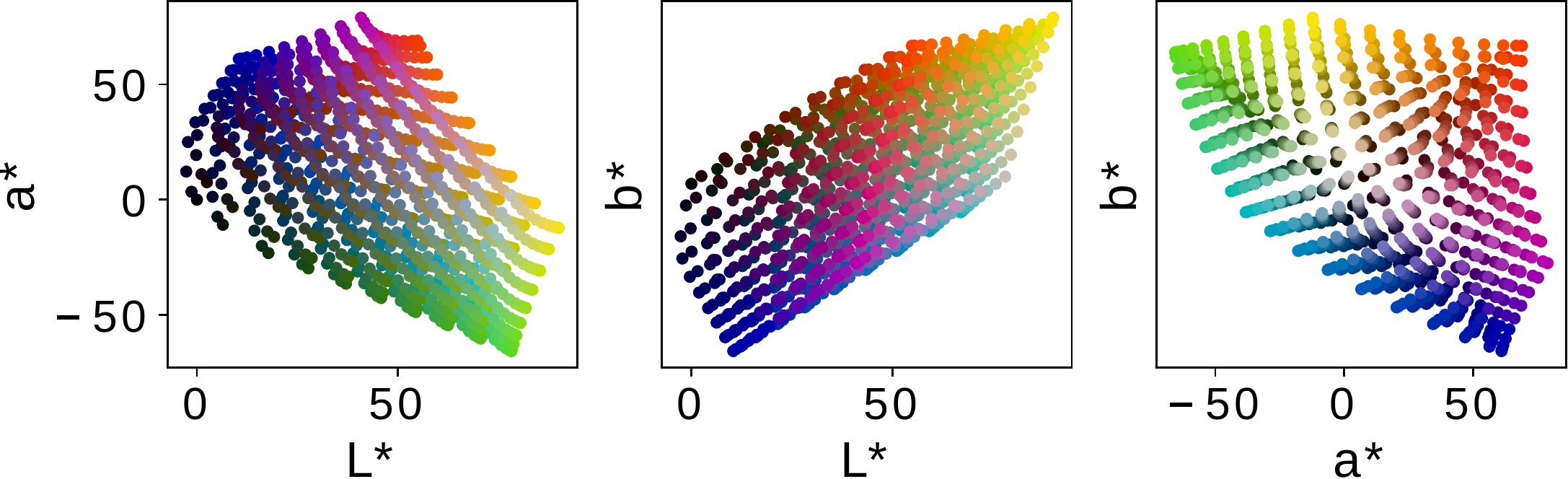} & \includegraphics[width=\linewidth]{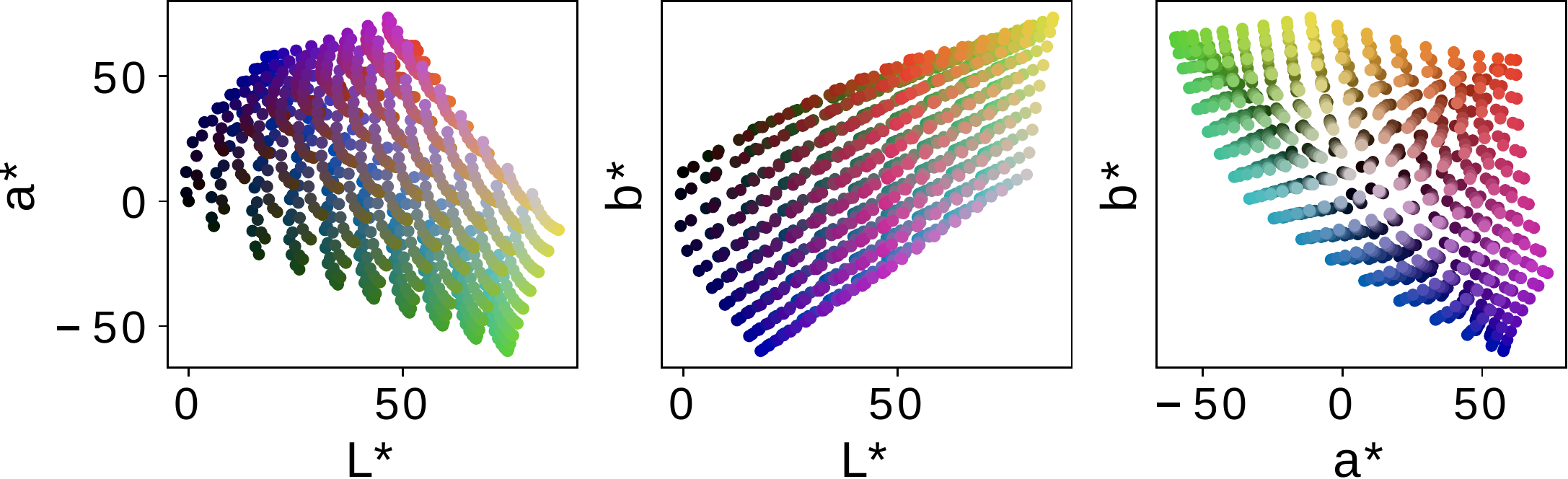} & \includegraphics[width=\linewidth]{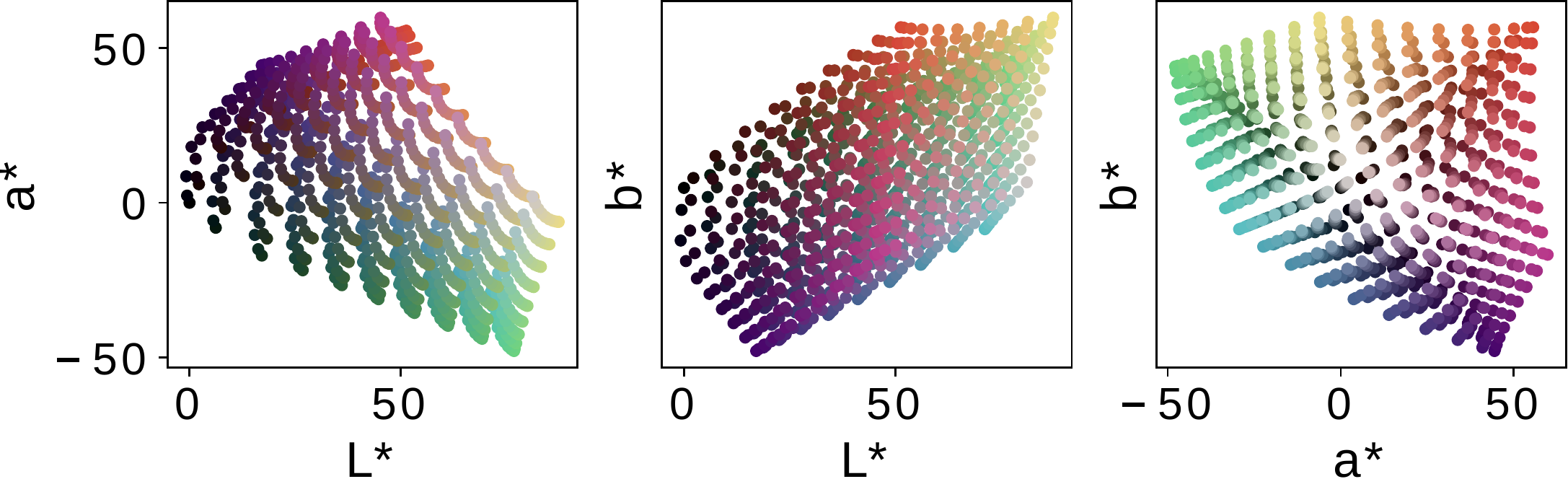} \\
    \end{tabular}
    \caption{The impact of vector lesion on RGB cube plotted in CIE L*a*b* coordinate,}
    \label{fig:trans_mats}
\end{figure}

In Figure \ref{fig:trans_mats} we have illustrated the impact of each linear transformation applied to the entire RGB cube. This gives an intuitive idea of what each vector performs in a simple glance. Absence of some vectors results in the collapse of a chromatic direction. Others shear, shrink or expand the colour space.



\end{document}